\documentclass[11pt]{article}
\usepackage[margin=1in]{geometry}
\usepackage[T1]{fontenc}
\usepackage{times}
\usepackage{microtype}
\usepackage{amsthm}

\newtheorem{theorem}{Theorem}[section]
\newtheorem{proposition}[theorem]{Proposition}
\newtheorem{lemma}[theorem]{Lemma}

\theoremstyle{remark}

\theoremstyle{plain}

\usepackage{amssymb}
\usepackage{graphicx}
\usepackage{natbib}
\usepackage[algo2e,vlined]{algorithm2e}

\usepackage{microtype}
\usepackage{mathtools}
\usepackage{booktabs}
\usepackage{array}
\usepackage{float}

\usepackage{amsfonts}
\usepackage{bm}
\usepackage{xcolor}
\usepackage{mathtools}
\usepackage{xspace}
\usepackage{booktabs}
\usepackage{microtype}
\usepackage{hyperref}
\usepackage{cleveref}

\hypersetup{
  colorlinks=true,
  breaklinks=true,
  linkcolor=blue!55!black,
  citecolor=blue!55!black,
  urlcolor=blue!55!black
}
\allowdisplaybreaks

\newcommand{\wh}{\widehat}

\newcommand{\eps}{\varepsilon}

\newcommand{\inner}[1]{\left\langle #1\right\rangle}

\DeclarePairedDelimiter{\norm}{\lVert}{\rVert}

\def\R{\mathbb{R}}
\def\E{\mathbb{E}}

\newcommand{\ind}{\mathbb{I}}

\newcommand{\Et}{\mathbb{E}_t}

 \newcommand{\Bcal}{\mathcal B}

 \newcommand{\cF}{\mathcal F}

 \newcommand{\ellhat}{\wh\ell}

\DeclareMathOperator*{\argmin}{argmin}

\newcommand{\order}{\mathcal O}
\newcommand{\otil}{\widetilde{\mathcal O}}

\newcommand{\ip}[2]{\left\langle #1,#2\right\rangle}

\newcommand{\Reg}{\mathsf{Reg}}
\newcommand{\REM}{\mathsf{Rem}}

\newcommand{\halftick}{\ensuremath{\checkmark\kern-1.1ex\raisebox{.7ex}{\rotatebox[origin=c]{125}{--}}}}

\usepackage{prettyref}
\newcommand{\savehyperref}[2]{\texorpdfstring{\hyperref[#1]{#2}}{#2}}
\newcommand{\pref}[1]{\prettyref{#1}}

\newrefformat{sec}{\savehyperref{#1}{Section~\ref*{#1}}}
\newrefformat{app}{\savehyperref{#1}{Appendix~\ref*{#1}}}
\newrefformat{eq}{\savehyperref{#1}{Eq.~\textup{(\ref*{#1})}}}
\newrefformat{thm}{\savehyperref{#1}{Theorem~\ref*{#1}}}
\newrefformat{lem}{\savehyperref{#1}{Lemma~\ref*{#1}}}
\newrefformat{prop}{\savehyperref{#1}{Proposition~\ref*{#1}}}
\newrefformat{propt}{\savehyperref{#1}{Proposition~\ref*{#1}}}
\newrefformat{assum}{\savehyperref{#1}{Assumption~\ref*{#1}}}
\newrefformat{def}{\savehyperref{#1}{Definition~\ref*{#1}}}
\newrefformat{cor}{\savehyperref{#1}{Corollary~\ref*{#1}}}
\newrefformat{rem}{\savehyperref{#1}{Remark~\ref*{#1}}}
\newrefformat{alg}{\savehyperref{#1}{Algorithm~\ref*{#1}}}
\newrefformat{fig}{\savehyperref{#1}{Figure~\ref*{#1}}}
\newrefformat{tab}{\savehyperref{#1}{Table~\ref*{#1}}}
\newrefformat{line}{\savehyperref{#1}{Line~\ref*{#1}}}

\newcommand{\DeltaK}{\Delta_K}
\newcommand{\Vself}{V_{\mathrm{self}}}
\newcommand{\Vbarself}{\bar{V}_{\mathrm{self}}}
\newcommand{\Gself}{G_{\mathrm{self}}}
\newcommand{\Fu}{F}

\definecolor{wine_red}{RGB}{228,48,64}
\definecolor{DSgray}{cmyk}{0,1,0,0}

\newcommand{\tstart}{t^{\mathrm{start}}}
\newcommand{\tend}{t^{\mathrm{end}}}

\setcitestyle{numbers,square,comma}
\let\citet\citep

\title{Near-Optimal First-Order Regret for Bandits with Feedback Graphs}
\author{
    Mengxiao Zhang\\
    University of Iowa\\
    \texttt{mengxiao-zhang@uiowa.edu}
    \and
    Yuheng Zhang\\
    University of Illinois Urbana-Champaign \\
    \texttt{yuhengz2@illinois.edu}
    \and 
\hspace*{-10pt} Haipeng Luo\\
\hspace*{-10pt} University of Southern California\\
\hspace*{-14pt}\texttt{haipengl@usc.edu}
\and 
\hspace*{20pt} Paul Mineiro\\
\hspace*{20pt} Microsoft Research\\
\hspace*{20pt} \texttt{pmineiro@microsoft.com}
}
\date{}

\SetKwInput{KwIn}{Input}
\RestyleAlgo{ruled}
\SetAlgoVlined
\DontPrintSemicolon
\LinesNumbered
\SetAlFnt{\small}

\begin{document}
\raggedbottom
\maketitle
\thispagestyle{empty}
\begin{abstract}We study first-order regret guarantees for adversarial bandits with a fixed, known directed feedback graph. For strongly observable graphs, \citet{alon2015feedback} establishes a minimax regret of $\widetilde{\Theta}(\sqrt{\alpha T})$, where $\alpha$ denotes the independence number of the graph and $T$ denotes the horizon. On the other hand, the best existing first-order bound scales as $\otil(\sqrt{(\kappa+1) L^\star})$~\citep{lee2020graph}, where $\kappa$ denotes the clique partition number, satisfying $\kappa+1\geq\alpha$, and $L^\star$ denotes the cumulative loss of the best arm, leaving a gap between the graph dependence in worst-case and first-order guarantees. In this paper, we bridge this gap by obtaining an expected regret bound of $\widetilde{O}(\sqrt{\alpha L^\star}+\alpha)$. Our first result is an algorithm for the special case of self-aware graphs, in which every vertex has a self-loop. The algorithm combines the update of EXP3~\citep{auer2002nonstochastic} with a sampling rule inspired by exploration by optimization \citep{lattimore2020exploration}. More specifically, we construct a map whose fixed points yield sampling distributions that control the one-round regret remainder in terms of the learner's loss and the graph's independence number, which is crucial in deriving optimal first-order guarantees. We further show that a sufficiently accurate fixed point can be computed in polynomial time by a geometrically convergent iteration. Finally, we extend the algorithm to general strongly observable graphs and achieve the same $\widetilde{O}(\sqrt{\alpha L^\star}+\alpha)$ guarantee by adapting the two-level aggregation approach of \citet{lee2020graph}.
\end{abstract}

\newpage
{\hypersetup{linkcolor=black}\tableofcontents}
\thispagestyle{empty}
\newpage
\setcounter{page}{1}

\section{Introduction}
\label{sec:introduction}

In this paper, we study adversarial bandits with directed feedback graphs, a model that interpolates between prediction with expert advice and multi-armed bandits \citep{mannor2011side,alon2017nonstochastic}.
The vertices of a directed graph represent actions, and an edge from $i$ to $j$ means that playing action $i$ reveals the loss of action $j$.
The learner may therefore observe losses beyond that of its chosen action, and the graph structure determines how these additional observations improve regret.
For a fixed strongly observable graph, in which every action is observed either by itself or by all other actions, \citet{alon2015feedback} established a worst-case regret bound of $\otil(\sqrt{\alpha T})$, where $\alpha$ denotes the independence number of the graph and $T$ denotes the horizon.\footnote{The notation $\otil(\cdot)$ hides factors logarithmic in the number of actions $K$ and the horizon $T$.} This bound is minimax optimal up to logarithmic factors for sufficiently large $T$.

However, a worst-case guarantee does not capture the smaller regret achievable when the loss sequence is favorable. First-order, or small-loss, guarantees adapt to the cumulative loss $L^\star$ of the best fixed action, which can be much smaller than $T$. For adversarial bandits with graph feedback, \citet{lykouris2018small} was the first to study small-loss bounds, albeit only for the special case of self-aware undirected graphs, in which every vertex has a self-loop.
Subsequently, \citet{lee2020graph} improved these results, obtaining a bound of $\otil(\sqrt{(\kappa+1)L^\star})$ for general strongly observable directed graphs, where $\kappa$ denotes the clique partition number of the subgraph induced by vertices with self-loops, namely the smallest number of bidirected cliques whose vertex sets partition this subgraph, and an improved bound of
$\otil\left(
    \min\left\{\sqrt{\alpha T},\sqrt{\kappa L^\star}\right\}
  \right)$ for self-aware directed graphs. 
Note that it holds generally that $\alpha\le\kappa+1$, and the two quantities can be far apart in some graphs where $\alpha=1$ but $\kappa$ equals the total number of actions.
These remain the best known first-order bounds for the respective class of directed graphs, and a natural question that was already raised in~\citet{lee2020graph} is whether $\otil(\sqrt{\alpha L^\star})$ regret is achievable, even for self-aware graphs. This leads to the central question studied in this paper:
\begin{center}
\emph{For multi-armed bandits with a strongly observable directed feedback graph, what is the best possible first-order regret bound and how to design algorithms to achieve it?}
\end{center}

\paragraph{Contribution.} In this paper, we show that $\otil(\sqrt{\alpha L^\star})$ near-optimal first-order regret is indeed achievable. Specifically, we first consider self-aware graphs in \pref{sec:uncorrected}, where our main technical contribution already appears.
Our algorithm builds on EXP3~\citep{auer2002nonstochastic}, viewed as online mirror descent with negative entropy. The new ingredient is the rule for selecting the sampling distribution.
Following the principle of exploration by optimization \citep{lattimore2020exploration}, we derive this distribution based on a one-round regret inequality, with the additional requirement that the regret remainder be controlled by the learner's loss.
Intuitively, the rule reduces the sampling probability of actions whose losses are poorly observed, limiting the discrepancy between the learner's loss and the loss estimates used in the update.
Because observation probabilities themselves depend on the sampling distribution, this rule defines a fixed-point condition.
At the fixed point, we control the one-round regret remainder by the learner's loss multiplied by a factor governed by the graph's independence number. To make our algorithm computationally efficient, we show that an approximate fixed point suffices, and a geometrically convergent fixed-point iteration computes one in polynomial time. The resulting algorithm achieves expected regret $\otil(\sqrt{\alpha L^\star}+\alpha)$ for self-aware graphs (\pref{thm:u-master}).

In \pref{sec:strongly-observable}, we extend this result to general strongly observable graphs, in which some actions have no self-loop but can be observed whenever any other action is played. Following the two-level design of \citet{lee2020graph}, we treat all actions with self-loops as a single meta-action handled by the fixed-point learner we have designed for the self-aware graphs, while a master algorithm, which is an online mirror descent (OMD) algorithm with a hybrid regularizer, allocates probability between this meta-action and the actions without self-loops.
Finally, a doubling scheme based on observed quantities removes the need to know $L^\star$ or $\alpha$, giving the same regret bound for every strongly observable graph (\pref{thm:strong-adaptive}).
\subsection{Related Work}

The feedback-graph model was introduced in \citet{mannor2011side} to study how side observations improve learning in adversarial bandits.
Subsequent work related regret to graph parameters such as the independence and domination numbers \citep{alon2013domination,alon2017nonstochastic}, and \citet{alon2015feedback} classified feedback graphs according to their observability and characterized the corresponding minimax rates.
More recent results sharpen this understanding in several directions.
\citet{kocak2023shape} characterized the minimax regret across different horizon regimes, while \citet{eldowa2023minimax} improved the logarithmic dependence for strongly observable undirected graphs.
Other work developed high-probability guarantees for time-varying graphs \citep{luo2023highprob} and regret decompositions based on graph partitions \citep{he2023decomposition}.

The model has also been studied under different assumptions on the feedback and loss sequences.
\citet{cohen2016without} considered feedback graphs that are never fully revealed to the learner, and \citet{esposito2022edge} studied stochastic feedback graphs with unknown edge probabilities.
With switching costs, \citet{arora2019switching} obtained regret guarantees governed by the domination number.
For stochastic losses, \citet{marinov2022stochastic} investigated the distinction between finite-time and asymptotic optimality.
Best-of-both-worlds algorithms achieve favorable guarantees for stochastic losses together with near-optimal regret for adversarial  losses~\citep{ito2022nearly,rouyer2022near}.
The feedback-graph model has also been extended to adversarial combinatorial semi-bandits \citep{wen2025combinatorial} and contextual bandits.
For contextual bandits, \citet{zhang2023practical} developed efficient algorithms with informed feedback graphs through a reduction to online regression under realizability.
\citet{zhang2024uninformed} extended this approach to uninformed graphs by learning both losses and feedback probabilities, using log loss for the latter.
\citet{wen2024contextual} studied graph feedback across actions together with cross-learning across contexts and identified graph parameters governing the statistical complexity.

Small-loss guarantees have also been studied extensively in online learning, including full-information prediction \citep{freund1997decision}, adversarial bandits and semi-bandits \citep{neu2015firstorder,wei2018adaptive}, and contextual bandits \citep{allenzhu2018minority}.
Under partial information, these guarantees require controlling the cost of estimating unobserved losses.
Relevant techniques include implicit exploration \citep{kocak2014implicit,neu2015explore}, log-barrier regularization and correction terms \citep{wei2018adaptive}, and recursive freezing \citep{allenberg2006hannan, lykouris2018small}.
For self-aware undirected feedback graphs, \citet{lykouris2018small} obtained both an independence-number-related bound of $\otil(\alpha^{1/3}(L^\star)^{2/3})$ and a clique-partition number-related bound of $\otil(\sqrt{\kappa L^\star})$.

The closest work to ours is \citet{lee2020graph}, which extended small-loss guarantees to general directed strongly observable graphs and obtained $\otil(\sqrt{(\kappa+1) L^\star})$ regret or
$\otil(\min(\sqrt{\alpha T}, \sqrt{\kappa L^\star}))$ if every vertex has a self-loop.
Their leading small-loss term depends on the clique partition number, and they left open whether the independence-number dependence of worst-case guarantees is achievable in a first-order bound.
For general strongly observable graphs, their construction treats each clique in a partition of the self-loop subgraph as a meta-action and combines the resulting learners through a master algorithm.
We build on this two-level construction, using a single fixed-point learner for the entire self-loop subgraph and a complement log barrier for actions without self-loops.
This construction gives independence-number dependence without requiring a clique partition or an independent set to be computed. For weakly observable graphs, which have an incoming edge at every vertex but are not strongly observable, \citet{lee2020graph} also established an impossibility result for standard small-loss bounds in terms of $L^\star$ (explaining why we focus on the strongly observable case).
They instead obtained alternative loss-dependent guarantees based on the losses of specific subsets of arms.

Our sampling rule is also related to exploration by optimization (ExO), which jointly optimizes a sampling distribution and a loss estimator to minimize a worst-case one-round regret bound \citep{lattimore2020exploration}.
This approach has connections to information-ratio and minimax analyses \citep{lattimore2019information,zimmert2019connections} and has been extended using hybrid regularizers \citep{tsuchiya2024exo}.
We follow the same principle of deriving exploration from the regret bound: we modify the distribution produced by OMD so that the one-round regret remainder scales with the learner's loss.

\section{Preliminaries}
\label{sec:prelim}
\paragraph{Notation.}
Denote by $[K]\triangleq\{1,\ldots,K\}$ the set of actions and by $\DeltaK$ the probability simplex over $[K]$.
For any finite nonempty set $U$, denote by $\Delta(U)\triangleq\{x\in\R_+^U:\sum_{i\in U}x_i=1\}$ the probability simplex over $U$, so $\DeltaK=\Delta([K])$.
Its relative interior $\operatorname{int}(\DeltaK)$ consists of distributions with strictly positive coordinates.
Denote by $e_k$ the $k$-th standard basis vector and by $\ind\{\cdot\}$ the indicator function.
For $u,v\in\R^K$, denote by $\ip{u}{v}\triangleq\sum_{{i=1}}^{{K}}u_iv_i$ the inner product and by $\norm{v}_1\triangleq\sum_{{i=1}}^{{K}}|v_i|$ the $\ell_1$ norm, and denote by $v^2$ the coordinatewise square of $v$.
For $\eps\ge0$, a distribution $q\in\DeltaK$ is an $\eps$-fixed point of a map $F:\DeltaK\to\DeltaK$ if $\norm{q-F(q)}_1\le\eps$.
For a differentiable convex function $h$, denote by $D_h(u,v)\triangleq h(u)-h(v)-\ip{\nabla h(v)}{u-v}$ its Bregman divergence. We use $\order(\cdot)$ to hide universal constants.

\paragraph{Problem setting.}
We consider adversarial multi-armed bandits with a feedback graph~\citep{mannor2011side}.
Let $G=(V,E)$ be a fixed and known directed graph with $V=[K]$ and $K\ge2$, and fix a horizon $T>2$.
For $i,j\in V$, denote by $i\to j$ the condition $(i,j)\in E$, meaning that playing arm $i$ reveals the loss of arm $j$.
We assume that $G$ is \emph{strongly observable} in the sense of \citet{alon2015feedback}: for every $i\in V$, either $i\to i$, or $j\to i$ for every $j\ne i$.
{Denote by $\Vself\triangleq\{i\in V:i\to i\}$ the set of vertices with self-loops, by $\Vbarself\triangleq V\setminus \Vself$ its complement, and by $s\triangleq|\Vself|$ the number of vertices with self-loops.}
{Denote by $\Gself$ the subgraph induced by these vertices.}
We call $G$ self-aware if $\Vself=V$.

Following~\citep{lykouris2018small,lee2020graph}, we consider an oblivious adversary for simplicity, which chooses arbitrary loss vectors $\ell_1,\ldots,\ell_T\in[0,1]^K$ before the interaction begins.
At round $t$, the learner selects a distribution $q_t\in\DeltaK$, draws $I_t\sim q_t$, incurs loss $\ell_{t,I_t}$, and observes $\ell_{t,i}$ for every $i$ such that $I_t\to i$.
This model interpolates between prediction with expert advice, where a complete graph with self-loops reveals all losses, and standard multi-armed bandits (MAB), where a graph containing only self-loops reveals only the selected arm's loss.

Denote by $\cF_{t-1}$ the $\sigma$-field generated by the learner's draws and internal randomness before round $t$, and by $\Et[\cdot]\triangleq\E[\cdot\mid\cF_{t-1}]$ the corresponding conditional expectation.
Every sampling distribution $q_t$ constructed below is $\cF_{t-1}$-measurable and therefore depends only on information available before the feedback at round $t$. For a fixed comparator $u\in\DeltaK$, denote by $L_T(u)\triangleq\sum_{t=1}^T\ip{u}{\ell_t}$ its cumulative loss and by $\Reg_T(u)$ the learner's expected regret against $u$, defined as
$$
  \Reg_T(u)\triangleq\E\left[\sum_{t=1}^T\ell_{t,I_t}\right]-L_T(u).
$$
For a fixed action $k$, denote by $L_{T,k}\triangleq L_T(e_k)$ its cumulative loss and by $\Reg_{T,k}\triangleq \Reg_T(e_k)$ the expected regret against it. The expected regret against the best fixed action is $\Reg_T\triangleq\max_{k\in[K]}\Reg_{T,k}=\E\bigl[\sum_{t=1}^T\ell_{t,I_t}\bigr]-L^\star$, where $L^\star\triangleq\min_{k\in[K]}L_{T,k}$.

\paragraph{Graph quantities.}

We next relate the graph structure to the observation probabilities that govern the cost of loss estimation. Denote by $\alpha$ the independence number of $G$, defined as the largest cardinality of a vertex set with no edge between any two distinct vertices in either direction. When $s>0$, $\alpha$ is also the independence number of $\Gself$, since strong observability makes every vertex without a self-loop adjacent to every other vertex. 

For a distribution $q\in\DeltaK$, denote by $Q_i^G(q)$ the probability of observing the loss of action $i$, that is,
$
  Q_i^G(q)\triangleq\sum_{j\in V:j\to i}q_j.
$
We abbreviate $Q_i^G(q)$ as $Q_i(q)$ when the graph is clear. If $i\in \Vself$, then $Q_i(q)\ge q_i$;
otherwise, strong observability gives $Q_i(q)=1-q_i$. 
Further denote by $\mathcal H^G(q)\triangleq\sum_{i\in \Vself}\frac{q_i}{Q_i^G(q)+1/(KT)}$ the decision-observation ratio sum over the vertices with self-loops, with fixed offset $1/(KT)$. The following lemma is a direct implication of \citet[Lemma~1]{kocak2014implicit} (see \pref{app:aux} for the proof), which bounds $\mathcal H^G(q)$ in terms of the independence number of $G$ and is utilized in various places of our analysis.
\begin{lemma}\label[lemma]{lem:soft-ratio}
For every distribution $q\in\DeltaK$, we have
$$
  \mathcal H^G(q)
  \le
  2\alpha\ln\left(
  1+\frac{K^3T+K}{\alpha}\right)+2
  =\order(\alpha\ln(KT)).
$$
\end{lemma}

\section{A Fixed-Point Sampling Rule for Self-Aware Graphs}
\label{sec:uncorrected}
In this section, we first consider the case where the feedback graph is self-aware, that is {$V=\Vself$}.
For such graphs, we start from EXP3 \citep{auer2002nonstochastic} and modify its sampling distribution to obtain a first-order guarantee.

\paragraph{Review of EXP3.}
We present the EXP3 update as online mirror descent (OMD) with negative entropy $h(p)\triangleq\sum_{{j=1}}^{{K}}p_j\ln p_j$ and learning rate $\eta>0$.
Starting from the uniform distribution $p_1$, EXP3 draws $I_t\sim p_t$ in round $t$ and forms the unbiased estimates $\ellhat_{t,j}\triangleq\ell_{t,j}\ind\{I_t\to j\}/Q_j(p_t)$.
It then updates
\begin{equation}
  p_{t+1}
  =\argmin_{p\in\DeltaK}\left\{\eta\ip{p}{\ellhat_t}+D_h(p,p_t)\right\}.
  \label{eq:u-omd-update}
\end{equation}
For any nonnegative estimates, a slightly more refined analysis for EXP3 (proved in \pref{prop:ordinary-potential} of \pref{app:self-aware}) gives, for every $i\in[K]$,
\begin{equation}
  \sum_{t=1}^T\ip{p_t-e_i}{\ellhat_t}
  \le\frac{\ln K}{\eta}
    +\eta\sum_{t=1}^T\sum_{j=1}^K\frac{p_{t,j}\ellhat_{t,j}^2}{1+\eta\ellhat_{t,j}}.
  \label{eq:ordinary-potential}
\end{equation}
Since $\Et[\ellhat_{t,j}]=\ell_{t,j}$, it remains to upper-bound the conditional expectation of the second term on the right hand side. Direct calculation gives
\begin{align}
  \Et\left[\frac{\ellhat_{t,j}^2}{1+\eta\ellhat_{t,j}}\right]
  =\frac{\ell_{t,j}^2}{Q_j(p_t)+\eta\ell_{t,j}}
    \le\frac{\ell_{t,j}}{Q_j(p_t)+\eta},
  \label{eq:u-exponential-moment}
\end{align}
where the inequality uses $\ell_{t,j}\le1$.
Taking expectations in \pref{eq:ordinary-potential} and applying \pref{eq:u-exponential-moment} gives
\begin{align}
  \Reg_{T,i}
  \le\frac{\ln K}{\eta}
    +\eta\E\left[\sum_{t=1}^T\sum_{j=1}^K
      \frac{p_{t,j}\ell_{t,j}}{Q_j(p_t)+\eta}\right].
  \label{eq:u-exp3-regret}
\end{align}
For $\eta\ge1/(KT)$, the last summation is at most $\sum_{t=1}^T\mathcal H^G(p_t)$ (recall $V=\Vself$), so \pref{lem:soft-ratio} implies that the regret is at most $\frac{\ln K}{\eta}+\order(\eta\alpha T\ln(KT))$.
While this leads to the worst-case near-optimal rate after tuning $\eta$, the bound scales with $T$ and does not provide a first-order guarantee.

\paragraph{New sampling distributions.}
To obtain a first-order bound, we change the sampling distribution from $p_t$ to a distribution $q_t\in\operatorname{int}(\DeltaK)$ computed from $p_t$ before observing the current feedback.
We draw $I_t\sim q_t$ and use the unbiased estimator
\begin{equation}
  \ellhat_{t,j}
  =\frac{\ell_{t,j}\ind\{I_t\to j\}}{Q_j(q_t)},
  \label{eq:u-unbiased-estimator}
\end{equation}
while using the same update \pref{eq:u-omd-update}.
Under \pref{eq:u-unbiased-estimator}, $\Et[\ellhat_{t,j}]=\ell_{t,j}$, and the calculation in \pref{eq:u-exponential-moment} holds with $Q_j(q_t)$ in place of $Q_j(p_t)$.
Denote by $\REM_t(p,q)\triangleq\sum_{j=1}^K\ell_{t,j}(q_j-\frac{p_jQ_j(q)}{Q_j(q)+\eta})$ the resulting one-round regret remainder.
The same reasoning as in \pref{eq:u-exp3-regret} now gives
\begin{equation}
  \Reg_{T,i}
  \le\frac{\ln K}{\eta} + \E\left[\sum_{t=1}^T\left(\inner{q_t-p_t,\ell_t}+\sum_{j=1}^K\frac{\eta p_{t,j}\ell_{t,j}}{Q_j(q_t)+\eta}\right)\right]=\frac{\ln K}{\eta}
    +\E\left[\sum_{t=1}^T\REM_t(p_t,q_t)\right].
  \label{eq:u-design-remainder}
\end{equation}
Suppose that, for each $p$, we can find $q$ such that
\begin{equation}
  \REM_t(p,q)\le\otil(\eta\alpha)\ip{q}{\ell_t},
  \label{eq:u-target-remainder}
\end{equation}
for every $\ell_t\in[0,1]^K$. Then, substituting it into \pref{eq:u-design-remainder} and using $\E[\sum_{{t=1}}^{{T}}\ip{q_t}{\ell_t}]=L_{T,i}+\Reg_{T,i}$ gives $\Reg_{T,i}\le\frac{\ln K}{\eta}+\otil(\eta\alpha)(L_{T,i}+\Reg_{T,i})$.
When the coefficient $\otil(\eta\alpha)$ is at most $1/2$, rearranging the terms yields $\Reg_{T,i}\le\otil\left(\frac{\ln K}{\eta} + \eta\alpha L_{T,i}\right)$. Tuning $\eta$ optimally thus leads to the desired bound.

\subsection{Fixed-Point Construction}
\label{sec:u-computation}

It remains to find $q$ from $p$ so that \pref{eq:u-target-remainder} holds.
By the definition of the remainder, we have
\begin{equation}
  \REM_t(p,q)
  =\sum_{j=1}^K\ell_{t,j}\bigl(q_j-p_j\phi(Q_j(q))\bigr),
  \label{eq:u-envelope}
\end{equation}
where $\phi(s)\triangleq s/(s+\eta)$.
To ensure \pref{eq:u-target-remainder}, we want to make sure $p_j\phi(Q_j(q))\propto q_j$, or in other words,
\begin{equation}
q_j = \frac{p_j\phi(Q_j(q))}{\sum_{i=1}^K p_i\phi(Q_i(q))},
  \label{eq:u-coordinate-design}
\end{equation}
which then leads to
$$
\REM_t(p,q) = \left(1-\sum_{i=1}^K p_i\phi(Q_i(q))\right)\inner{q, \ell_t} = Z_0(p,q) \inner{q, \ell_t}
$$
with $Z_0(p,q)\triangleq1-\sum_{{i=1}}^{{K}}p_i\phi(Q_i(q))=\sum_{j=1}^K\frac{\eta p_j}{Q_j(q)+\eta}$ being the remainder coefficient.
The remaining task would then be to show $Z_0(p,q)=\otil(\eta\alpha)$.

Note that \pref{eq:u-coordinate-design} establishes a fixed point problem.
We argue below that a solution must exist for any $p\in\operatorname{int}(\DeltaK)$.
Indeed, for every $q\in\DeltaK$, self-awareness gives $Q_j(q)\ge q_j$, so at least one observation probability is positive.
Since $\phi(s)\in(0,1)$ for $s>0$, the normalizer $1-Z_0(p,q)=\sum_{{j=1}}^{{K}}p_j\phi(Q_j(q))$ is strictly positive on the simplex.
The normalized map with coordinates $p_j\phi(Q_j(q))/(1-Z_0(p,q))$ is therefore continuous and maps the simplex into itself.
Brouwer's fixed point theorem \citep{brouwer1911abbildung} then proves the existence of a solution to \pref{eq:u-coordinate-design}.

However, there are three caveats in this derivation.
First, Brouwer's theorem shows the existence of the fixed point but gives no efficient computation procedure. The normalized map associated with \pref{eq:u-coordinate-design} need not admit a uniform contraction factor below one. Second, the fixed point may lie on the boundary with $Q_j(q)=0$ for some action $j$, where the unbiased estimator \pref{eq:u-unbiased-estimator} is not defined for this action.
Third, the remaining goal of bounding $Z_0(p,q)=\sum_{j=1}^K\frac{\eta p_j}{Q_j(q)+\eta}$ by $\otil(\eta\alpha)$ cannot be achieved in general.
To see this, note that 
for every coordinate with $Q_j(q)>0$, \pref{eq:u-coordinate-design} gives the exact identity
\begin{equation}\label{eq:u-ratio}
  \frac{p_j}{Q_j(q)+\eta}
  =(1-Z_0(p,q))\frac{q_j}{Q_j(q)}.
\end{equation}
This produces the unregularized ratios $q_j/Q_j(q)$, whereas \pref{lem:soft-ratio} bounds the sum with denominator $Q_j(q)+1/(KT)$.
Without the additional $\frac{1}{KT}$ term in the denominator, if the coordinates of $q$ have no positive lower bound, then $\sum_{j=1}^K\frac{q_j}{Q_j(q)}$ can be as large as $\Omega(K)$ even when $\alpha = 1$. In fact, \pref{prop:u-unsmoothed-obstruction} in \pref{app:u-unsmoothed} gives an example with $\alpha=1$ and $\eta\in(0,\frac{1}{K+1})$ in which every fixed point must have $Z_0(p,q)\ge \eta(K+1)/3$, invalidating the goal $Z_0(p,q)=\otil(\eta\alpha)$. 

\paragraph{Smoothing and efficient computation.}
Fortunately, all three caveats can be addressed by the following smoothing process.
Specifically, we add the positive term $\frac{1}{KT(s+\eta)}$ to the multiplier $\phi(s)$, leading to a new smoothed multiplier
$\psi(s)\triangleq\phi(s)+\frac{1}{KT(s+\eta)}=\frac{s+1/(KT)}{s+\eta}$, and replace the normalized map in \pref{eq:u-coordinate-design} by the following new normalized map
\begin{equation}
  [\Fu_p(q)]_j
  \triangleq\frac{p_j\psi(Q_j(q))}{\sum_{{i=1}}^{{K}}p_i\psi(Q_i(q))}.
  \label{eq:u-map}
\end{equation}
Denote by $Z(p,q)\triangleq1-\sum_{{j=1}}^{{K}}p_j\psi(Q_j(q))$ the corresponding remainder coefficient.

To resolve the first caveat, we compute the smoothed fixed point by Picard iteration~\citep{picard1890memoire,banach1922operations}, repeatedly replacing $q$ by $\Fu_p(q)$. Via some calculation, we show that its convergence is controlled by the derivative of $\ln\psi(s)$ with respect to $\ln s$, namely $s\psi'(s)/\psi(s)$: if $0\le s\psi'(s)/\psi(s)\le C<1$ for every $s>0$, then the iterates converge geometrically at rate $C$. Note that without smoothing, $s\phi'(s)/\phi(s)=\eta/(s+\eta)$ tends to one as $s$ approaches zero, so no such $C$ exists. After smoothing, on the other hand, $s\psi'(s)/\psi(s)\le1-1/\sqrt{\eta KT}$ for every $s\ge0$ when $\eta\ge1/(KT)$, so such $C$ exists.
The final guarantee is summarized in the next proposition, where we bound the $\ell_1$ residual of each iterate and give an iteration count for the stopping rule. The proof is deferred to \pref{app:computation}.

\begin{proposition}
\label[proposition]{prop:u-contraction}
Let $G$ be self-aware, $p\in\operatorname{int}(\DeltaK)$ and $1/(KT)\le\eta\le1$. Fix any $q^{(0)}\in\operatorname{int}(\DeltaK)$ and set $q^{(n+1)}=\Fu_p(q^{(n)})$ for $n\ge0$. Then, for every $n\ge1$,
\begin{equation}
  \norm{q^{(n)}-\Fu_p(q^{(n)})}_1
  \le2\left(1-\frac1{\sqrt{\eta KT}}\right)^{n-1}\ln(\eta KT).
  \label{eq:u-contraction}
\end{equation}
Consequently, for $0<\eps\le1/2$, the iteration reaches an $\eps$-fixed point after \begin{equation}
  N
  =\order\left(\sqrt{\eta KT}\ln\frac {1+\ln(\eta KT)}{\eps}\right)
  \label{eq:u-iterations}
\end{equation}
iterations, each of which takes $\order(K^2)$ time.
\end{proposition}

As for the second caveat, note that $\psi(s)>0$ for every $s\ge0$.
Thus, \pref{eq:u-map} is continuous and maps the simplex into its \emph{interior}, and the output of Picard iteration always makes sure that every observation probability is  positive and hence the estimator \pref{eq:u-unbiased-estimator} is well defined.

Finally, we resolve the third caveat by showing that $Z(p,q)=\otil(\eta\alpha)$ indeed holds.
At the fixed point, \pref{eq:u-map} gives $(1-Z(p,q))q_j=p_j\psi(Q_j(q))$.
For $\eta\ge1/(KT)$, we have $\psi(s)\le1$; thus, $1-Z(p,q) \leq 1$ and 
\begin{equation}
  \frac{p_j}{Q_j(q)+\eta}
  =(1-Z(p,q))\frac{q_j}{Q_j(q)+1/(KT)}
  \le\frac{q_j}{Q_j(q)+1/(KT)}
  \label{eq:u-smoothed-ratio}
\end{equation}
(note the extra term $1/(KT)$ in the equality compared to \pref{eq:u-ratio}).
The definition of $Z(p,q)$ now gives
\begin{align*}
  Z(p,q)
  =\sum_{j=1}^Kp_j\bigl(1-\psi(Q_j(q))\bigr)=\left(\eta-\frac1{KT}\right)\sum_{j=1}^K\frac{p_j}{Q_j(q)+\eta}\le\eta\mathcal H^G(q)
  =\otil(\eta\alpha),
\end{align*}
where the second equality uses $\psi(s)=\frac{s+1/(KT)}{s+\eta}$, the inequality is due to \pref{eq:u-smoothed-ratio}, and the last step uses \pref{lem:soft-ratio}.

It remains to bound the smoothing and numerical residual terms in the regret decomposition.
At an exact fixed point of the smoothed map, \pref{eq:u-envelope} gives
\begin{align}
  \REM_t(p,q)
  &=\sum_{j=1}^K\ell_{t,j}\left(q_j-p_j\psi(Q_j(q))
    +\frac{p_j}{KT(Q_j(q)+\eta)}\right)\nonumber\\*
  &=\sum_{j=1}^K\ell_{t,j}\left(Z(p,q)q_j
    +\frac{p_j}{KT(Q_j(q)+\eta)}\right)\nonumber\\*
  &=Z(p,q)\ip{q}{\ell_t}
    +\frac1{KT}\sum_{j=1}^K\frac{p_j\ell_{t,j}}{Q_j(q)+\eta},
  \label{eq:u-target-bridge-exact}
\end{align}
where the first equality uses $\phi(s)=\psi(s)-\frac1{KT(s+\eta)}$, and the second uses $(1-Z(p,q))q_j=p_j\psi(Q_j(q))$ from \pref{eq:u-map}.
\pref{eq:u-smoothed-ratio} thus also bounds the second term of \pref{eq:u-target-bridge-exact} by $\mathcal H^G(q)/(KT)$.

For $\eta\ge1/(KT)$ and an approximate fixed point with $\norm{q-\Fu_p(q)}_1\le\eps\le1/(KT^2)$, we show in \pref{app:u-proof} that this approximation adds at most $\eta KT\eps\ip{q}{\ell_t}+2\eps$ to the bound obtained from \pref{eq:u-target-bridge-exact}. Specifically,
\begin{equation}
  \REM_t(p,q)
  \le\eta\left(\mathcal H^G(q)+KT\eps\right)\ip{q}{\ell_t}
    +\frac{\mathcal H^G(q)}{KT}+2\eps.
  \label{eq:u-bridge-constants}
\end{equation}
Setting $\eps=1/(KT^2)$ makes the coefficient of the learner's loss $\ip{q}{\ell_t}$ at most $\order(\eta\alpha\ln(KT))$ by \pref{lem:soft-ratio}. Since $\mathcal H^G(q)\le K$, the remaining terms sum to at most $1+2/(KT)$ over $T$ rounds. Thus we obtain the loss-dependent control in \pref{eq:u-target-remainder} up to a constant cumulative cost. 

Putting everything together, we show the complete algorithm in \pref{alg:uncorrected} and its first-order regret bound $\otil(\sqrt{\alpha L^\star}+\alpha)$ in the following theorem. We state the guarantee for any bound $A$ such that $\mathcal H^G(q_t)\le A$ for all $t\in[T]$. By \pref{lem:soft-ratio}, $A=\order(\alpha\ln(KT))$ is always valid, though the independence number $\alpha$ is NP-hard to compute in general.

\begin{algorithm2e}[t]
\caption{EXP3 with sampling from an approximate fixed point}
\label{alg:uncorrected}
\KwIn{horizon $T$, learning rate $\eta>0$}
Set $\eps=1/(KT^2)$ and $p_1$ to the uniform distribution\;
\For{$t=1,\ldots,T$}{
  Starting from the uniform distribution, iterate the map \pref{eq:u-map} until the current iterate $q_t$ satisfies $\norm{q_t-\Fu_{p_t}(q_t)}_1\le\eps$\;
  Draw $I_t\sim q_t$, observe $\ell_{t,j}$ for every $j$ with $I_t\to j$, and form loss estimators \pref{eq:u-unbiased-estimator}\;
  Update $p_{t+1}$ by \pref{eq:u-omd-update}\;
}
\end{algorithm2e}

\begin{theorem}
\label{thm:u-master}
Fix a self-aware graph $G$ and a bound $A\ge1$ such that $\mathcal H^G(q_t)\le A$ for all $t\in[T]$.
For $1/(KT)\le\eta\le1/(12A)$, \pref{alg:uncorrected} guarantees that, for every $i\in[K]$,
\begin{equation}
  \Reg_{T,i}\le
    \frac{3\ln K}{\eta}
    +10A\eta L_{T,i}.
  \label{eq:u-master}
\end{equation}
Picking $\eta=\max\{1/(KT),\min\{1/(12A),\sqrt{\ln K/(AL^\star)}\}\}$ gives $\Reg_T=\order(\sqrt{AL^\star\ln K }+A\ln K)$.
By \pref{lem:soft-ratio}, we can take $A=\order(\alpha\ln(KT))$, yielding $\Reg_T=\otil(\sqrt{\alpha L^\star}+\alpha)$.
\end{theorem}

The complete proof is deferred to \pref{app:u-proof}. Each round takes $\otil(K^2\sqrt{\eta KT})$ time by \pref{prop:u-contraction}. While the choice of $\eta$ in \pref{thm:u-master} requires the knowledge of $\alpha$ and $L^\star$, we show in \pref{thm:u-adaptive} of \pref{app:u-adaptive} that a doubling procedure achieves the same first-order rate without knowing $L^\star$ or $\alpha$. To our knowledge, this is the first first-order guarantee with independence-number dependence for self-aware directed graphs, improving the $\otil(\min\{\sqrt{\alpha T},\sqrt{\kappa L^\star}\})$ result obtained in \citet{lee2020graph}, where $\kappa$ is the clique partition number which can be significantly larger than $\alpha$. For example, orient every edge of a complete graph in one direction and add self-loops. Then $\alpha=1$ but $\kappa=K$, since no two distinct vertices observe each other in both directions.

\section{From Self-Aware to Strongly Observable Graphs}
\label{sec:strongly-observable}

In this section, we extend our algorithm to general strongly observable graphs, where a vertex without a self-loop is observed by every other vertex.
We follow the master-base construction of \citet{lee2020graph}, replacing their clique learners by a single self-aware learner  on the entire self-loop subgraph. The self-aware learner is a variant of \pref{alg:uncorrected} with an adaptive learning rate, similar to the clique learners of \citet{lee2020graph}.

\subsection{Algorithm Description}
Recall the sets $\Vself$ and $\Vbarself$ and the induced subgraph $\Gself$ from \pref{sec:prelim}.
Denote by $\Bcal$ the master action set, consisting of one action for each vertex in $\Vbarself$ and, when $s>0$, a meta-action $0$ representing $\Gself$. Denote by $M\triangleq|\Bcal|$ its size.
Recall that $\Delta(\Vself)$ and $\Delta(\Bcal)$ denote the probability simplices on these sets.

Our algorithm is given in \pref{alg:strong}, following Algorithm~1 of \citet{lee2020graph}.
Specifically, we maintain a master distribution $x_t\in\Delta(\Bcal)$ and, when $0\in\Bcal$, a self-aware base learner with an OMD distribution $p_t\in\Delta(\Vself)$ and a sampling distribution $q_t\in\Delta(\Vself)$.
The master draws $B_t\sim x_t$. If $B_t=0$, the self-aware learner draws $I_t\sim q_t$, and otherwise the master plays $I_t=B_t$ directly.
Denote by $\pi_t\in\DeltaK$ the resulting distribution over actions (\pref{line:strong-sample}), so that
\begin{equation}
  \pi_{t,i}=\begin{cases}
    x_{t,0}q_{t,i}, & i\in \Vself,\\
    x_{t,i}, & i\in\Vbarself.
  \end{cases}
  \label{eq:strong-play-distribution}
\end{equation}
After playing $I_t$, we construct loss estimates for both the self-aware learner and the master.
For the self-aware learner, we use every observed loss and set
\begin{equation}
  \widetilde\ell_{t,i}
  =\frac{\ell_{t,i}\ind\{I_t\to i\}}
    {Q_i(\pi_t)},
  \label{eq:strong-base-estimator}
\end{equation}
for every $i\in \Vself$ (\pref{line:strong-base-est}).
For the master learner, denote by $g_t\in\R^{\Bcal}$ its loss vector, with $g_{t,0}\triangleq\ip{q_t}{\ell_t}$ when $0\in\Bcal$ and $g_{t,i}\triangleq\ell_{t,i}$ for $i\in\Vbarself$. Then, we estimate the loss of meta-action $0$ by its sampled loss weighted by $\frac{1}{x_{t,0}}$, and use the standard inverse propensity weighting estimator on $\Vbarself$ (\pref{line:strong-master-est}):
\begin{equation}
  \begin{aligned}
    \widetilde g_{t,0}&=\ell_{t,I_t}\ind\{B_t=0\}/x_{t,0},\\
    \widetilde g_{t,i}&=\ell_{t,i}\ind\{I_t\ne i\}/(1-x_{t,i}).
  \end{aligned}
  \label{eq:strong-master-estimator}
\end{equation}
Direct calculation shows that $\Et[\widetilde g_t]=g_t$.
In the following, we introduce how we update the strategy for both the base and the master learner. 

\paragraph{Base learner.}
For the base learner, we combine the fixed-point sampling rule of \pref{alg:uncorrected} with an adaptive EXP3 update on a clipped simplex, with the latter one similar to Algorithm~3 of \citet{lee2020graph}. The adaptive learning rate is necessary as the self-aware learner's loss estimates are scaled by $1/x_{t,0}$, which is random and set by the master (see \pref{eq:base_stab}). With an adaptive learning rate, the resulting regret grows only with the expected largest scaling factor, a cost that the master is designed to absorb (see \pref{lem:strong-components}). Similarly, the clique learners of \citet{lee2020graph} receive loss estimates scaled by the inverse probability of their meta-action and run an adaptive learning rate whose regret bound scales with the largest loss estimate (see Algorithm~3 and Lemma~11 of \citet{lee2020graph}).

Specifically, we use OMD with an entropy regularizer on the clipped simplex $\mathcal P\triangleq\{p\in\Delta(\Vself):p_i\ge1/(sT)\text{ for all }i\in\Vself\}$, where the clipped domain helps control the Bregman divergence when the learning rate changes, as in~\citet{lee2020graph}. We then update $p_t$ with the loss estimates in \pref{eq:strong-base-estimator} and an adaptively tuned learning rate $\eta_t$ chosen before sampling (\pref{line:strong-p-update}).
Note that selecting meta-action $0$ alone reveals $i$ with probability $x_{t,0}Q_i^{\Gself}(q_t)$ while sampling other actions in ${\Vbarself}$ may provide additional observations. Thus $Q_i(\pi_t)\ge x_{t,0}Q_i^{\Gself}(q_t)$, and the calculation in \pref{eq:u-exponential-moment} gives
\begin{align}
  \Et\left[\frac{\eta_t\widetilde\ell_{t,i}^2}{1+\eta_t\widetilde\ell_{t,i}}\right]
  &=\frac{\eta_t\ell_{t,i}^2}{Q_i(\pi_t)+\eta_t\ell_{t,i}}
    \le\frac{(\eta_t/x_{t,0})\ell_{t,i}}{Q_i^{\Gself}(q_t)+\eta_t/x_{t,0}}.
  \label{eq:strong-base-moment}
\end{align}
Therefore, the sampling map uses parameter $\eta_t/x_{t,0}$:
\begin{equation}
  \begin{aligned}
    [\Fu^{\mathrm{self}}_t(q)]_i
    &\triangleq\frac{p_{t,i}\psi_t(Q_i^{\Gself}(q))}{\sum_{j\in\Vself}p_{t,j}\psi_t(Q_j^{\Gself}(q))},\\
    \psi_t(v)&\triangleq\frac{v+1/(KT)}{v+\eta_t/x_{t,0}}.
  \end{aligned}
  \label{eq:strong-base-map}
\end{equation}
We then compute an interior $\eps$-fixed point $q_t$ of this map (\pref{line:strong-fixed-point}), that is, to find $q_t$ such that $\|q_t-\Fu^{\mathrm{self}}_t(q_t)\|_1\leq \eps$ with $\eps=\frac{1}{KT^2}$. To tune this adaptive EXP3 update, we combine the last term of \pref{eq:ordinary-potential} with the loss discrepancy $\ip{q_t-p_t}{\ell_t}$ between playing $q_t$ and $p_t$, and use the fixed-point construction to show
\begin{align}
  \ip{q_t-p_t}{\ell_t}
  +\Et\left[\sum_{i\in\Vself}\frac{\eta_t p_{t,i}\widetilde\ell_{t,i}^2}{1+\eta_t\widetilde\ell_{t,i}}\right]\le\frac{2A\eta_t}{x_{t,0}}g_{t,0}+\frac2T+4\eps, \label{eq:base_stab}
\end{align}
where $A$ is such that $\mathcal H^{\Gself}(q_t)\le A$ (note the extra $1/x_{t,0}$ factor compared to the derivations in \pref{sec:uncorrected}).
Since $\Et[\widetilde g_{t,0}/x_{t,0}]=g_{t,0}/x_{t,0}$, we use these estimates (ignoring the lower-order $\frac{2}{T}+4\eps$ terms) in the adaptive learning-rate tuning and choose (\pref{line:strong-eta})
\begin{equation}
  \eta_t=\sqrt{\frac{\ln(sT)}{2A\left(1+\sum_{\tau=1}^{t-1}\frac{\widetilde g_{\tau,0}}{x_{\tau,0}}\right)}}.
  \label{eq:strong-base-learning-rate}
\end{equation}
It remains to see how to compute $q_t$ efficiently. The map \pref{eq:strong-base-map} is the smoothed map \pref{eq:u-map} with $\eta$ replaced by $\eta_t/x_{t,0}$, and the offset $1/(KT)$ in $\psi_t$ plays the same smoothing role as in \pref{prop:u-contraction}. However, \pref{prop:u-contraction} requires its parameter $\eta$ to lie in $[1/(KT),1]$, while $\eta_t/x_{t,0}$ grows without bound as $x_{t,0}$ approaches zero. 
To address this,
we first extend \pref{prop:u-contraction} to an arbitrary parameter $\gamma>0$ and provide a bound on the number of Picard iterations that is polynomial whenever $\gamma KT$ and $1/(\gamma KT)$ are polynomially bounded; see \pref{prop:u-arbitrary-iterations} in \pref{app:computation}.
We therefore need to control $\gamma = \eta_t/x_{t,0}$ from both above and below.
For this reason, 
we restrict the master to the clipped domain $\mathcal X\triangleq\{x\in\Delta(\Bcal):x_b\ge1/(2MT)\text{ for all }b\in\Bcal\}$, so that $x_{t,0}\ge1/(2MT)$. This bound controls $\eta_t/x_{t,0}$ from above (since $\eta_t\le\sqrt{\ln(sT)/(2A)}$) and at the same time, through $\widetilde g_{\tau,0}/x_{\tau,0}\le1/x_{\tau,0}^2$ in \pref{eq:strong-base-learning-rate}, bounds $\eta_t$ from below.
Together, this makes sure that $q_t$ can be computed in polynomial time.

\begin{algorithm2e}[t]
\caption{The self-aware learner with a complement-barrier master}
\label{alg:strong}
\KwIn{strongly observable graph $G$, horizon $T$, bound $A\ge1$, master rate $\theta>0$}
Set $J=\lceil\log_2(2MT)\rceil$, $\sigma=e^{1/J}$, $\rho_1=2$ and $\eps=1/(KT^2)$\;
Define $\mathcal P\triangleq\{p\in\Delta(\Vself):p_i\ge1/(sT)\text{ for all }i\in\Vself\}$ and $\mathcal X\triangleq\{x\in\Delta(\Bcal):x_b\ge1/(2MT)\text{ for all }b\in\Bcal\}$\;
Initialize $p_1$ uniformly on $\Vself$, $x_{1,0}=(M+1)/(2M)$, $x_{1,i}=1/(2M)$ for $i\in\Vbarself$, and $\theta_{1,0}=\theta$\;
\For{$t=1,\ldots,T$}{
  Set $\eta_t=\sqrt{\frac{\ln(sT)}{2A(1+\sum_{\tau=1}^{t-1}\frac{\widetilde g_{\tau,0}}{x_{\tau,0}})}}$ and $\psi_t(v)=\frac{v+1/(KT)}{v+\eta_t/x_{t,0}}$\label{line:strong-eta}\;
  Compute an interior $\eps$-fixed point $q_t$ of $q\mapsto\left(\frac{p_{t,i}\psi_t(Q_i^{\Gself}(q))}{\sum_{j\in\Vself}p_{t,j}\psi_t(Q_j^{\Gself}(q))}\right)_{i\in\Vself}$ by Picard iteration\label{line:strong-fixed-point}\;
  Set $\pi_{t,i}=x_{t,0}q_{t,i}$ for $i\in\Vself$ and $\pi_{t,i}=x_{t,i}$ for $i\in\Vbarself$; draw $B_t\sim x_t$\label{line:strong-sample}\;
  \lIf{$B_t=0$}{draw $I_t\sim q_t$}
  \lElse{set $I_t=B_t$}
  Play $I_t$ and observe the graph feedback\;
  Set $\widetilde\ell_{t,i}=\ell_{t,i}\ind\{I_t\to i\}/Q_i(\pi_t)$ for every $i\in\Vself$\label{line:strong-base-est}\;
  Set $\widetilde g_{t,0}=\ell_{t,I_t}\ind\{B_t=0\}/x_{t,0}$ and $\widetilde g_{t,i}=\ell_{t,i}\ind\{I_t\ne i\}/(1-x_{t,i})$ for $i\in\Vbarself$\label{line:strong-master-est}\;
  Update base learner: $p_{t+1}=\argmin_{p\in\mathcal P}\left\{\ip{p}{\widetilde\ell_t}+\frac{1}{\eta_t}D_h(p,p_t)\right\}$\label{line:strong-p-update}\;
  Update master learner: $x_{t+1}=\argmin_{x\in\mathcal X}\{\ip{x}{\widetilde g_t}+D_{\Psi_t}(x,x_t)\}$, with $\Psi_t$ given by \pref{eq:strong-regularizer}\label{line:strong-x-update}\;
  \lIf{$1/x_{t+1,0}>\rho_t$}{
    set $\rho_{t+1}=2/x_{t+1,0}$ and $\theta_{t+1,0}=\sigma\theta_{t,0}$\label{line:strong-threshold}
  }
  \lElse{
    set $\rho_{t+1}=\rho_t$ and $\theta_{t+1,0}=\theta_{t,0}$
  }
}
\end{algorithm2e}

\paragraph{Master learner.}
For the master learner, we update $x_t$ by yet another OMD (\pref{line:strong-x-update}). Specifically, we use the following hybrid regularizer at round $t$:
\begin{equation}
  \Psi_t(x)\triangleq\frac{1}{\theta_{t,0}}\ln\frac{1}{x_0}
  +\frac{1}{\theta}\sum_{i\in\Vbarself}\ln\frac{1}{1-x_i}
  +\frac{1}{\theta}\sum_{i\in\Vbarself}x_i\ln x_i,
  \label{eq:strong-regularizer}
\end{equation}
where $\theta>0$ is the learning rate for the actions in $\Vbarself$, and $\theta_{t,0}$, initialized at $\theta_{1,0}=\theta$, is the learning rate for meta-action $0$.
The first and third terms of this hybrid regularizer follow Eq.~(7) of \citet{lee2020graph}: a log barrier on meta-action $0$ and entropy on $\Vbarself$. The second term handles the actions without self-loops. For $i\in\Vbarself$, the estimator in \pref{eq:strong-master-estimator} leads to $\Et[\widetilde g_{t,i}^2]=\ell_{t,i}^2/(1-x_{t,i})$. With entropy alone, the OMD regret bound contains the term $\theta x_{t,i}\Et[\widetilde g_{t,i}^2]=\theta x_{t,i}\ell_{t,i}^2/(1-x_{t,i})$ for this coordinate, which is unbounded as $x_{t,i}$ approaches $1$. \citet{lee2020graph} addresses this by a ``loss shifting'' technique and adding the barrier $c\sum_{i\in\Vbarself}\ln\frac1{x_i}$ with a constant coefficient $c$ of order $M$, which adds an $\order(M^2\ln T)$ term to the regret. We instead add the complement barrier $\frac1\theta\sum_{i\in\Vbarself}\ln\frac{1}{1-x_i}$, whose curvature $\frac{1}{\theta(1-x_i)^2}$ offsets the factor $\frac1{1-x_{t,i}}$ in the second moment $\Et[\widetilde g_{t,i}^2]$. Together with the entropy, this barrier replaces $x_{t,i}$ by $\min\{x_{t,i},(1-x_{t,i})^2\}$ in the regret bound, which is then bounded directly without loss shifting:
\begin{align*}
  \theta\min\{x_{t,i},(1-x_{t,i})^2\}\Et[\widetilde g_{t,i}^2]
  &=\theta\min\left\{\frac{x_{t,i}}{1-x_{t,i}},1-x_{t,i}\right\}\ell_{t,i}^2
    \le2\theta x_{t,i}\ell_{t,i},
\end{align*}
where the inequality uses the entropy-related term when $x_{t,i}\le1/2$ and the barrier-related term otherwise, together with $\ell_{t,i}\le1$.
Moreover, on the clipped domain $\mathcal X$, the complement barrier contributes at most $(3+\ln(2MT))/\theta$ to the initial Bregman divergence (as shown in \pref{lem:strong-barrier-reward}), which is of the same order as the barrier on meta-action $0$. Therefore, the complement barrier removes the $\order(M^2\ln T)$ term of \citet{lee2020graph}.
When $0\in\Bcal$, we initialize $x_{1,0}=(M+1)/(2M)$ and $x_{1,i}=1/(2M)$ for $i\in\Vbarself$; otherwise, we initialize $x_1$ uniformly on $\Bcal$. If $M=1$, the master always selects the self-aware learner.

It remains to choose the learning rate $\theta_{t,0}$ for meta-action $0$. The base learner's estimates scale with $1/x_{t,0}$, so its regret can be large when $x_{t,0}$ becomes small (\textit{c.f.} \pref{eq:base_stab}). As in \citet{lee2020graph}, we handle this with the increasing learning-rate scheme of CORRAL \citep{agarwal2017corralling}. 
Specifically, the master maintains an inverse-probability threshold $\rho_t$ with $\rho_1=2$. Set $J=\lceil\log_2(2MT)\rceil$ and $\sigma=e^{1/J}$. If $1/x_{t+1,0}>\rho_t$, we set $\rho_{t+1}=2/x_{t+1,0}$ and $\theta_{t+1,0}=\sigma\theta_{t,0}$ (\pref{line:strong-threshold}). Otherwise, both remain unchanged. This increasing learning rate serves a technical purpose, as shown in \citet{agarwal2017corralling}: each increase of $\theta_{t,0}$ decreases the coefficient $1/\theta_{t,0}$ of the barrier on meta-action $0$, which produces the negative term that controls the base learner's regret dependence on $1/x_{t,0}$.
This concludes all details of our algorithm; see \pref{alg:strong}.

\subsection{Regret Guarantee}
\label{sec:strong-oracle}

We first split the regret between the master and the self-aware learner. By \pref{eq:strong-play-distribution}, the expected loss at round $t$ is $\Et[\ell_{t,I_t}]=\ip{x_t}{g_t}$, which is the master's loss under $x_t$.
For $b\in\Bcal$, denote by $\Reg^{\mathrm{master}}_{T,b}\triangleq\E[\sum_{t=1}^T(\ip{x_t}{g_t}-g_{t,b})]$ the master's expected regret against $b$, and for $k\in \Vself$, denote by $\Reg^{\mathrm{base}}_{T,k}\triangleq\E[\sum_{t=1}^T(g_{t,0}-\ell_{t,k})]$ the self-aware learner's expected regret against $k$.
Adding and subtracting $g_{t,0}$ for $k\in\Vself$, and using $g_{t,i}=\ell_{t,i}$ for $i\in\Vbarself$, we obtain that
\begin{align}
  \Reg_{T,k}
  &=\Reg^{\mathrm{master}}_{T,0}+\Reg^{\mathrm{base}}_{T,k},
    && k\in \Vself,
  \label{eq:block-decomp-S}\\
  \Reg_{T,i}
  &=\Reg^{\mathrm{master}}_{T,i},
    && i\in\Vbarself.
  \label{eq:block-decomp-U}
\end{align}

It remains to bound the two regrets. Following Lemma~11 and Lemma~12 of \citet{lee2020graph}, we bound them in the following lemma.

\begin{lemma}
\label[lemma]{lem:strong-components}
Suppose $\theta\le1/(8e)$. When $0\in\Bcal$, assume $\mathcal H^{\Gself}(q_t)\le A$ for all $t\in[T]$ and denote by $\bar\rho\triangleq\max_{1\le t\le T}1/x_{t,0}$ the largest inverse probability of meta-action $0$; otherwise, set $A=1$ and $\bar\rho=0$. Then the base bound below holds for every $k\in\Vself$, and the master bound holds for every $b\in\Bcal$:
\begin{align}
  \Reg^{\mathrm{base}}_{T,k}
  &\le74A\ln(KT)\E[\bar\rho]+8\sqrt{2A\ln(KT)\E[\bar\rho]L_{T,k}},
  \label{eq:strong-base-ledger}\\
  \Reg^{\mathrm{master}}_{T,b}
  &\le\frac{9\ln(MT)}{\theta}
    +12\theta\E\left[\sum_{t=1}^{T}\ip{x_t}{g_t}\right]
    -\frac{\E[\bar\rho]}{16e^2\theta J}\ind\{b=0\}.
  \label{eq:strong-master-ledger}
\end{align}
\end{lemma}

The proof is in \pref{app:strong-structure} and \pref{app:strong-master}. The base bound grows with $\E[\bar\rho]$ because of the $1/x_{t,0}$ factor in \pref{eq:base_stab}, while the increasing learning rate of meta-action $0$ gives the master bound the negative term $-\E[\bar\rho]/(16e^2\theta J)$. Denote by $\Gamma_A\triangleq AJ\ln(KT)$. When $\theta\le c/\Gamma_A$ for a sufficiently small universal constant $c>0$, this negative term cancels both terms involving $\E[\bar\rho]$ in \pref{eq:strong-base-ledger}, up to an additional $\order(\Gamma_A\theta L_{T,k})$. Consequently, we obtain the following guarantee.

\begin{theorem}
\label{thm:strong-fixed}
Let $A\ge1$ satisfy $\mathcal H^{\Gself}(q_t)\le A$ for all $t\in[T]$ when $0\in\Bcal$; otherwise, take $A=1$. For a sufficiently small universal constant $c>0$, choose $0<\theta\le c/\Gamma_A$. Then \pref{alg:strong} guarantees, for every $k\in V$,
\begin{equation*}
  \Reg_{T,k}\le\order\left(\frac{\ln(MT)}{\theta}+\Gamma_A\theta L_{T,k}\right).
\end{equation*}
Therefore, setting $\theta=\min\{c/\Gamma_A,\sqrt{\ln(MT)/(\Gamma_A L^\star)}\}$ and $A=\order(\alpha\ln(KT))$ from \pref{lem:soft-ratio} applied to $\Gself$ gives $\Reg_T=\otil(\sqrt{\alpha L^\star}+\alpha)$.
\end{theorem}

The proof is in \pref{app:strong-master}.
Similar to \pref{thm:u-master}, \pref{thm:strong-fixed} requires knowing $\alpha$ and $L^\star$. To remove this requirement, we follow the doubling trick of \pref{alg:u-adaptive} in \pref{app:u-adaptive}, using estimated losses as in \citet{lee2020graph}. Before sampling, we check $\mathcal H^{\Gself}(q_t)\le A$ and double $A$ if the check fails. For a fixed $A$, we halve $\theta$ when the accumulated observable loss, with per-round value $\widetilde Z_t\triangleq\ip{x_t}{\widetilde g_t}$, reaches a threshold. The pseudocode and the proof are given in \pref{app:strong-adaptive}.

\begin{theorem}
\label{thm:strong-adaptive}
\pref{alg:strong-adaptive} (in \pref{app:strong-adaptive}) guarantees that
\begin{equation}
  \Reg_T=\otil\left(\sqrt{\alpha L^\star}+\alpha\right),
  \label{eq:strong-adaptive-rate}
\end{equation}
for any strongly observable graph without knowing $L^\star$ and $\alpha$.
\end{theorem}

\section{Conclusion}
\label{sec:discussion}

In this paper, we obtain expected regret $\otil(\sqrt{\alpha L^\star}+\alpha)$ for adversarial bandits with a fixed, known directed strongly observable feedback graph, without knowing $L^\star$ or $\alpha$. This improves the clique-partition dependence of previous first-order bounds to the independence number. The main ingredient is a fixed-point sampling rule that controls the loss-estimation cost by the learner's loss while using the standard EXP3 update with unbiased estimates. Reducing the logarithmic factors in our regret bounds is left for future work.

\clearpage
\appendix
\section{Omitted Details in \texorpdfstring{\pref{sec:uncorrected}}{Section 3}}
\label{app:self-aware}

In this section, we provide the omitted details of \pref{sec:uncorrected}. We first prove \pref{thm:u-master}, following the regret decomposition and fixed-point calculation in the main text. We then present a doubling trick that removes the knowledge of $\alpha$ and $L^\star$, prove \pref{prop:u-contraction} together with its extension used in \pref{sec:strongly-observable}, and prove the slightly refined analysis of EXP3 in \pref{prop:ordinary-potential}.

\subsection{Proof of \texorpdfstring{\pref{thm:u-master}}{Theorem 3.2}}
\label{app:u-proof}

We first prove \pref{eq:u-bridge-constants} for an approximate fixed point, restated in the following proposition.

\begin{proposition}
\label[proposition]{prop:u-approx-fixed-point}
Let $G$ be self-aware, $\eta\ge1/(KT)$, $p\in\operatorname{int}(\DeltaK)$ and $\eps>0$. Then every $q\in\DeltaK$ with $\norm{q-\Fu_p(q)}_1\le\eps$ satisfies, for every $\ell_t\in[0,1]^K$,
\begin{equation*}
\REM_t(p,q)
\le\eta\left(\mathcal H^G(q)+KT\eps\right)\ip{q}{\ell_t}
+\frac{\mathcal H^G(q)}{KT}+2\eps.
\tag{\ref{eq:u-bridge-constants}}
\end{equation*}
\end{proposition}

\begin{proof}
Denote by $\xi\triangleq(1-Z(p,q))(q-\Fu_p(q))$ the residual of the unnormalized fixed-point equation.
Since $1-Z(p,q)\le1$, we have $\norm{\xi}_1\le\eps$, and \pref{eq:u-map} gives
\begin{equation}
  (1-Z(p,q))q_j-\xi_j
  =p_j\psi(Q_j(q))
  =\frac{p_j(Q_j(q)+1/(KT))}{Q_j(q)+\eta}.
  \label{eq:u-coordinate-identity}
\end{equation}
Dividing \pref{eq:u-coordinate-identity} by $Q_j(q)+1/(KT)$ and summing over $j$ yields
\begin{equation}
  \sum_{j=1}^K\frac{p_j}{Q_j(q)+\eta}
  \le\sum_{j=1}^K\frac{q_j+|\xi_j|}{Q_j(q)+1/(KT)}
  \le\mathcal H^G(q)+KT\eps,
  \label{eq:u-estimation-ratio}
\end{equation}
where the first inequality uses $1-Z(p,q)\le1$, and the second uses $\norm{\xi}_1\le\eps$.
Consequently, the definition of $Z(p,q)$ gives
\begin{equation}
  Z(p,q)
  =\left(\eta-\frac1{KT}\right)\sum_{j=1}^K\frac{p_j}{Q_j(q)+\eta}
  \le\eta\left(\mathcal H^G(q)+KT\eps\right).
  \label{eq:z-bound}
\end{equation}
Using $q_j-p_j\psi(Q_j(q))=Z(p,q)q_j+\xi_j$ from \pref{eq:u-coordinate-identity}, the calculation in \pref{eq:u-target-bridge-exact} becomes
\begin{align}
  \REM_t(p,q)
  &=Z(p,q)\ip{q}{\ell_t}+\ip{\xi}{\ell_t}
    +\frac1{KT}\sum_{j=1}^K\frac{p_j\ell_{t,j}}{Q_j(q)+\eta}\nonumber\\*
  &\le Z(p,q)\ip{q}{\ell_t}
    +\frac{\mathcal H^G(q)}{KT}+2\eps,
  \label{eq:reg-replace}
\end{align}
where the inequality uses \pref{eq:u-estimation-ratio}, $\ell_t\in[0,1]^K$ and $\norm{\xi}_1\le\eps$.
Combining \pref{eq:z-bound} and \pref{eq:reg-replace} proves \pref{eq:u-bridge-constants}.
\end{proof}

We are now ready to prove \pref{thm:u-master} by combining \pref{prop:ordinary-potential} with \pref{prop:u-approx-fixed-point}.

\begin{proof}
Recall that \pref{alg:uncorrected} samples $I_t\sim q_t$ and uses $\ellhat_{t,j}=\ell_{t,j}\ind\{I_t\to j\}/Q_j(q_t)$ in the EXP3 update \pref{eq:u-omd-update}.
Applying \pref{prop:ordinary-potential} gives, for every $i\in[K]$,
\begin{equation*}
  \sum_{t=1}^T\ip{p_t-e_i}{\ellhat_t}
  \le\frac{\ln K}{\eta}
    +\eta\sum_{t=1}^T\sum_{j=1}^K\frac{p_{t,j}\ellhat_{t,j}^2}{1+\eta\ellhat_{t,j}}.
\end{equation*}
Since $q_t$ is an interior distribution and $G$ is self-aware, $Q_j(q_t)>0$. Thus the conditional expectation satisfies $\Et[\ellhat_{t,j}]=\ell_{t,j}$ and
\begin{equation*}
  \Et\left[\frac{\eta\ellhat_{t,j}^2}{1+\eta\ellhat_{t,j}}\right]
  =\frac{\eta\ell_{t,j}^2}{Q_j(q_t)+\eta\ell_{t,j}}
  \le\frac{\eta\ell_{t,j}}{Q_j(q_t)+\eta},
\end{equation*}
where the inequality uses $\ell_{t,j}\le1$.
Taking expectations in the bound of \pref{prop:ordinary-potential} and using this calculation gives
\begin{align}
  \Reg_{T,i}
  &\le\frac{\ln K}{\eta}
    +\E\left[\sum_{t=1}^T\REM_t(p_t,q_t)\right]\nonumber\\*
  &=\frac{\ln K}{\eta}
    +\E\left[\sum_{t=1}^T\sum_{j=1}^K\ell_{t,j}
      \left(q_{t,j}-p_{t,j}\psi(Q_j(q_t))
        +\frac{p_{t,j}}{KT(Q_j(q_t)+\eta)}\right)\right],
  \label{eq:reg_fix}
\end{align}
where the equality uses $\phi(s)=s/(s+\eta)$ and $\psi(s)=(s+1/(KT))/(s+\eta)$ from \pref{sec:uncorrected}.
Substituting \pref{eq:u-bridge-constants} into \pref{eq:reg_fix} and using $\eps=1/(KT^2)$, $\mathcal H^G(q_t)\le A$ in the learner-loss term and $\mathcal H^G(q_t)\le K$ in the additive term gives
\begin{align*}
  \Reg_{T,i}
  &\le\frac{\ln K}{\eta}
    +\eta\left(A+\frac1T\right)\E\left[\sum_{t=1}^T\ip{q_t}{\ell_t}\right]
    +1+\frac2{KT}\\*
  &=\frac{\ln K}{\eta}
    +\eta\left(A+\frac1T\right)(L_{T,i}+\Reg_{T,i})
    +1+\frac2{KT}.
\end{align*}
Since $\eta\le1/(12A)$, rearranging the terms gives
\begin{equation*}
  \Reg_{T,i}
  \le\frac{\frac{\ln K}{\eta}+\eta(A+1/T)L_{T,i}+1+2/(KT)}{1-\eta(A+1/T)}
  \le\frac{3\ln K}{\eta}+10A\eta L_{T,i},
\end{equation*}
where the last inequality uses $A\ge1$, $T>2$ and $K\ge2$. This proves \pref{eq:u-master}.

It remains to substitute the learning rate in \pref{thm:u-master}. If $1/(12A)<1/(KT)$, then bounded losses give $\Reg_T\leq T<12A/K$, which proves the stated rate.
Otherwise, $1/(KT)\le\eta\le1/(12A)$, and the choice of $\eta$ gives $1/\eta\le\max\{12A,\sqrt{AL^\star/\ln K}\}$ and $A\eta L^\star\le\sqrt{AL^\star\ln K}+A$. Substituting these bounds into \pref{eq:u-master} with $i$ being the best action gives
\begin{equation*}
  \Reg_T
  \le\frac{3\ln K}{\eta}+10A\eta L^\star
  \le36A\ln K+13\sqrt{AL^\star\ln K}+10A
  =\order\left(\sqrt{AL^\star\ln K}+A\ln K\right).
\end{equation*}
Taking $A=\order(\alpha\ln(KT))$ by \pref{lem:soft-ratio} gives $\Reg_T=\otil(\sqrt{\alpha L^\star}+\alpha)$.
\end{proof}

\subsection{Doubling Trick}
\label{app:u-adaptive}
In this section, we show a parameter-free version of \pref{alg:uncorrected} that achieves $\otil(\sqrt{\alpha L^\star}+\alpha)$ regret without knowing $\alpha$ or $L^\star$, which are either unknown in advance or NP-hard to compute. Specifically, we propose \pref{alg:u-adaptive}, a doubling-trick variant of \pref{alg:uncorrected} which maintains a guess for the bound $A$ in \pref{thm:u-master} and the cumulative incurred loss $S$ since the last restart. Each instance of \pref{alg:uncorrected} defines an epoch with a fixed guess and learning rate. Consecutive epochs with the same guess form a meta-epoch. Denote by $m$ the meta-epoch index, by $\lambda$ the epoch index within it, by $A_m$ the guess in meta-epoch $m$, and by $\eta_{m,\lambda}$ the learning rate in epoch $(m,\lambda)$. Once $\mathcal H^G(q_t)>A_m$, we double the guess and start meta-epoch $m+1$ at $\lambda=1$ with its initial learning rate. Note that a meta-epoch may contain no rounds, which happens when $\mathcal H^G(q_t)>A_{m+1}$ still holds right after a doubling. Once $A_m\eta_{m,\lambda}^2S\ge\ln K$, we halve the learning rate, subject to the floor $1/(KT)$, and start epoch $\lambda+1$ within meta-epoch $m$. At the start of each epoch, we initialize \pref{alg:uncorrected} with $\eta_{m,\lambda}$ and set $S=0$. The complete pseudocode is shown in \pref{alg:u-adaptive}, and \pref{thm:u-adaptive} establishes the claimed regret guarantee.
\begin{algorithm2e}[t]
\caption{Doubling trick with \pref{alg:uncorrected} as an oracle}
\label{alg:u-adaptive}
\KwIn{self-aware graph $G$, horizon $T$}
Set $m=1$, $\lambda=1$, $A_1=1$, $\eta_{1,1}=\max\{\frac1{KT},\frac1{12A_1}\}$ and $S=0$\;
Start meta-epoch $m$ and epoch $(m,\lambda)$ by initializing an instance of \pref{alg:uncorrected} with $(T,\eta_{m,\lambda})$\;
\For{$t=1,\ldots,T$}{
  Compute $q_t$ by the instance of \pref{alg:uncorrected} for epoch $(m,\lambda)$\;
  \While{$\mathcal H^G(q_t)>A_m$\label{line:u-adaptive-check}}{
    $A_{m+1}=2A_m$, $\eta_{m+1,1}=\max\{\frac1{KT},\frac1{12A_{m+1}}\}$, $m\leftarrow m+1$, $\lambda\leftarrow1$ and $S\leftarrow0$\;
    Start meta-epoch $m$ and epoch $(m,1)$ by initializing a new instance of \pref{alg:uncorrected} with $(T,\eta_{m,1})$, starting at round $t$\;
    Compute $q_t$ by this new instance\;
  }
  Execute round $t$ of the instance of \pref{alg:uncorrected} for epoch $(m,\lambda)$, and denote by $I_t$ its sampled action\;
  $S\leftarrow S+\ell_{t,I_t}$\;
  \If{$t<T$ and $A_m\eta_{m,\lambda}^2S\ge\ln K$\label{line:u-adaptive-halve}}{
    $\eta_{m,\lambda+1}=\max\{\frac1{KT},\frac{\eta_{m,\lambda}}2\}$, $\lambda\leftarrow\lambda+1$ and $S\leftarrow0$\;
    Start epoch $(m,\lambda)$ by initializing a new instance of \pref{alg:uncorrected} with $(T,\eta_{m,\lambda})$, starting at round $t+1$\;
  }
}
\end{algorithm2e}

\begin{theorem}
\label{thm:u-adaptive}
Fix a self-aware graph $G$. Then \pref{alg:u-adaptive} guarantees that, for every $i\in[K]$,
\begin{equation}
  \Reg_{T,i}
  =\otil\left(\sqrt{\alpha L_{T,i}}+\alpha\right),
  \label{eq:u-adaptive}
\end{equation}
without knowing $\alpha$ or $L^\star$. In particular, $\Reg_T=\otil(\sqrt{\alpha L^\star}+\alpha)$.
\end{theorem}

\begin{proof}
Denote by $m^\star$ the index of the last meta-epoch started by \pref{alg:u-adaptive} and by $\lambda_m^\star$ the index of the last epoch started within meta-epoch $m$. We sum over these epochs before taking expectations. Denote by $A\triangleq2\alpha\ln(1+(K^3T+K)/\alpha)+2$ the bound in \pref{lem:soft-ratio}, so that $\mathcal H^G(q)\le A$ for every $q\in\DeltaK$.
Every reached meta-epoch $m$ uses the guess $A_m=2^{m-1}\le2K$, since self-awareness gives $\mathcal H^G(q)\le K$.
Since the condition $\mathcal H^G(q_t)>A_m$ in \pref{line:u-adaptive-check} never holds once $A_m\ge A$, and $A_{m+1}=2A_m$ with $A_1=1$, we have $A_{m^\star}<2A$ and $\sum_{m=1}^{m^\star}A_m<4A$.
Denote by $\mathcal T_m$ the rounds of meta-epoch $m$ and by $S_m\triangleq\sum_{t\in\mathcal T_m}\ell_{t,I_t}$ its incurred loss.
Denote by $\mathcal T_{m,\lambda}$ the rounds of epoch $(m,\lambda)$ and by $S_{m,\lambda}\triangleq\sum_{t\in\mathcal T_{m,\lambda}}\ell_{t,I_t}$ its incurred loss. For each $i\in[K]$, denote the expected regret in epoch $(m,\lambda)$ by
\begin{equation*}
  \Reg_{m,\lambda,i}
  \triangleq\E\left[\sum_{t\in\mathcal T_{m,\lambda}}(\ell_{t,I_t}-\ell_{t,i})\right].
\end{equation*}
To bound the total regret, we apply \pref{prop:ordinary-potential} in each epoch $(m,\lambda)$ to obtain
\begin{equation*}
  \sum_{t\in\mathcal T_{m,\lambda}}\left(
    \ip{p_t-e_i}{\ellhat_t}
    -\eta_{m,\lambda}\sum_{j=1}^K\frac{p_{t,j}\ellhat_{t,j}^2}{1+\eta_{m,\lambda}\ellhat_{t,j}}\right)
  \le\frac{\ln K}{\eta_{m,\lambda}}.
\end{equation*}
Summing over the epochs started by the algorithm gives the pathwise bound
\begin{equation}
  \sum_{m=1}^{m^\star}\sum_{\lambda=1}^{\lambda_m^\star}
    \sum_{t\in\mathcal T_{m,\lambda}}\left(
      \ip{p_t-e_i}{\ellhat_t}
      -\eta_{m,\lambda}\sum_{j=1}^K\frac{p_{t,j}\ellhat_{t,j}^2}{1+\eta_{m,\lambda}\ellhat_{t,j}}\right)
  \le\sum_{m=1}^{m^\star}\sum_{\lambda=1}^{\lambda_m^\star}
    \frac{\ln K}{\eta_{m,\lambda}}.
  \label{eq:u-adaptive-potential}
\end{equation}
To convert this bound into a bound on the true regret, denote by $(m_t,\lambda_t)$ the epoch containing round $t$. These indices, $p_t,q_t$ and $\eta_{m_t,\lambda_t}$ are determined before drawing $I_t$. Recall from \pref{sec:prelim} that $\Et$ conditions on $\cF_{t-1}$. Evaluating $\REM_t$ with the current learning rate gives
\begin{align}
  \Et[\ell_{t,I_t}-\ell_{t,i}]
  &\le\Et\left[\ip{p_t-e_i}{\ellhat_t}
    -\eta_{m_t,\lambda_t}\sum_{j=1}^K
    \frac{p_{t,j}\ellhat_{t,j}^2}{1+\eta_{m_t,\lambda_t}\ellhat_{t,j}}\right]+\REM_t(p_t,q_t),
  \label{eq:u-adaptive-conditional}
\end{align}
where the inequality uses the unbiasedness of \pref{eq:u-unbiased-estimator} and the conditional-moment calculation in \pref{eq:u-exponential-moment}, with $Q_j(q_t)$ in place of $Q_j(p_t)$. Taking outer expectations and summing over $t=1,\ldots,T$ therefore gives
\begin{align}
  \Reg_{T,i}
  &\le\E\left[\sum_{t=1}^T\left(
    \ip{p_t-e_i}{\ellhat_t}
    -\eta_{m_t,\lambda_t}\sum_{j=1}^K\frac{p_{t,j}\ellhat_{t,j}^2}{1+\eta_{m_t,\lambda_t}\ellhat_{t,j}}
    +\REM_t(p_t,q_t)\right)\right]\nonumber\\*
  &\le\E\left[\sum_{m=1}^{m^\star}\sum_{\lambda=1}^{\lambda_m^\star}
    \frac{\ln K}{\eta_{m,\lambda}}+\sum_{t=1}^T\REM_t(p_t,q_t)\right],
  \label{eq:u-adaptive-reg-decomposition}
\end{align}
where the first inequality uses \pref{eq:u-adaptive-conditional} and the tower property, and the second uses \pref{eq:u-adaptive-potential}.
To bound the remainder term in \pref{eq:u-adaptive-reg-decomposition}, apply \pref{eq:u-bridge-constants} on a played round $t\in\mathcal T_{m,\lambda}$ to obtain
\begin{equation}
  \REM_t(p_t,q_t)
  \le\eta_{m,\lambda}\left(\mathcal H^G(q_t)+KT\eps\right)\ip{q_t}{\ell_t}
    +\frac{\mathcal H^G(q_t)}{KT}+2\eps.
  \label{eq:u-adaptive-remainder}
\end{equation}
Plugging \pref{eq:u-adaptive-remainder} into \pref{eq:u-adaptive-reg-decomposition} gives
\begin{align}
  \Reg_{T,i}
  &\le\E\left[\sum_{m=1}^{m^\star}\sum_{\lambda=1}^{\lambda_m^\star}\frac{\ln K}{\eta_{m,\lambda}}\right]
    +\E\left[\sum_{t=1}^T\eta_{m_t,\lambda_t}\left(A_{m_t}+\frac1T\right)\ip{q_t}{\ell_t}\right]+1+\frac2{KT},
  \label{eq:u-adaptive-weighted-regret}
\end{align}
where we use $\eps=1/(KT^2)$, $\mathcal H^G(q_t)\le A_{m_t}$ by \pref{line:u-adaptive-check} in the loss-dependent term, and $\mathcal H^G(q_t)\le K$ from self-awareness in the additive term.
To bound the second summation in \pref{eq:u-adaptive-weighted-regret}, we use $\Et[\ell_{t,I_t}]=\ip{q_t}{\ell_t}$ over the full horizon to obtain
\begin{align}
  \E\left[\sum_{t=1}^T\eta_{m_t,\lambda_t}\left(A_{m_t}+\frac1T\right)\ip{q_t}{\ell_t}\right]
  &=\E\left[\sum_{t=1}^T\Et\left[\eta_{m_t,\lambda_t}\left(A_{m_t}+\frac1T\right)\ell_{t,I_t}\right]\right]\nonumber\\*
  &=\E\left[\sum_{t=1}^T\eta_{m_t,\lambda_t}\left(A_{m_t}+\frac1T\right)\ell_{t,I_t}\right]\nonumber\\*
  &=\E\left[\sum_{m=1}^{m^\star}\sum_{\lambda=1}^{\lambda_m^\star}
    \eta_{m,\lambda}\left(A_m+\frac1T\right)S_{m,\lambda}\right],
  \label{eq:u-adaptive-loss-conversion}
\end{align}
where the first equality uses that the coefficient is $\cF_{t-1}$-measurable, the second is the tower property, and the last groups the realized rounds by epoch. Plugging \pref{eq:u-adaptive-loss-conversion} into \pref{eq:u-adaptive-weighted-regret} gives
\begin{equation}
  \Reg_{T,i}
  \le\E\left[\sum_{m=1}^{m^\star}\sum_{\lambda=1}^{\lambda_m^\star}
    \left(\frac{\ln K}{\eta_{m,\lambda}}+\eta_{m,\lambda}\left(A_m+\frac1T\right)S_{m,\lambda}\right)\right]
    +2.
  \label{eq:u-adaptive-epoch-bound}
\end{equation}
To bound $S_{m,\lambda}$ in \pref{eq:u-adaptive-epoch-bound}, denote by $\tend_{m,\lambda}$ the last round of a nonempty epoch $(m,\lambda)$. Separating this round from the preceding rounds gives
\begin{align}
  S_{m,\lambda}
  &\le\sum_{\substack{t\in\mathcal T_{m,\lambda}\\t<\tend_{m,\lambda}}}\ell_{t,I_t}+1\le\frac{\ln K}{A_m\eta_{m,\lambda}^2}+1,
  \label{eq:u-adaptive-epoch-loss}
\end{align}
where the first inequality bounds the last incurred loss by one, and the second uses the restart condition before that round.
Substituting \pref{eq:u-adaptive-epoch-loss} and the learning-rate schedule into \pref{eq:u-adaptive-epoch-bound}, using $A_m\le2K$, yields
\begin{equation}
  \Reg_{T,i}
  \le\order\left(\E\left[\sum_{m=1}^{m^\star}\sum_{\lambda=1}^{\lambda_m^\star}\frac{\ln K}{\eta_{m,\lambda}}\right]\right).
  \label{eq:u-adaptive-rate-bound}
\end{equation}

It remains to bound the sum of inverse learning rates in \pref{eq:u-adaptive-rate-bound}. Fix a meta-epoch $m\in[m^\star]$. By \pref{alg:u-adaptive}, $\eta_{m,\lambda}=\max\{\frac1{KT},\frac{2^{1-\lambda}}{12A_m}\}$. Once $\eta_{m,\lambda}=\frac1{KT}$, the condition in \pref{line:u-adaptive-halve} never holds again, since $A_m\eta_{m,\lambda}^2S\le2K\cdot\frac{T}{(KT)^2}=\frac2{KT}<\ln K$ by $A_m\le2K$ and $S\le T$. Hence only the last epoch of meta-epoch $m$ can be at the floor, the inverse rates of the other epochs double from one epoch to the next, and
\begin{align}
\sum_{\lambda=1}^{\lambda_m^\star}\frac1{\eta_{m,\lambda}}\le\frac3{\eta_{m,\lambda_m^\star}}.
\label{eq:u-adaptive-geometric}
\end{align}
To bound $\frac1{\eta_{m,\lambda_m^\star}}$ in \pref{eq:u-adaptive-geometric}, we consider two cases. If $\lambda_m^\star=1$, then $\frac1{\eta_{m,1}}\le12A_m$. If $\lambda_m^\star\ge2$, then epoch $(m,\lambda_m^\star-1)$ ended because the condition in \pref{line:u-adaptive-halve} held, that is, $A_m\eta_{m,\lambda_m^\star-1}^2S_{m,\lambda_m^\star-1}\ge\ln K$, and $S_{m,\lambda_m^\star-1}\le S_m$. Therefore,
\begin{align*}
\frac1{\eta_{m,\lambda_m^\star}}\le\frac2{\eta_{m,\lambda_m^\star-1}}\le2\sqrt{\frac{A_mS_m}{\ln K}}.
\end{align*}
In both cases, $\frac1{\eta_{m,\lambda_m^\star}}\le12A_m+2\sqrt{\frac{A_mS_m}{\ln K}}$. Plugging this bound into \pref{eq:u-adaptive-geometric} and taking the summation over $m\in[m^\star]$, we have
\begin{align}
  \sum_{m=1}^{m^\star}\sum_{\lambda=1}^{\lambda_m^\star}\frac{\ln K}{\eta_{m,\lambda}}
  &\le\order\left(\sum_{m=1}^{m^\star}\left(\sqrt{A_m\ln K\,S_m}+A_m\ln K\right)\right)\nonumber\\*
  &\le\order\left(\sqrt{A\ln K\sum_{t=1}^T\ell_{t,I_t}}+A\ln K\right),
  \label{eq:u-adaptive-rate-sum}
\end{align}
where the last inequality uses Cauchy--Schwarz, $\sum_{m=1}^{m^\star}A_m<4A$, and $\sum_{m=1}^{m^\star}S_m=\sum_{t=1}^T\ell_{t,I_t}$.
Plugging \pref{eq:u-adaptive-rate-sum} into \pref{eq:u-adaptive-rate-bound} and applying Jensen's inequality yields
\begin{equation*}
  \Reg_{T,i}
  \le\order\left(\sqrt{A\ln K\left(L_{T,i}+\Reg_{T,i}\right)}+A\ln K\right),
\end{equation*}
where we use $\E[\sum_{t=1}^T\ell_{t,I_t}]=L_{T,i}+\Reg_{T,i}$.
For $\Reg_{T,i}>0$, using $\sqrt{x+y}\le\sqrt{x}+\sqrt{y}$ together with
the AM-GM inequality gives
\begin{equation*}
  \Reg_{T,i}
  \le\frac{\Reg_{T,i}}2+\order\left(\sqrt{A\ln K\,L_{T,i}}+A\ln K\right).
\end{equation*}
Rearranging terms proves $\Reg_{T,i}=\order(\sqrt{A\ln K\,L_{T,i}}+A\ln K)=\otil(\sqrt{\alpha L_{T,i}}+\alpha)$ by \pref{lem:soft-ratio}. The case $\Reg_{T,i}\le0$ is immediate.
\end{proof}

\subsection{Proof of the Residual and Iteration Bounds}
\label{app:computation}

We prove \pref{prop:u-contraction} by bounding the $\ell_1$ residual through the spread of coordinate ratios between consecutive iterates. We show that this spread decreases geometrically and then derive the iteration count for the stopping rule.

\begin{proof}[Proof of \pref{prop:u-contraction}.]
Recalling \pref{eq:u-map}, the iterates satisfy
\begin{align*}
  q_i^{(n+1)}
  &=[\Fu_p(q^{(n)})]_i
    =\frac{p_i\psi(Q_i(q^{(n)}))}{\sum_{j=1}^Kp_j\psi(Q_j(q^{(n)}))},\\*
  \psi(s)
  &=\frac{s+1/(KT)}{s+\eta},
\end{align*}
for every $n\ge0$ and $i\in[K]$.
To bound the residual, denote by $D_n$ the spread of the log coordinate ratios between consecutive iterates, for $n\ge1$:
\begin{equation}
  D_n
  \triangleq\max_{i\in[K]}\ln\frac{q_i^{(n+1)}}{q_i^{(n)}}
    -\min_{i\in[K]}\ln\frac{q_i^{(n+1)}}{q_i^{(n)}}.
  \label{eq:u-log-ratio-spread}
\end{equation}
For $r,q\in\operatorname{int}(\DeltaK)$, denote by $r_{\min}\triangleq\min_jr_j/q_j$ and $r_{\max}\triangleq\max_jr_j/q_j$ the extreme coordinate ratios.
As distributions sum to one, we have $r_{\min}\le1\le r_{\max}$.
Using $1-x\le-\ln x$, we know that
\begin{align*}
  \frac12\norm{r-q}_1
  &=\sum_{\substack{i\in[K]\\r_i>q_i}}(r_i-q_i)
    \le1-\frac1{r_{\max}}\le\ln r_{\max},\\*
  \frac12\norm{r-q}_1
  &=\sum_{\substack{i\in[K]\\r_i<q_i}}(q_i-r_i)
    \le1-r_{\min}\le-\ln r_{\min}.
\end{align*}
Adding these estimates and taking $r=q^{(n+1)}$ and $q=q^{(n)}$ gives
\begin{equation}
  \norm{q^{(n)}-\Fu_p(q^{(n)})}_1
  =\norm{q^{(n+1)}-q^{(n)}}_1\le D_n.
  \label{eq:u-residual-spread}
\end{equation}

To bound $D_n$ in \pref{eq:u-residual-spread}, we compare $D_{n+1}$ with $D_n$. Substituting \pref{eq:u-map} into the coordinate ratio gives
\begin{align}
  \ln\frac{q_i^{(n+2)}}{q_i^{(n+1)}}
  &=\ln\frac{[\Fu_p(q^{(n+1)})]_i}{[\Fu_p(q^{(n)})]_i}\nonumber\\*
  &=\ln\left(
    \frac{p_i\psi(Q_i(q^{(n+1)}))}{\sum_{j=1}^Kp_j\psi(Q_j(q^{(n+1)}))}
    \frac{\sum_{j=1}^Kp_j\psi(Q_j(q^{(n)}))}{p_i\psi(Q_i(q^{(n)}))}
    \right)\nonumber\\*
  &=\ln\frac{\psi(Q_i(q^{(n+1)}))}{\psi(Q_i(q^{(n)}))}
    +\ln\frac{\sum_{j=1}^Kp_j\psi(Q_j(q^{(n)}))}
      {\sum_{j=1}^Kp_j\psi(Q_j(q^{(n+1)}))}.
  \label{eq:u-log-coordinate-ratio}
\end{align}
The second term in \pref{eq:u-log-coordinate-ratio} is independent of $i$, so it cancels when we take the maximum minus the minimum in \pref{eq:u-log-ratio-spread} with index $n+1$. Hence,
\begin{equation}
  D_{n+1}
  =\max_{i\in[K]}\ln\frac{\psi(Q_i(q^{(n+1)}))}{\psi(Q_i(q^{(n)}))}
    -\min_{i\in[K]}\ln\frac{\psi(Q_i(q^{(n+1)}))}{\psi(Q_i(q^{(n)}))}.
  \label{eq:u-iterate-spread}
\end{equation}
To bound the multiplier ratios in \pref{eq:u-iterate-spread}, we first expand the ratio of their arguments. By the definition of the observation probabilities, we have
\begin{align}
  \frac{Q_i(q^{(n+1)})}{Q_i(q^{(n)})}
  =\frac{\sum_{\substack{j\in[K]\\j\to i}}q_j^{(n+1)}}
    {\sum_{\substack{j\in[K]\\j\to i}}q_j^{(n)}}=\sum_{\substack{j\in[K]\\j\to i}}
    \frac{q_j^{(n)}}{Q_i(q^{(n)})}
    \frac{q_j^{(n+1)}}{q_j^{(n)}}.
  \label{eq:u-observation-ratio}
\end{align}
The weights $q_j^{(n)}/Q_i(q^{(n)})$ in \pref{eq:u-observation-ratio} sum to one over the in-neighborhood of $i$. Thus, this ratio lies between the smallest and largest coordinate ratios, and taking logarithms gives
\begin{equation}
  \min_{j\in[K]}\ln\frac{q_j^{(n+1)}}{q_j^{(n)}}
  \le\ln\frac{Q_i(q^{(n+1)})}{Q_i(q^{(n)})}
  \le\max_{j\in[K]}\ln\frac{q_j^{(n+1)}}{q_j^{(n)}}.
  \label{eq:u-observation-log-ratio}
\end{equation}
We now return to the multiplier log ratios in \pref{eq:u-iterate-spread}. They satisfy
\begin{align}
  \ln\frac{\psi(Q_i(q^{(n+1)}))}{\psi(Q_i(q^{(n)}))}
  =\int_{Q_i(q^{(n)})}^{Q_i(q^{(n+1)})}
    \frac{\psi'(s)}{\psi(s)}\,\mathrm{d}s=\int_{Q_i(q^{(n)})}^{Q_i(q^{(n+1)})}
    \frac{s\psi'(s)}{\psi(s)}\frac{\mathrm{d}s}{s}.
  \label{eq:u-multiplier-integral}
\end{align}
Suppose that $0\le s\psi'(s)/\psi(s)\le C<1$ for every $s>0$. Substituting \pref{eq:u-multiplier-integral} into \pref{eq:u-iterate-spread} gives
\begin{align}
  D_{n+1}
  &=\max_{i\in[K]}\int_{Q_i(q^{(n)})}^{Q_i(q^{(n+1)})}
    \frac{s\psi'(s)}{\psi(s)}\frac{\mathrm{d}s}{s}
    -\min_{i\in[K]}\int_{Q_i(q^{(n)})}^{Q_i(q^{(n+1)})}
    \frac{s\psi'(s)}{\psi(s)}\frac{\mathrm{d}s}{s}\nonumber\\*
  &\le C\left(\max_{j\in[K]}\ln\frac{q_j^{(n+1)}}{q_j^{(n)}}
    -\min_{j\in[K]}\ln\frac{q_j^{(n+1)}}{q_j^{(n)}}\right)
    =CD_n,
  \label{eq:projective-multiplier-contraction}
\end{align}
where the inequality holds because each integral lies between $0$ and $C\ln\frac{Q_i(q^{(n+1)})}{Q_i(q^{(n)})}$, which by \pref{eq:u-observation-log-ratio} lies between $C\min_{j\in[K]}\ln\frac{q_j^{(n+1)}}{q_j^{(n)}}\le0$ and $C\max_{j\in[K]}\ln\frac{q_j^{(n+1)}}{q_j^{(n)}}\ge0$.

We now verify this condition for the multiplier in \pref{eq:u-map}. If $\eta=1/(KT)$, then $\psi(s)=1$ and $\Fu_p(q)=p$, so the residual is zero after the first iteration. Suppose $\eta>1/(KT)$. Direct differentiation gives
\begin{equation}
  0\le\frac{s\psi'(s)}{\psi(s)}
  =\frac{\eta-1/(KT)}{s+\eta+1/(KT)+\eta/(KTs)}
  \le\frac{\sqrt{\eta KT}-1}{\sqrt{\eta KT}+1}
  \le1-\frac1{\sqrt{\eta KT}}.
  \label{eq:u-elasticity-bound}
\end{equation}
The first upper bound uses $s+\eta/(KTs)\ge2\sqrt{\eta/(KT)}$ by the AM-GM inequality.
Applying \pref{eq:projective-multiplier-contraction} with $C=1-1/\sqrt{\eta KT}$ therefore gives $D_{n+1}\le(1-1/\sqrt{\eta KT})D_n$. To initialize this recursion, note that $q^{(1)}$ and $q^{(2)}$ are both outputs of $\Fu_p$. Since $1/(\eta KT)\le\psi(s)\le1$, their multiplier ratios lie in $[1/(\eta KT),\eta KT]$, so $D_1\le2\ln(\eta KT)$.
Combining this bound with \pref{eq:u-residual-spread} and \pref{eq:projective-multiplier-contraction} gives
\begin{align*}
  \norm{q^{(n)}-\Fu_p(q^{(n)})}_1
  \le D_n
    \le\left(1-\frac1{\sqrt{\eta KT}}\right)^{n-1}D_1
    \le2\left(1-\frac1{\sqrt{\eta KT}}\right)^{n-1}\ln(\eta KT),
\end{align*}
which proves \pref{eq:u-contraction}.
Since $1-x\le e^{-x}$, \pref{eq:u-contraction} gives
\begin{equation*}
\norm{q^{(n)}-\Fu_p(q^{(n)})}_1\le2e^{-(n-1)/\sqrt{\eta KT}}\ln(\eta KT)\le\eps,
\end{equation*}
for $n\ge1+\sqrt{\eta KT}\,\ln\frac{2\ln(\eta KT)}{\eps}$,
which gives \pref{eq:u-iterations}.
Each iteration computes the $K$ observation probabilities by a matrix-vector product, then evaluates and normalizes the multiplier, taking $\order(K^2)$ time.
With $\eps=1/(KT^2)$, the per-round running time is $\otil(K^2\sqrt{\eta KT})$.

\end{proof}

To apply the iteration bound in \pref{sec:strongly-observable}, we extend \pref{prop:u-contraction} to the subgraph $\Gself$ and an arbitrary sampling parameter $\gamma>0$, keeping the original offset $1/(KT)$.

\begin{proposition}
\label[proposition]{prop:u-arbitrary-iterations}
Let $p\in\operatorname{int}(\Delta(\Vself))$ and $\gamma>0$. Denote by $\psi_\gamma(v)\triangleq\frac{v+\frac{1}{KT}}{v+\gamma}$ the multiplier and by $\Fu_{p,\gamma}$ the map on $\Delta(\Vself)$
\begin{equation*}
  [\Fu_{p,\gamma}(q)]_i
  \triangleq\frac{p_i\psi_\gamma(Q_i^{\Gself}(q))}
    {\sum_{j\in \Vself}p_j\psi_\gamma(Q_j^{\Gself}(q))},
\end{equation*}
for every $i\in \Vself$. Fix any $q^{(0)}\in\operatorname{int}(\Delta(\Vself))$ and set $q^{(n+1)}=\Fu_{p,\gamma}(q^{(n)})$. For $0<\eps\le1/2$, the iteration reaches an $\eps$-fixed point in $\ell_1$ after
\begin{equation}
  N=\order\left(\max\left\{\sqrt{\gamma KT},\frac1{\sqrt{\gamma KT}}\right\}
    \ln\frac{1+|\ln(\gamma KT)|}{\eps}\right)
  \label{eq:u-arbitrary-iterations}
\end{equation}
iterations, each of which takes $\order(K^2)$ time.
In particular, if $1/\Gamma\le\gamma KT\le\Gamma$ for some $\Gamma\ge1$, then $N=\order\left(\sqrt{\Gamma}\ln\frac{1+\ln\Gamma}{\eps}\right)$.
Consequently, the base distribution $q_t$ in \pref{alg:strong}, satisfying $\norm{q_t-\Fu_t^{\mathrm{self}}(q_t)}_1\le\eps$ for the map in \pref{eq:strong-base-map}, can be computed by Picard iteration in time polynomial in $K$, $T$, $A$ and $\ln(1/\eps)$.
\end{proposition}

\begin{proof}
To bound the residual, as in \pref{eq:u-log-ratio-spread} but with the coordinate indices restricted to $\Vself$ and the iterates $q^{(n)}$ generated by $\Fu_{p,\gamma}$, denote by $D_n$ the spread of the log coordinate ratios between consecutive iterates, for $n\ge1$:
\begin{equation*}
D_n
\triangleq\max_{i\in\Vself}\ln\frac{q_i^{(n+1)}}{q_i^{(n)}}
-\min_{i\in\Vself}\ln\frac{q_i^{(n+1)}}{q_i^{(n)}}.
\end{equation*}
The argument for \pref{eq:u-residual-spread} gives $\norm{q^{(n)}-\Fu_{p,\gamma}(q^{(n)})}_1\le D_n$.

To bound $D_n$, we compare $D_{n+1}$ with $D_n$. Direct differentiation gives
\begin{align}
\frac{v\psi_\gamma'(v)}{\psi_\gamma(v)}
&=\frac{\gamma-\frac{1}{KT}}{v+\gamma+\frac{1}{KT}+\frac{\gamma}{KTv}}.
\label{eq:u-arbitrary-elasticity}
\end{align}
Therefore, we have $\left|\frac{v\psi_\gamma'(v)}{\psi_\gamma(v)}\right|\le C_\gamma\triangleq\frac{|\sqrt{\gamma KT}-1|}{\sqrt{\gamma KT}+1}<1$, where the first inequality uses $v+\frac{\gamma}{KTv}\ge2\sqrt{\frac{\gamma}{KT}}$. Thus $\frac{v\psi_\gamma'(v)}{\psi_\gamma(v)}$ has the constant sign of $\gamma-\frac{1}{KT}$, and flipping this sign does not change the spread of the log multiplier ratios. Therefore, the derivations of \pref{eq:u-iterate-spread} and \pref{eq:u-multiplier-integral}, with $Q_i^{\Gself}$ and $\psi_\gamma$ in place of $Q_i$ and $\psi$, give
\begin{align}
D_{n+1}
&=\max_{i\in\Vself}\int_{Q_i^{\Gself}(q^{(n)})}^{Q_i^{\Gself}(q^{(n+1)})}\left|\frac{v\psi_\gamma'(v)}{\psi_\gamma(v)}\right|\frac{\mathrm{d}v}{v}\nonumber-\min_{i\in\Vself}\int_{Q_i^{\Gself}(q^{(n)})}^{Q_i^{\Gself}(q^{(n+1)})}\left|\frac{v\psi_\gamma'(v)}{\psi_\gamma(v)}\right|\frac{\mathrm{d}v}{v}\nonumber\\
&\le C_\gamma\left(\max_{j\in\Vself}\ln\frac{q_j^{(n+1)}}{q_j^{(n)}}-\min_{j\in\Vself}\ln\frac{q_j^{(n+1)}}{q_j^{(n)}}\right)
=C_\gamma D_n,
\label{eq:u-arbitrary-spread-decay}
\end{align}
where the inequality holds because each integral lies between $0$ and $C_\gamma\ln\frac{Q_i^{\Gself}(q^{(n+1)})}{Q_i^{\Gself}(q^{(n)})}$, which by \pref{eq:u-observation-log-ratio} lies between $C_\gamma\min_{j\in\Vself}\ln\frac{q_j^{(n+1)}}{q_j^{(n)}}\le0$ and $C_\gamma\max_{j\in\Vself}\ln\frac{q_j^{(n+1)}}{q_j^{(n)}}\ge0$.
To initialize this recursion, $\psi_\gamma$ takes values between $\frac{1}{\gamma KT}$ and $1$, so $D_1\le2|\ln(\gamma KT)|$. Combining this bound with \pref{eq:u-arbitrary-spread-decay} gives
\begin{equation*}
\norm{q^{(n)}-\Fu_{p,\gamma}(q^{(n)})}_1
\le D_n
\le C_\gamma^{n-1}D_1
\le2C_\gamma^{n-1}|\ln(\gamma KT)|.
\end{equation*}
Since $1-x\le e^{-x}$, this residual is at most $2e^{-(n-1)(1-C_\gamma)}|\ln(\gamma KT)|$, which is at most $\eps$ once $n\ge1+\frac{1}{1-C_\gamma}\ln\frac{2|\ln(\gamma KT)|}{\eps}$. Since $\frac{1}{1-C_\gamma}=\frac{1}{2}\left(1+\max\left\{\sqrt{\gamma KT},\frac{1}{\sqrt{\gamma KT}}\right\}\right)$, this gives \pref{eq:u-arbitrary-iterations}. Each iteration takes $\order(K^2)$ time by the same matrix-vector computation as in \pref{prop:u-contraction}.

To prove the claim for \pref{eq:strong-base-map}, take $p=p_t$ and $\gamma=\frac{\eta_t}{x_{t,0}}$. Then \pref{eq:strong-base-map} gives
\begin{equation*}
  \Fu_{p_t,\gamma}(q)=\Fu_t^{\mathrm{self}}(q).
\end{equation*}
Thus the iteration bound above applies with $\gamma=\frac{\eta_t}{x_{t,0}}$, and it remains to bound this parameter. The clipped master domain gives $x_{t,0}\ge\frac{1}{2MT}$ and $\frac{\widetilde g_{t,0}}{x_{t,0}}\le(2MT)^2$. Substituting these bounds into \pref{eq:strong-base-learning-rate} gives
\begin{equation*}
  \sqrt{\frac{\ln(sT)}{2A(1+4M^2T^3)}}
  \le\frac{\eta_t}{x_{t,0}}
  \le2MT\sqrt{\frac{\ln(sT)}{2A}}.
\end{equation*}
Therefore, $\frac{1}{\Gamma}\le\frac{\eta_tKT}{x_{t,0}}\le\Gamma$ with $\Gamma$ polynomial in $K$, $T$ and $A$, and the iteration count $\order(\sqrt{\Gamma}\ln\frac{1+\ln\Gamma}{\eps})$ with $\order(K^2)$ time per iteration gives the claimed running time.
\end{proof}

\subsection{A Refined Analysis of EXP3}
\label{app:potentials}

In this section, we prove the slightly refined analysis of EXP3 in \pref{eq:ordinary-potential}, stated as follows.

\begin{proposition}
\label[proposition]{prop:ordinary-potential}
For the EXP3 update \pref{eq:u-omd-update} with fixed learning rate $\eta>0$, uniform $p_1$ and any nonnegative estimates, every comparator $i\in[K]$ satisfies
\begin{equation*}
  \sum_{t=1}^T\ip{p_t-e_i}{\ellhat_t}
  \le\frac{\ln K}{\eta}
    +\eta\sum_{t=1}^T\sum_{j=1}^K\frac{p_{t,j}\ellhat_{t,j}^2}{1+\eta\ellhat_{t,j}}.
\end{equation*}
\end{proposition}

\begin{proof}
Denote by $W_t\triangleq\sum_{j=1}^Kp_{t,j}e^{-\eta\ellhat_{t,j}}$ the normalizer in \pref{eq:u-omd-update}.
To bound the change in the log weight of comparator $i$, we first bound $\ln W_t$ as follows:
\begin{align}
  \ln W_t
  \le W_t-1\le-\eta\ip{p_t}{\ellhat_t}
    +\eta^2\sum_{j=1}^K\frac{p_{t,j}\ellhat_{t,j}^2}{1+\eta\ellhat_{t,j}},
  \label{eq:ordinary-upper}
\end{align}
where the first inequality uses $\ln x\le x-1$ for $x>0$, and the second uses $e^{-x}\le1/(1+x)=1-x+x^2/(1+x)$ for $x\ge0$.
Substituting \pref{eq:ordinary-upper} into $\ln p_{t+1,i}-\ln p_{t,i}=-\eta\ellhat_{t,i}-\ln W_t$ gives
\begin{equation*}
  \ip{p_t-e_i}{\ellhat_t}
  \le\frac{\ln p_{t+1,i}-\ln p_{t,i}}{\eta}
    +\eta\sum_{j=1}^K\frac{p_{t,j}\ellhat_{t,j}^2}{1+\eta\ellhat_{t,j}}.
\end{equation*}
Summing over $t$ and using $p_{1,i}=1/K$ and $p_{T+1,i}\le1$ proves the claim.
\end{proof}

\section{Omitted Details in \texorpdfstring{\pref{sec:strongly-observable}}{Section 4}}
\label{app:strongly-observable}

In this section, we provide the omitted details of \pref{sec:strongly-observable}.
\pref{app:strong-structure} proves the regret bound of the self-aware learner, \pref{app:strong-master} proves \pref{lem:strong-components} and \pref{thm:strong-fixed}, and \pref{app:strong-adaptive} presents the doubling trick and proves \pref{thm:strong-adaptive}.
Throughout, $c>0$ is the universal constant in \pref{thm:strong-fixed}.

\subsection{The Self-Aware Learner}
\label{app:strong-structure}

We adapt the analysis of Lemma~11 of \citet{lee2020graph} to the fixed-point sampling rule.
Since the doubling trick in \pref{app:strong-adaptive} restarts \pref{alg:strong} at random rounds, we state the bound for an instance that runs on an interval of rounds.

\begin{lemma}
\label[lemma]{lem:scaled-stability-base}
Consider an instance of \pref{alg:strong} with $0\in\Bcal$ that starts at round $\tau$ and runs on an interval $\mathcal I$ of rounds, where whether $t\in\mathcal I$ is determined before round $t$ is sampled, and suppose $\mathcal H^{\Gself}(q_t)\le A$ for all $t\in\mathcal I$.
Denote by $\E_\tau$ the expectation conditional on the history before $\tau$, by $\bar\rho_{\mathcal I}\triangleq\max_{t\in\mathcal I}\frac1{x_{t,0}}$ the largest inverse probability of meta-action $0$, and by $r_{\mathcal I}\triangleq\E_\tau[\bar\rho_{\mathcal I}]$ and $L_{\mathcal I,k}\triangleq\E_\tau[\sum_{t\in\mathcal I}\ell_{t,k}]$.
Then, for every $k\in\Vself$,
\begin{align}
R_{\mathcal I,k}\triangleq\E_\tau\left[\sum_{t\in\mathcal I}(\ip{q_t}{\ell_t}-\ell_{t,k})\right]
\le(56A\ln(sT)+6J)r_{\mathcal I}
+8\sqrt{2A\ln(sT)r_{\mathcal I}L_{\mathcal I,k}}.
\label{eq:scaled-base-bound}
\end{align}
\end{lemma}

\begin{proof}
If $s=1$, then $q_t=e_k$ and the regret is zero, so suppose $s\ge2$.
Fix $k\in\Vself$ and the clipped comparator $u\triangleq(1-\frac1T)e_k+\frac1{sT}\sum_{i\in\Vself}e_i$, which lies in $\mathcal P$ since its coordinates sum to one and are at least $\frac1{sT}$.
Denote by $\widetilde p_{t+1}\in\R_+^{\Vself}$ the EXP3 step from $p_t$ without normalization and clipping, with coordinates $\widetilde p_{t+1,i}\triangleq p_{t,i}e^{-\eta_t\widetilde\ell_{t,i}}$, and by $W_t\triangleq\sum_{i\in\Vself}\widetilde p_{t+1,i}$ its total mass, as in the proof of \pref{prop:ordinary-potential}. Recall that $D_h(z,y)=\sum_{i\in\Vself}(z_i\ln\frac{z_i}{y_i}-z_i+y_i)$ for $z,y\in\R_+^{\Vself}$. For every $z\in\Delta(\Vself)$, substituting the definition of $\widetilde p_{t+1}$ gives
\begin{align}
D_h(z,\widetilde p_{t+1})
=D_h(z,p_t)+\eta_t\ip{z}{\widetilde\ell_t}+W_t-1.
\label{eq:scaled-unnormalized}
\end{align}
Since $W_t-1$ does not depend on $z$, \pref{eq:scaled-unnormalized} shows that the update in \pref{line:strong-p-update} minimizes $D_h(\cdot,\widetilde p_{t+1})$ over the convex set $\mathcal P$. As $u\in\mathcal P$, the generalized Pythagorean inequality for this minimizer gives $D_h(u,p_{t+1})\le D_h(u,\widetilde p_{t+1})$. Applying \pref{eq:scaled-unnormalized} with $z=u$ and this inequality, we obtain
\begin{align}
\ip{p_t-u}{\widetilde\ell_t}
&\le\frac{D_h(u,p_t)-D_h(u,p_{t+1})}{\eta_t}+\frac{W_t-1}{\eta_t}+\ip{p_t}{\widetilde\ell_t}\nonumber\\
&\le\frac{D_h(u,p_t)-D_h(u,p_{t+1})}{\eta_t}
+\sum_{i\in\Vself}\frac{\eta_t p_{t,i}\widetilde\ell_{t,i}^2}{1+\eta_t\widetilde\ell_{t,i}},
\label{eq:scaled-omd-step}
\end{align}
where the second inequality is the second inequality of \pref{eq:ordinary-upper} with $\eta_t$ in place of $\eta$.
Since $\eta_t$ is nonincreasing and $D_h(u,p)\le\ln(sT)$ for every $p\in\mathcal P$, summing the first term of \pref{eq:scaled-omd-step} over $t\in\mathcal I$ gives
\begin{align}
\sum_{t\in\mathcal I}\frac{D_h(u,p_t)-D_h(u,p_{t+1})}{\eta_t}
\le\frac{\ln(sT)}{\eta_{t'}}
\le\sqrt{2A\ln(sT)\left(1+\sum_{t\in\mathcal I}\frac{\widetilde g_{t,0}}{x_{t,0}}\right)},
\label{eq:scaled-divergence-sum}
\end{align}
where $t'$ denotes the last round of $\mathcal I$, and the second inequality uses \pref{eq:strong-base-learning-rate}.

It remains to bound the loss discrepancy $\ip{q_t-p_t}{\ell_t}$ together with the last term of \pref{eq:scaled-omd-step}. Multiplying \pref{eq:strong-base-moment} by $p_{t,i}$ and summing over $i\in\Vself$, we have
\begin{align}
&\ip{q_t-p_t}{\ell_t}+\Et\left[\sum_{i\in\Vself}\frac{\eta_tp_{t,i}\widetilde\ell_{t,i}^2}{1+\eta_t\widetilde\ell_{t,i}}\right]\nonumber\\
&\le\ip{q_t-p_t}{\ell_t}+\sum_{i\in\Vself}\frac{\frac{\eta_t}{x_{t,0}}p_{t,i}\ell_{t,i}}{Q_i^{\Gself}(q_t)+\frac{\eta_t}{x_{t,0}}}
=\sum_{i\in\Vself}\ell_{t,i}\left(q_{t,i}-\frac{p_{t,i}Q_i^{\Gself}(q_t)}{Q_i^{\Gself}(q_t)+\frac{\eta_t}{x_{t,0}}}\right).
\label{eq:scaled-remainder}
\end{align}
To bound the right-hand side of \pref{eq:scaled-remainder}, denote by $\widehat q_t\triangleq\Fu^{\mathrm{self}}_t(q_t)$ the updated iterate and by $Z^{\mathrm{self}}_t\triangleq1-\sum_{i\in\Vself}p_{t,i}\psi_t(Q_i^{\Gself}(q_t))$ the remainder coefficient. By \pref{eq:strong-base-map} and the definition of $\psi_t$, for every $i\in\Vself$,
\begin{align}
(1-Z^{\mathrm{self}}_t)\widehat q_{t,i}
=p_{t,i}\psi_t(Q_i^{\Gself}(q_t))
=\frac{p_{t,i}\left(Q_i^{\Gself}(q_t)+\frac1{KT}\right)}{Q_i^{\Gself}(q_t)+\frac{\eta_t}{x_{t,0}}}.
\label{eq:scaled-coordinate-identity}
\end{align}
Since $\norm{q_t-\widehat q_t}_1\le\eps$, we have
\begin{align}
\sum_{i\in\Vself}\frac{\widehat q_{t,i}}{Q_i^{\Gself}(q_t)+\frac1{KT}}
\le\sum_{i\in\Vself}\frac{q_{t,i}}{Q_i^{\Gself}(q_t)+\frac1{KT}}+KT\norm{q_t-\widehat q_t}_1
\le\mathcal H^{\Gself}(q_t)+KT\eps.
\label{eq:scaled-estimation-ratio}
\end{align}
Moreover,
\begin{align}
Z^{\mathrm{self}}_t
&=\sum_{i\in\Vself}p_{t,i}\left(1-\psi_t(Q_i^{\Gself}(q_t))\right)
=\left(\frac{\eta_t}{x_{t,0}}-\frac1{KT}\right)\sum_{i\in\Vself}\frac{p_{t,i}}{Q_i^{\Gself}(q_t)+\frac{\eta_t}{x_{t,0}}}\nonumber\\
&=\left(\frac{\eta_t}{x_{t,0}}-\frac1{KT}\right)(1-Z^{\mathrm{self}}_t)\sum_{i\in\Vself}\frac{\widehat q_{t,i}}{Q_i^{\Gself}(q_t)+\frac1{KT}},
\label{eq:scaled-Z-identity}
\end{align}
where the first equality uses $\sum_{i\in\Vself}p_{t,i}=1$, the second uses the definition of $\psi_t$, and the third uses \pref{eq:scaled-coordinate-identity} divided by $Q_i^{\Gself}(q_t)+\frac1{KT}$.

We now bound $Z^{\mathrm{self}}_t$ in two cases, using $1-Z^{\mathrm{self}}_t=\sum_{i\in\Vself}p_{t,i}\psi_t(Q_i^{\Gself}(q_t))>0$. If $\frac{\eta_t}{x_{t,0}}\ge\frac1{KT}$, then \pref{eq:scaled-Z-identity} gives $0\le Z^{\mathrm{self}}_t<1$, and with $1-Z^{\mathrm{self}}_t\le1$, \pref{eq:scaled-estimation-ratio}, $\mathcal H^{\Gself}(q_t)\le A$ and $KT\eps=\frac1T\le A$, it gives $Z^{\mathrm{self}}_t\le\frac{2A\eta_t}{x_{t,0}}$. If $\frac{\eta_t}{x_{t,0}}<\frac1{KT}$, then \pref{eq:scaled-Z-identity} gives $Z^{\mathrm{self}}_t<0$, and solving \pref{eq:scaled-coordinate-identity} for $p_{t,i}$ gives, for every $i\in\Vself$,
\begin{align*}
p_{t,i}
=(1-Z^{\mathrm{self}}_t)\frac{\widehat q_{t,i}\left(Q_i^{\Gself}(q_t)+\frac{\eta_t}{x_{t,0}}\right)}{Q_i^{\Gself}(q_t)+\frac1{KT}}.
\end{align*}
Summing this identity over $i\in\Vself$, using $\sum_{i\in\Vself}p_{t,i}=1$ and dividing both sides by $1-Z^{\mathrm{self}}_t>0$, we obtain
\begin{align*}
\frac1{1-Z^{\mathrm{self}}_t}
&=\sum_{i\in\Vself}\frac{\widehat q_{t,i}\left(Q_i^{\Gself}(q_t)+\frac{\eta_t}{x_{t,0}}\right)}{Q_i^{\Gself}(q_t)+\frac1{KT}}
\ge\sum_{i\in\Vself}\frac{\widehat q_{t,i}Q_i^{\Gself}(q_t)}{Q_i^{\Gself}(q_t)+\frac1{KT}}
=1-\frac1{KT}\sum_{i\in\Vself}\frac{\widehat q_{t,i}}{Q_i^{\Gself}(q_t)+\frac1{KT}}\\
&\ge1-\frac{\mathcal H^{\Gself}(q_t)+KT\eps}{KT}
\ge1-\frac1T-\eps
>\frac12,
\end{align*}
where the second equality uses $\sum_{i\in\Vself}\widehat q_{t,i}=1$, the second inequality uses \pref{eq:scaled-estimation-ratio}, and the third uses $\mathcal H^{\Gself}(q_t)\le K$. In both cases, $Z^{\mathrm{self}}_t\le\min\{1,\frac{2A\eta_t}{x_{t,0}}\}$ whenever $Z^{\mathrm{self}}_t\ge0$, and $1-Z^{\mathrm{self}}_t\le2$. Therefore, the right-hand side of \pref{eq:scaled-remainder} satisfies
\begin{align}
&\sum_{i\in\Vself}\ell_{t,i}\left(q_{t,i}-\frac{p_{t,i}Q_i^{\Gself}(q_t)}{Q_i^{\Gself}(q_t)+\frac{\eta_t}{x_{t,0}}}\right)\nonumber\\
&=\sum_{i\in\Vself}\ell_{t,i}\left(q_{t,i}-p_{t,i}\psi_t(Q_i^{\Gself}(q_t))+\frac{p_{t,i}}{KT\left(Q_i^{\Gself}(q_t)+\frac{\eta_t}{x_{t,0}}\right)}\right)
\tag*{(definition of $\psi_t$)}\\
&=\sum_{i\in\Vself}\ell_{t,i}\left(q_{t,i}-(1-Z^{\mathrm{self}}_t)\widehat q_{t,i}+\frac{(1-Z^{\mathrm{self}}_t)\widehat q_{t,i}}{KT\left(Q_i^{\Gself}(q_t)+\frac1{KT}\right)}\right)
\tag*{(by \pref{eq:scaled-coordinate-identity})}\\
&=Z^{\mathrm{self}}_t\ip{q_t}{\ell_t}+(1-Z^{\mathrm{self}}_t)\ip{q_t-\widehat q_t}{\ell_t}+\frac{1-Z^{\mathrm{self}}_t}{KT}\sum_{i\in\Vself}\frac{\widehat q_{t,i}\ell_{t,i}}{Q_i^{\Gself}(q_t)+\frac1{KT}},
\label{eq:scaled-remainder-split}
\end{align}
where the last equality regroups the terms. We bound the three terms on the right-hand side of \pref{eq:scaled-remainder-split} separately:
\begin{align*}
Z^{\mathrm{self}}_t\ip{q_t}{\ell_t}&\le\min\left\{1,\frac{2A\eta_t}{x_{t,0}}\right\}g_{t,0},\\
(1-Z^{\mathrm{self}}_t)\ip{q_t-\widehat q_t}{\ell_t}&\le2\eps,\\
\frac{1-Z^{\mathrm{self}}_t}{KT}\sum_{i\in\Vself}\frac{\widehat q_{t,i}\ell_{t,i}}{Q_i^{\Gself}(q_t)+\frac1{KT}}&\le\frac{2(K+KT\eps)}{KT},
\end{align*}
where the first inequality uses the two cases above and $g_{t,0}=\ip{q_t}{\ell_t}$, the second uses $1-Z^{\mathrm{self}}_t\le2$ and $\norm{q_t-\widehat q_t}_1\le\eps$, and the third uses $1-Z^{\mathrm{self}}_t\le2$, $\ell_t\in[0,1]^K$, \pref{eq:scaled-estimation-ratio} and $\mathcal H^{\Gself}(q_t)\le K$.
Summing these three bounds gives
\begin{align}
\sum_{i\in\Vself}\ell_{t,i}\left(q_{t,i}-\frac{p_{t,i}Q_i^{\Gself}(q_t)}{Q_i^{\Gself}(q_t)+\frac{\eta_t}{x_{t,0}}}\right)
\le\min\left\{1,\frac{2A\eta_t}{x_{t,0}}\right\}g_{t,0}+2\eps+\frac{2(K+KT\eps)}{KT}.
\label{eq:scaled-remainder-bound}
\end{align}
Combining \pref{eq:scaled-remainder} and \pref{eq:scaled-remainder-bound} with $\min\{1,\frac{2A\eta_t}{x_{t,0}}\}\le\frac{2A\eta_t}{x_{t,0}}$ proves \pref{eq:base_stab}.

We now sum the bounds above over $\mathcal I$. Since $\Et[\widetilde\ell_{t,i}]=\ell_{t,i}$ for $i\in\Vself$ and whether $t\in\mathcal I$ is determined before round $t$ is sampled, we have
\begin{align*}
R_{\mathcal I,k}
&=\E_\tau\left[\sum_{t\in\mathcal I}\left(\ip{q_t-p_t}{\ell_t}+\ip{p_t-u}{\widetilde\ell_t}+\ip{u-e_k}{\ell_t}\right)\right]\\
&\le\E_\tau\left[\sum_{t\in\mathcal I}\left(\ip{q_t-p_t}{\ell_t}+\sum_{i\in\Vself}\frac{\eta_tp_{t,i}\widetilde\ell_{t,i}^2}{1+\eta_t\widetilde\ell_{t,i}}\right)\right]
+\E_\tau\left[\sum_{t\in\mathcal I}\frac{D_h(u,p_t)-D_h(u,p_{t+1})}{\eta_t}\right]\tag*{(by \pref{eq:scaled-omd-step})}\\
&\phantom{\le}+\E_\tau\left[\sum_{t\in\mathcal I}\ip{u-e_k}{\ell_t}\right].
\end{align*}
We bound the three terms on the right-hand side separately:
\begin{align*}
\E_\tau\left[\sum_{t\in\mathcal I}\left(\ip{q_t-p_t}{\ell_t}+\sum_{i\in\Vself}\frac{\eta_tp_{t,i}\widetilde\ell_{t,i}^2}{1+\eta_t\widetilde\ell_{t,i}}\right)\right]
&\le\E_\tau\left[\sum_{t\in\mathcal I}\min\left\{1,\frac{2A\eta_t}{x_{t,0}}\right\}g_{t,0}\right]+2+\frac4{KT},\\
\E_\tau\left[\sum_{t\in\mathcal I}\frac{D_h(u,p_t)-D_h(u,p_{t+1})}{\eta_t}\right]
&\le\E_\tau\left[\sqrt{2A\ln(sT)\left(1+\sum_{t\in\mathcal I}\frac{\widetilde g_{t,0}}{x_{t,0}}\right)}\right],\\
\E_\tau\left[\sum_{t\in\mathcal I}\ip{u-e_k}{\ell_t}\right]
&\le1,
\end{align*}
where the first inequality takes the conditional expectation $\Et$ of the term of each round $t\in\mathcal I$ and applies \pref{eq:scaled-remainder} and \pref{eq:scaled-remainder-bound}, together with $|\mathcal I|\le T$ and $T\left(2\eps+\frac{2(K+KT\eps)}{KT}\right)=2+\frac4{KT}$, the second uses \pref{eq:scaled-divergence-sum}, and the third uses $\ip{u-e_k}{\ell_t}\le\frac1T$ and $|\mathcal I|\le T$. Summing these three bounds gives
\begin{align}
R_{\mathcal I,k}
\le\E_\tau\left[\sqrt{2A\ln(sT)\left(1+\sum_{t\in\mathcal I}\frac{\widetilde g_{t,0}}{x_{t,0}}\right)}
+\sum_{t\in\mathcal I}\min\left\{1,\frac{2A\eta_t}{x_{t,0}}\right\}g_{t,0}\right]+3+\frac4{KT}.
\label{eq:scaled-base-decomposition}
\end{align}

To bound the second sum $\sum_{t\in\mathcal I}\min\{1,\frac{2A\eta_t}{x_{t,0}}\}g_{t,0}$ in \pref{eq:scaled-base-decomposition}, note that $\eta_t$ and $x_{t,0}$ are determined before round $t$ and $\Et[\widetilde g_{t,0}]=g_{t,0}$, so
\begin{align*}
\E_\tau\left[\sum_{t\in\mathcal I}\min\left\{1,\frac{2A\eta_t}{x_{t,0}}\right\}g_{t,0}\right]
=\E_\tau\left[\sum_{t\in\mathcal I}\min\left\{\widetilde g_{t,0},\frac{2A\eta_t\widetilde g_{t,0}}{x_{t,0}}\right\}\right].
\end{align*}
By \pref{eq:strong-base-learning-rate}, for every $t\in\mathcal I$,
\begin{align*}
\frac1{\eta_{t+1}^2}-\frac1{\eta_t^2}=\frac{2A\widetilde g_{t,0}}{\ln(sT)x_{t,0}},
\end{align*}
where, for the last round $t'$ of $\mathcal I$, we let $\eta_{t'+1}$ denote the value of \pref{eq:strong-base-learning-rate} with the sum up to $t'$. Thus, if $\eta_{t+1}\ge\frac{\eta_t}{\sqrt2}$, then
\begin{align*}
\frac{2A\eta_t\widetilde g_{t,0}}{x_{t,0}}
&=\ln(sT)\eta_t\left(\frac1{\eta_{t+1}^2}-\frac1{\eta_t^2}\right)
=\ln(sT)\left(1+\frac{\eta_t}{\eta_{t+1}}\right)\left(\frac1{\eta_{t+1}}-\frac1{\eta_t}\right)\\
&\le(1+\sqrt2)\ln(sT)\left(\frac1{\eta_{t+1}}-\frac1{\eta_t}\right).
\end{align*}
Taking the summation over the rounds $t\in\mathcal I$ with $\eta_{t+1}\ge\frac{\eta_t}{\sqrt2}$, we have
\begin{align*}
\sum_{t\in\mathcal I:\,\eta_{t+1}\ge\eta_t/\sqrt2}\frac{2A\eta_t\widetilde g_{t,0}}{x_{t,0}}
\le\frac{(1+\sqrt2)\ln(sT)}{\eta_{t'+1}}
=(1+\sqrt2)\sqrt{2A\ln(sT)\left(1+\sum_{t\in\mathcal I}\frac{\widetilde g_{t,0}}{x_{t,0}}\right)}.
\end{align*}
Each round $t\in\mathcal I$ with $\eta_{t+1}<\frac{\eta_t}{\sqrt2}$ contributes at most $\widetilde g_{t,0}\le\bar\rho_{\mathcal I}$.
There are at most $\log_2(1+\sum_{t\in\mathcal I}\frac{\widetilde g_{t,0}}{x_{t,0}})\le3J$ such rounds, since each of them satisfies $\frac1{\eta_{t+1}^2}>\frac2{\eta_t^2}$, while by \pref{eq:strong-base-learning-rate}, $\frac1{\eta_t^2}$ grows over $\mathcal I$ by the factor $1+\sum_{t\in\mathcal I}\frac{\widetilde g_{t,0}}{x_{t,0}}\le1+T(2MT)^2\le(2MT)^3$, where we use $\frac{\widetilde g_{t,0}}{x_{t,0}}\le\bar\rho_{\mathcal I}^2\le(2MT)^2$.
Combining the two cases, we have
\begin{align}
\E_\tau\left[\sum_{t\in\mathcal I}\min\left\{1,\frac{2A\eta_t}{x_{t,0}}\right\}g_{t,0}\right]
\le(1+\sqrt2)\sqrt{2A\ln(sT)}\,\E_\tau\left[\sqrt{1+\sum_{t\in\mathcal I}\frac{\widetilde g_{t,0}}{x_{t,0}}}\right]+3Jr_{\mathcal I}.
\label{eq:scaled-second-sum}
\end{align}
Substituting \pref{eq:scaled-second-sum} into \pref{eq:scaled-base-decomposition} gives
\begin{align}
R_{\mathcal I,k}\le(2+\sqrt{2})\sqrt{2A\ln(sT)}\,\E_\tau\left[\sqrt{1+\sum_{t\in\mathcal I}\frac{\widetilde g_{t,0}}{x_{t,0}}}\right]+3Jr_{\mathcal I}+3+\frac4{KT}.
\label{eq:scaled-before-cs}
\end{align}
Since $\sum_{t\in\mathcal I}\frac{\widetilde g_{t,0}}{x_{t,0}}\le\bar\rho_{\mathcal I}\sum_{t\in\mathcal I}\widetilde g_{t,0}$ and $\E_\tau[\sum_{t\in\mathcal I}\widetilde g_{t,0}]=\E_\tau[\sum_{t\in\mathcal I}\ip{q_t}{\ell_t}]=L_{\mathcal I,k}+R_{\mathcal I,k}$, we have
\begin{align*}
\E_\tau\left[\sqrt{1+\sum_{t\in\mathcal I}\frac{\widetilde g_{t,0}}{x_{t,0}}}\right]
&\le1+\E_\tau\left[\sqrt{\bar\rho_{\mathcal I}\sum_{t\in\mathcal I}\widetilde g_{t,0}}\right]\\
&\le1+\sqrt{\E_\tau[\bar\rho_{\mathcal I}]\,\E_\tau\left[\sum_{t\in\mathcal I}\widetilde g_{t,0}\right]}\tag*{(Cauchy--Schwarz)}\\
&=1+\sqrt{r_{\mathcal I}(L_{\mathcal I,k}+R_{\mathcal I,k})}.
\end{align*}
If $R_{\mathcal I,k}\le0$, then \pref{eq:scaled-base-bound} holds trivially. Otherwise, substituting this bound into \pref{eq:scaled-before-cs} and using $\sqrt{L_{\mathcal I,k}+R_{\mathcal I,k}}\le\sqrt{L_{\mathcal I,k}}+\sqrt{R_{\mathcal I,k}}$ gives
\begin{align*}
R_{\mathcal I,k}
&\le(2+\sqrt2)\left(\sqrt{2A\ln(sT)}+\sqrt{2A\ln(sT)r_{\mathcal I}L_{\mathcal I,k}}+\sqrt{2A\ln(sT)r_{\mathcal I}R_{\mathcal I,k}}\right)+3Jr_{\mathcal I}+3+\frac4{KT}\\
&\le4\sqrt{2A\ln(sT)}+4\sqrt{2A\ln(sT)r_{\mathcal I}L_{\mathcal I,k}}+4\sqrt{2A\ln(sT)r_{\mathcal I}R_{\mathcal I,k}}+3Jr_{\mathcal I}+4\\
&\le4\sqrt{2A\ln(sT)}+4\sqrt{2A\ln(sT)r_{\mathcal I}L_{\mathcal I,k}}+\frac{R_{\mathcal I,k}}2+16A\ln(sT)r_{\mathcal I}+3Jr_{\mathcal I}+4,
\end{align*}
where the last inequality uses the AM--GM inequality $4\sqrt{2A\ln(sT)r_{\mathcal I}R_{\mathcal I,k}}\le\frac{R_{\mathcal I,k}}2+16A\ln(sT)r_{\mathcal I}$. Rearranging gives
\begin{align*}
R_{\mathcal I,k}\le8\sqrt{2A\ln(sT)r_{\mathcal I}L_{\mathcal I,k}}+32A\ln(sT)r_{\mathcal I}+6Jr_{\mathcal I}+8\sqrt{2A\ln(sT)}+8.
\end{align*}
Finally, since $r_{\mathcal I}\ge1$ and $A\ln(sT)\ge1$, we have $8\sqrt{2A\ln(sT)}+8\le24A\ln(sT)r_{\mathcal I}$, which proves \pref{eq:scaled-base-bound}.
\end{proof}

\subsection{The Hybrid Master and the Regret Guarantee}
\label{app:strong-master}

The master analysis starts from the one-step mirror-descent inequality, bounds the remaining terms of this inequality coordinate by coordinate, and then bounds the change of the regularizer when the learning rate of meta-action $0$ increases.
Denote by $\varphi_0(z)\triangleq-\ln z$ and $\varphi(z)\triangleq z\ln z-\ln(1-z)$ the scalar components of \pref{eq:strong-regularizer}. The next lemma bounds $\sup_{0<y<1}\{(p-y)g-\frac1aD_f(y,p)\}$ for $f\in\{\varphi_0,\varphi\}$.

\begin{lemma}
\label[lemma]{lem:hybrid-coordinate-stability}
For $p\in(0,1)$, $g\ge0$, $a>0$ and $\theta>0$,
\begin{align}
\sup_{0<y<1}\left\{(p-y)g-\frac1aD_{\varphi_0}(y,p)\right\}
&\le\frac a2p^2g^2,
\label{eq:block-coordinate-stability}\\
\sup_{0<y<1}\left\{(p-y)g-\frac1{\theta} D_{\varphi}(y,p)\right\}
&\le\theta\min\{p,(1-p)^2\}g^2,
\label{eq:complement-coordinate-stability}
\end{align}
where the second inequality assumes $\theta\le\frac{1}{8}$ and $g\le\frac1{1-p}$.
\end{lemma}

\begin{proof}
Both inequalities trivially hold when $y\ge p$, since then $(p-y)g\le0$ and Bregman divergences are nonnegative. Now fix $y<p$.

For \pref{eq:block-coordinate-stability}, direct calculation gives
\begin{align*}
D_{\varphi_0}(y,p)=\frac yp-1-\ln\frac yp\ge\frac{(p-y)^2}{2p^2},
\end{align*}
where the inequality uses $z-1-\ln z\ge\frac{(1-z)^2}{2}$ for $z\in(0,1]$, which holds since the difference is nonincreasing in $z$ and vanishes at $z=1$. Therefore,
\begin{align*}
(p-y)g-\frac1aD_{\varphi_0}(y,p)
\le(p-y)g-\frac{(p-y)^2}{2ap^2}
\le\frac a2p^2g^2,
\end{align*}
where the last inequality is by the AM--GM inequality.

For \pref{eq:complement-coordinate-stability}, direct calculation gives
\begin{align*}
D_\varphi(y,p)
=\left(y\ln\frac yp-y+p\right)
+\left(\frac{p-y}{1-p}-\ln\frac{1-y}{1-p}\right),
\end{align*}
where the two parenthesized terms are the Bregman divergences of $z\ln z$ and $-\ln(1-z)$, and hence both nonnegative. If $p\le\frac12$, we lower bound $D_\varphi(y,p)$ by $y\ln\frac yp-y+p$, which is at least $\frac{(p-y)^2}{2p}$ since $z\ln z-z+1\ge\frac{(1-z)^2}{2}$ for $z\in(0,1]$ (again, the difference is nonincreasing in $z$ and vanishes at $z=1$). This gives
\begin{align*}
(p-y)g-\frac1\theta D_\varphi(y,p)
\le(p-y)g-\frac{(p-y)^2}{2\theta p}
\le\frac\theta2pg^2
\le\theta\min\{p,(1-p)^2\}g^2,
\end{align*}
where the second inequality is by the AM--GM inequality and the last uses $\frac p2\le(1-p)^2$ for $p\le\frac12$. If $p>\frac12$, we instead lower bound $D_\varphi(y,p)$ by $\frac{p-y}{1-p}-\ln\frac{1-y}{1-p}$. Note that $\theta(1-p)g\le\frac18$, since $\theta\le\frac18$ and $g\le\frac1{1-p}$. This gives
\begin{align*}
(p-y)g-\frac1\theta D_\varphi(y,p)
&\le(p-y)g-\frac1\theta\left(\frac{p-y}{1-p}-\ln\frac{1-y}{1-p}\right)\\
&\le-(1-p)g-\frac1\theta\ln\left(1-\theta(1-p)g\right)
\le\theta(1-p)^2g^2,
\end{align*}
where the second inequality holds since the right-hand side of the first inequality is concave in $y<1$ and maximized at $1-y=\frac{1-p}{1-\theta(1-p)g}$ and the last uses $\ln(1-x)\ge-x-x^2$ for $x=\theta(1-p)g\in[0,\frac12]$. Since $(1-p)^2\le p$ for $p>\frac12$, this proves \pref{eq:complement-coordinate-stability}.
\end{proof}

The next lemma bounds one step of the master's update in \pref{line:strong-x-update}.

\begin{lemma}
\label[lemma]{lem:strong-master-step}
If $\theta\le\frac18$, then for every $u\in\mathcal X$,
\begin{align}
\ip{x_t-u}{\widetilde g_t}
&\le D_{\Psi_t}(u,x_t)-D_{\Psi_t}(u,x_{t+1})
+2\theta\mathcal V_t,\label{eq:strong-master-step-bound}\\
\Et[\mathcal V_t]&\le2\ip{x_t}{g_t},
\label{eq:strong-master-step}
\end{align}
where $\mathcal V_t\triangleq\ind\{0\in\Bcal\}x_{t,0}^2\widetilde g_{t,0}^2+\sum_{i\in\Vbarself}\min\{x_{t,i},(1-x_{t,i})^2\}\widetilde g_{t,i}^2$.
\end{lemma}

\begin{proof}
Since $x_{t+1}$ minimizes $\ip{x}{\widetilde g_t}+D_{\Psi_t}(x,x_t)$ over $x\in\mathcal X$ by \pref{line:strong-x-update}, the first-order optimality condition gives
\begin{align}
\ip{\widetilde g_t+\nabla\Psi_t(x_{t+1})-\nabla\Psi_t(x_t)}{u-x_{t+1}}\ge0.
\label{eq:strong-first-order}
\end{align}
By the definition of the Bregman divergence, direct calculation gives the three-point identity
\begin{align}
\ip{\nabla\Psi_t(x_{t+1})-\nabla\Psi_t(x_t)}{u-x_{t+1}}
=D_{\Psi_t}(u,x_t)-D_{\Psi_t}(u,x_{t+1})-D_{\Psi_t}(x_{t+1},x_t).
\label{eq:strong-three-point}
\end{align}
Plugging \pref{eq:strong-three-point} into \pref{eq:strong-first-order} and adding $\ip{x_t-x_{t+1}}{\widetilde g_t}$ to both sides, we have
\begin{align}
\ip{x_t-u}{\widetilde g_t}
\le D_{\Psi_t}(u,x_t)-D_{\Psi_t}(u,x_{t+1})
+\ip{x_t-x_{t+1}}{\widetilde g_t}
-D_{\Psi_t}(x_{t+1},x_t).
\label{eq:strong-omd-step}
\end{align}
To bound $\ip{x_t-x_{t+1}}{\widetilde g_t}-D_{\Psi_t}(x_{t+1},x_t)$ in \pref{eq:strong-omd-step}, we expand $D_{\Psi_t}(x_{t+1},x_t)$ by \pref{eq:strong-regularizer}, group the terms of meta-action $0$ and of each $i\in\Vbarself$, and replace $x_{t+1,0}$ and $x_{t+1,i}$ by a supremum over $y\in(0,1)$:
\begin{align}
&\ip{x_t-x_{t+1}}{\widetilde g_t}-D_{\Psi_t}(x_{t+1},x_t)\nonumber\\
&=\ind\{0\in\Bcal\}\left((x_{t,0}-x_{t+1,0})\widetilde g_{t,0}-\frac1{\theta_{t,0}}D_{\varphi_0}(x_{t+1,0},x_{t,0})\right)\nonumber\\
&\qquad+\sum_{i\in\Vbarself}\left((x_{t,i}-x_{t+1,i})\widetilde g_{t,i}-\frac1\theta D_{\varphi}(x_{t+1,i},x_{t,i})\right)\nonumber\\
&\le\ind\{0\in\Bcal\}\sup_{0<y<1}\left\{(x_{t,0}-y)\widetilde g_{t,0}-\frac1{\theta_{t,0}}D_{\varphi_0}(y,x_{t,0})\right\}\nonumber\\
&\qquad+\sum_{i\in\Vbarself}\sup_{0<y<1}\left\{(x_{t,i}-y)\widetilde g_{t,i}-\frac1\theta D_{\varphi}(y,x_{t,i})\right\}\nonumber\\
&\le\frac{\theta_{t,0}}2\ind\{0\in\Bcal\}x_{t,0}^2\widetilde g_{t,0}^2+\theta\sum_{i\in\Vbarself}\min\{x_{t,i},(1-x_{t,i})^2\}\widetilde g_{t,i}^2,
\label{eq:strong-coordinate-bound}
\end{align}
where the first inequality uses $x_{t+1,0},x_{t+1,i}\in(0,1)$, and the second uses \pref{eq:block-coordinate-stability} with $a=\theta_{t,0}$ for meta-action $0$, and \pref{eq:complement-coordinate-stability} for $i\in\Vbarself$, whose conditions hold since $\theta\le\frac18$ and $\widetilde g_{t,i}=\frac{\ell_{t,i}\ind\{I_t\ne i\}}{1-x_{t,i}}\le\frac1{1-x_{t,i}}$.

It remains to bound $\theta_{t,0}$ in \pref{eq:strong-coordinate-bound}. Recall from \pref{alg:strong} that $J=\lceil\log_2(2MT)\rceil$ and $\sigma=e^{1/J}$, and that by \pref{line:strong-threshold}, $\theta_{t,0}$ equals $\theta$ multiplied by $\sigma$ once for each round $s<t$ with $\rho_{s+1}\ne\rho_s$. For each such round $s$, we have $\rho_{s+1}=\frac2{x_{s+1,0}}>2\rho_s$. Since $\rho_1=2$ and $\rho_{s+1}=\frac2{x_{s+1,0}}\le4MT$ for each such round $s$, where the inequality uses $x_{s+1,0}\ge\frac1{2MT}$ on $\mathcal X$, there are fewer than $\log_2(2MT)$ such rounds, and hence at most $J$. Therefore, $\theta_{t,0}\le\sigma^J\theta=e\theta$, and the right-hand side of \pref{eq:strong-coordinate-bound} is at most $2\theta\mathcal V_t$. Plugging this bound into \pref{eq:strong-omd-step} proves the first claim.

For the second claim, by \pref{eq:strong-master-estimator}, we have
\begin{align*}
\Et[x_{t,0}^2\widetilde g_{t,0}^2]
&=x_{t,0}\ip{q_t}{\ell_t^2}\le x_{t,0}\ip{q_t}{\ell_t}=x_{t,0}g_{t,0},\\*
\min\{x_{t,i},(1-x_{t,i})^2\}\Et[\widetilde g_{t,i}^2]
&=\min\left\{\frac{x_{t,i}}{1-x_{t,i}},1-x_{t,i}\right\}\ell_{t,i}^2
\le2x_{t,i}\ell_{t,i}=2x_{t,i}g_{t,i},
\end{align*}
where both inequalities use $\ell_t\in[0,1]^K$, and the inequality in the second line also uses $\frac{x_{t,i}}{1-x_{t,i}}\le2x_{t,i}$ when $x_{t,i}\le\frac12$ and $1-x_{t,i}\le2x_{t,i}$ when $x_{t,i}>\frac12$. Taking the summation of the first line multiplied by $\ind\{0\in\Bcal\}$ and the second line over $i\in\Vbarself$ gives $\Et[\mathcal V_t]\le2\ip{x_t}{g_t}$.
\end{proof}

The next lemma bounds the initial divergence, the smoothing of the comparator and the changes of the regularizer.

\begin{lemma}
\label[lemma]{lem:strong-barrier-reward}
For $b\in\Bcal$, let $u^{(b)}\triangleq\left(1-\frac{M-1}{2MT}\right)e_b+\frac1{2MT}\sum_{j\in\Bcal\setminus\{b\}}e_j\in\mathcal X$. Then $D_{\Psi_1}(u^{(b)},x_1)\le\frac{{8}\ln(MT)}{\theta}$ and $\sum_{t=1}^{T}\ip{u^{(b)}-e_b}{g_t}\le\frac12$. Moreover, pathwise, for every $b\in\Bcal$,
\begin{align}
\sum_{t=1}^{T-1}\left[D_{\Psi_{t+1}}(u^{(b)},x_{t+1})-D_{\Psi_t}(u^{(b)},x_{t+1})\right]\le0,
\label{eq:strong-barrier-zero}
\end{align}
and if $0\in\Bcal$, then for $b=0$, the following stronger bound holds:
\begin{align}
\sum_{t=1}^{T-1}\left[D_{\Psi_{t+1}}(u^{(0)},x_{t+1})-D_{\Psi_t}(u^{(0)},x_{t+1})\right]
\le-\frac{\bar\rho}{16e^2\theta J}+\frac1{e^2\theta J}.
\label{eq:strong-barrier-reward}
\end{align}
\end{lemma}

\begin{proof}
To bound $D_{\Psi_1}(u^{(b)},x_1)$, by \pref{eq:strong-regularizer} and $\theta_{1,0}=\theta$, we have
\begin{align}
D_{\Psi_1}(u^{(b)},x_1)
=\frac1\theta\left(\ind\{0\in\Bcal\}D_{\varphi_0}(u^{(b)}_0,x_{1,0})+\sum_{i\in\Vbarself}D_\varphi(u^{(b)}_i,x_{1,i})\right).
\label{eq:strong-initial-split}
\end{align}
To bound $D_{\varphi_0}(u^{(b)}_0,x_{1,0})$ in \pref{eq:strong-initial-split}, direct calculation gives
\begin{align*}
D_{\varphi_0}(u^{(b)}_0,x_{1,0})
=\frac{u^{(b)}_0}{x_{1,0}}-1+\ln\frac{x_{1,0}}{u^{(b)}_0}
\le1+\ln(2MT),
\end{align*}
where the inequality uses $x_{1,0}\ge\frac12$ and $\frac1{2MT}\le u^{(b)}_0\le1$. To bound $\sum_{i\in\Vbarself}D_\varphi(u^{(b)}_i,x_{1,i})$ in \pref{eq:strong-initial-split}, direct calculation gives
\begin{align}
\sum_{i\in\Vbarself}D_\varphi(u^{(b)}_i,x_{1,i})
=\sum_{i\in\Vbarself}\left(u^{(b)}_i\ln\frac{u^{(b)}_i}{x_{1,i}}-u^{(b)}_i+x_{1,i}\right)
+\sum_{i\in\Vbarself}\frac{x_{1,i}-u^{(b)}_i}{1-x_{1,i}}
+\sum_{i\in\Vbarself}\ln\frac{1-x_{1,i}}{1-u^{(b)}_i}.
\label{eq:strong-initial-phi}
\end{align}
Recall that $x_{1,i}=\frac1{2M}$ for $i\in\Vbarself$ and that $|\Vbarself|\le M$. We bound the three sums on the right-hand side of \pref{eq:strong-initial-phi} as follows:
\begin{align*}
\sum_{i\in\Vbarself}\left(u^{(b)}_i\ln\frac{u^{(b)}_i}{x_{1,i}}-u^{(b)}_i+x_{1,i}\right)
&\le\ln(2M)\sum_{i\in\Vbarself}u^{(b)}_i+\sum_{i\in\Vbarself}x_{1,i}\le\ln(2M)+1,\\
\sum_{i\in\Vbarself}\frac{x_{1,i}-u^{(b)}_i}{1-x_{1,i}}
&\le\sum_{i\in\Vbarself}\frac{x_{1,i}}{1-x_{1,i}}\le1,\\
\sum_{i\in\Vbarself}\ln\frac{1-x_{1,i}}{1-u^{(b)}_i}
&\le\sum_{i\in\Vbarself\setminus\{b\}}\ln\frac{1}{1-\frac1{2MT}}+\ind\{b\in\Vbarself\}\ln\frac{2MT}{M-1}
\le1+\ln(2MT),
\end{align*}
where the third uses $u^{(b)}_i=\frac1{2MT}$ for $i\ne b$, $u^{(b)}_b=1-\frac{M-1}{2MT}$, and $\ln\frac1{1-z}\le\frac z{1-z}$ for $z=\frac1{2MT}$. Plugging these bounds into \pref{eq:strong-initial-phi}, and then \pref{eq:strong-initial-phi} and the bound on $D_{\varphi_0}(u^{(b)}_0,x_{1,0})$ into \pref{eq:strong-initial-split}, gives
\begin{align*}
D_{\Psi_1}(u^{(b)},x_1)\le\frac{4+\ln(2M)+2\ln(2MT)}{\theta}\le\frac{{8}\ln(MT)}{\theta},
\end{align*}
{where the last inequality uses $T\ge3$.}
To bound $\sum_{t=1}^{T}\ip{u^{(b)}-e_b}{g_t}$, note that $\ip{u^{(b)}-e_b}{g_t}\le\frac{M-1}{2MT}\le\frac1{2T}$ for every $t\in[T]$, since $g_t\in[0,1]^{\Bcal}$. Taking the summation over $t\in[T]$ gives $\sum_{t=1}^{T}\ip{u^{(b)}-e_b}{g_t}\le\frac12$.

To prove \pref{eq:strong-barrier-zero}, note that by \pref{eq:strong-regularizer}, $\Psi_{t+1}-\Psi_t=\left(\frac1{\theta_{t+1,0}}-\frac1{\theta_{t,0}}\right)\varphi_0$, so for every $t\in[T-1]$,
\begin{align}
D_{\Psi_{t+1}}(u^{(b)},x_{t+1})-D_{\Psi_t}(u^{(b)},x_{t+1})
=\left(\frac1{\theta_{t+1,0}}-\frac1{\theta_{t,0}}\right)D_{\varphi_0}(u^{(b)}_0,x_{t+1,0})\le0,
\label{eq:strong-coefficient-change}
\end{align}
where the inequality uses $\theta_{t+1,0}\ge\theta_{t,0}$. Summing \pref{eq:strong-coefficient-change} over $t\in[T-1]$ proves \pref{eq:strong-barrier-zero}.

To prove \pref{eq:strong-barrier-reward}, first suppose that $\rho_{t+1}=\rho_t$ for every $t\in[T-1]$. Then the left-hand side of \pref{eq:strong-barrier-reward} is zero by \pref{eq:strong-coefficient-change}, while $\bar\rho\le\rho_1=2$ implies that the right-hand side of \pref{eq:strong-barrier-reward} is positive, so \pref{eq:strong-barrier-reward} holds. Otherwise, let $t'$ be the last round with $\rho_{t'+1}\ne\rho_{t'}$. By \pref{line:strong-threshold}, $\rho_{t'+1}=\frac2{x_{t'+1,0}}$, $\theta_{t'+1,0}=\sigma\theta_{t',0}$, and $\frac1{x_{t,0}}\le\rho_t\le\rho_{t'+1}$ for every $t\in[T]$, so $\frac1{x_{t'+1,0}}\ge\frac{\bar\rho}2$. Since every term of the sum on the left-hand side of \pref{eq:strong-barrier-reward} is nonpositive by \pref{eq:strong-coefficient-change}, keeping only the term $t=t'$ gives
\begin{align*}
&\sum_{t=1}^{T-1}\left[D_{\Psi_{t+1}}(u^{(0)},x_{t+1})-D_{\Psi_t}(u^{(0)},x_{t+1})\right]\\
&\le\left(\frac1{\sigma\theta_{t',0}}-\frac1{\theta_{t',0}}\right)D_{\varphi_0}(u^{(0)}_0,x_{t'+1,0})
\le-\frac1{e^2\theta J}\left(\frac{u^{(0)}_0}{x_{t'+1,0}}-1-\ln\frac{u^{(0)}_0}{x_{t'+1,0}}\right)\\
&\le-\frac1{e^2\theta J}\left(\frac{u^{(0)}_0}{2x_{t'+1,0}}-1\right)
\le-\frac1{e^2\theta J}\left(\frac{\bar\rho}8-1\right)
\le-\frac{\bar\rho}{16e^2\theta J}+\frac1{e^2\theta J},
\end{align*}
where the second inequality uses $\frac{\sigma-1}{\sigma\theta_{t',0}}\ge\frac1{e^2\theta J}$, which holds since $\sigma-1\ge\frac1J$, $\sigma\le e$ and $\theta_{t',0}\le e\theta$ by the proof of \pref{lem:strong-master-step}, the third uses $\ln z\le\frac z2$, and the fourth uses $u^{(0)}_0\ge\frac12$ and $\frac1{x_{t'+1,0}}\ge\frac{\bar\rho}2$.
\end{proof}

We are now ready to prove \pref{lem:strong-components} by combining \pref{lem:scaled-stability-base}, \pref{lem:strong-master-step} and \pref{lem:strong-barrier-reward}.

\begin{proof}[Proof of \pref{lem:strong-components}.]
To prove \pref{eq:strong-base-ledger}, fix $k\in\Vself$. Applying \pref{lem:scaled-stability-base} with $\mathcal I=[T]$ and $\tau=1$, we have $R_{[T],k}=\Reg^{\mathrm{base}}_{T,k}$ and $r_{[T]}=\E[\bar\rho]$, so \pref{eq:scaled-base-bound} gives
\begin{align*}
&\Reg^{\mathrm{base}}_{T,k}\\
&\le(56A\ln(sT)+6J)\E[\bar\rho]+8\sqrt{2A\ln(sT)\E[\bar\rho]L_{T,k}}
\le74A\ln(KT)\E[\bar\rho]+8\sqrt{2A\ln(KT)\E[\bar\rho]L_{T,k}},
\end{align*}
where the last inequality uses $s\le K$, $A\ge1$, and $J\le3\ln(KT)$, which holds since $M\le K$.

To prove \pref{eq:strong-master-ledger}, fix $b\in\Bcal$. Summing \pref{eq:strong-master-step-bound} over $t\in[T]$ with $u=u^{(b)}$, defined in \pref{lem:strong-barrier-reward}, we have
\begin{align*}
&\sum_{t=1}^T\ip{x_t-u^{(b)}}{\widetilde g_t}\\
&\le\sum_{t=1}^T\left(D_{\Psi_t}(u^{(b)},x_t)-D_{\Psi_t}(u^{(b)},x_{t+1})\right)+2\theta\sum_{t=1}^T\mathcal V_t\\
&=D_{\Psi_1}(u^{(b)},x_1)-D_{\Psi_T}(u^{(b)},x_{T+1})+\sum_{t=1}^{T-1}\left(D_{\Psi_{t+1}}(u^{(b)},x_{t+1})-D_{\Psi_t}(u^{(b)},x_{t+1})\right)+2\theta\sum_{t=1}^T\mathcal V_t.
\end{align*}
We take expectations of both sides, using $\Et[\widetilde g_t]=g_t$ from \pref{eq:strong-master-estimator}, $\Et[\mathcal V_t]\le2\ip{x_t}{g_t}$ from \pref{eq:strong-master-step}, and $D_{\Psi_T}(u^{(b)},x_{T+1})\ge0$. Adding $\E[\sum_{t=1}^T\ip{u^{(b)}-e_b}{g_t}]$ to both sides, we have
\begin{align*}
\Reg^{\mathrm{master}}_{T,b}
\le&D_{\Psi_1}(u^{(b)},x_1)
+4\theta\E\left[\sum_{t=1}^{T}\ip{x_t}{g_t}\right]
+\E\left[\sum_{t=1}^{T}\ip{u^{(b)}-e_b}{g_t}\right]\\
&+\E\left[\sum_{t=1}^{T-1}\bigl(D_{\Psi_{t+1}}(u^{(b)},x_{t+1})
-D_{\Psi_t}(u^{(b)},x_{t+1})\bigr)\right].
\end{align*}
By \pref{lem:strong-barrier-reward}, $D_{\Psi_1}(u^{(b)},x_1)\le\frac{{8}\ln(MT)}{\theta}$, $\sum_{t=1}^T\ip{u^{(b)}-e_b}{g_t}\le\frac12$, and by \pref{eq:strong-barrier-zero} and \pref{eq:strong-barrier-reward}, $\E[\sum_{t=1}^{T-1}(D_{\Psi_{t+1}}(u^{(b)},x_{t+1})-D_{\Psi_t}(u^{(b)},x_{t+1}))]\le-\frac{\E[\bar\rho]}{16e^2\theta J}\ind\{b=0\}+\frac1{e^2\theta J}$. Therefore,
\begin{align*}
\Reg^{\mathrm{master}}_{T,b}
\le\frac{{8}\ln(MT)}{\theta}+\frac12+\frac1{e^2\theta J}+4\theta\E\left[\sum_{t=1}^{T}\ip{x_t}{g_t}\right]-\frac{\E[\bar\rho]}{16e^2\theta J}\ind\{b=0\},
\end{align*}
which gives \pref{eq:strong-master-ledger} since $\frac12+\frac1{e^2\theta J}\le\frac{\ln(MT)}{\theta}$ for $\theta\le\frac1{8e}$, $J\ge1$ and $MT\ge2$.
\end{proof}

We now prove \pref{thm:strong-fixed} by combining the two bounds of \pref{lem:strong-components}.

\begin{proof}[Proof of \pref{thm:strong-fixed}.]
Choose $c=\frac1{2^{13}e^2}$. {Since $\Gamma_A=AJ\ln(KT)\ge1$, we have $\theta\le\frac c{\Gamma_A}\le c\le\frac1{8e}$, so \pref{lem:strong-components} applies.} We first consider $k\in\Vself$. Plugging \pref{eq:strong-base-ledger} and \pref{eq:strong-master-ledger} with $b=0$ into \pref{eq:block-decomp-S}, and using $\E[\sum_{t=1}^T\ip{x_t}{g_t}]=L_{T,k}+\Reg_{T,k}$, we have
\begin{align}
\Reg_{T,k}
\le\frac{{9}\ln(MT)}{\theta}+12\theta(L_{T,k}+\Reg_{T,k})
-\frac{\E[\bar\rho]}{16e^2\theta J}
+74A\ln(KT)\E[\bar\rho]+8\sqrt{2A\ln(KT)\E[\bar\rho]L_{T,k}}.
\label{eq:strong-fixed-combined}
\end{align}
To bound the last two terms in \pref{eq:strong-fixed-combined}, we have
\begin{align}
&74A\ln(KT)\E[\bar\rho]+8\sqrt{2A\ln(KT)\E[\bar\rho]L_{T,k}}\nonumber\\
&\le\frac{\E[\bar\rho]}{64e^2\theta J}+\left(\frac{\E[\bar\rho]}{64e^2\theta J}+2^{11}e^2\Gamma_A\theta L_{T,k}\right)
=\frac{\E[\bar\rho]}{32e^2\theta J}+2^{11}e^2\Gamma_A\theta L_{T,k},
\label{eq:adaptive-base-cancellation}
\end{align}
where the inequality bounds the two terms separately. For $74A\ln(KT)\E[\bar\rho]$, we use $64e^2\theta J\cdot74A\ln(KT)=4736e^2\Gamma_A\theta\le4736e^2c\le1$, which holds since $\Gamma_A=AJ\ln(KT)$, $\theta\le\frac c{\Gamma_A}$. For $8\sqrt{2A\ln(KT)\E[\bar\rho]L_{T,k}}$, we use the AM--GM inequality.
Plugging \pref{eq:adaptive-base-cancellation} into \pref{eq:strong-fixed-combined}, the terms $-\frac{\E[\bar\rho]}{16e^2\theta J}+\frac{\E[\bar\rho]}{32e^2\theta J}$ sum to $-\frac{\E[\bar\rho]}{32e^2\theta J}\le0$, so
\begin{align}
\Reg_{T,k}\le\frac{{9}\ln(MT)}{\theta}+12\theta(L_{T,k}+\Reg_{T,k})+\order(\Gamma_A\theta L_{T,k}).
\label{eq:strong-fixed-rearrange}
\end{align}
Since $\theta\le\frac c{\Gamma_A}\le c=\frac1{2^{13}e^2}$, we have $12\theta\le\frac12$, so rearranging \pref{eq:strong-fixed-rearrange} gives $\Reg_{T,k}\le\order(\frac{\ln(MT)}{\theta}+\Gamma_A\theta L_{T,k})$. For $k\in\Vbarself$, by \pref{eq:block-decomp-U} and \pref{eq:strong-master-ledger} with $b=k$, \pref{eq:strong-fixed-rearrange} holds without the term $\order(\Gamma_A\theta L_{T,k})$, and rearranging in the same way gives the claimed bound.
Finally, with $\theta=\min\{\frac c{\Gamma_A},\sqrt{\frac{\ln(MT)}{\Gamma_AL^\star}}\}$, we have $\frac{\ln(MT)}{\theta}\le\frac{\Gamma_A\ln(MT)}{c}+\sqrt{\Gamma_A\ln(MT)L^\star}$ and $\Gamma_A\theta L^\star\le\sqrt{\Gamma_A\ln(MT)L^\star}$, which gives $\Reg_T=\order(\sqrt{\Gamma_A\ln(MT)L^\star}+\Gamma_A\ln(MT))$, and $A=\order(\alpha\ln(KT))$ by \pref{lem:soft-ratio} gives $\Reg_T=\otil(\sqrt{\alpha L^\star}+\alpha)$.
\end{proof}

\subsection{Doubling Trick}
\label{app:strong-adaptive}

In this section, we show a parameter-free version of \pref{alg:strong} that achieves $\otil(\sqrt{\alpha L^\star}+\alpha)$ regret without knowing $\alpha$ or $L^\star$. Specifically, we propose \pref{alg:strong-adaptive}, a doubling-trick variant of \pref{alg:strong} that follows \pref{alg:u-adaptive}: it maintains a guess for the bound $A$ in \pref{thm:strong-fixed} and the cumulative observable loss $S$ since the last restart. Each instance of \pref{alg:strong} defines an epoch with a fixed guess and master learning rate. Consecutive epochs with the same guess form a meta-epoch. Denote by $m$ the meta-epoch index, by $\lambda$ the epoch index within it, by $A_m$ the guess in meta-epoch $m$, and by $\theta_{m,\lambda}$ the master learning rate in epoch $(m,\lambda)$. Once $\mathcal H^{\Gself}(q_t)>A_m$ (\pref{line:strong-adaptive-check}), we double the guess and start meta-epoch $m+1$ at $\lambda=1$ with the master learning rate $\frac c{\Gamma_{A_{m+1}}}$. Similar to \pref{alg:u-adaptive}, a meta-epoch may contain no rounds, which happens when $\mathcal H^{\Gself}(q_t)>A_{m+1}$ still holds right after a doubling. Compared with \pref{alg:u-adaptive}, the main change is the loss that triggers the learning-rate restarts. Since the played action may have no self-loop, its loss may not be observed, so we accumulate the observable loss $\widetilde Z_t=\ip{x_t}{\widetilde g_t}$, defined in \pref{sec:strong-oracle}, in $S$ instead (\pref{line:strong-adaptive-S}). Once $\Gamma_{A_m}\theta_{m,\lambda}^2S\ge\ln(MT)$, we halve the master learning rate and start epoch $\lambda+1$ within meta-epoch $m$ (\pref{line:strong-adaptive-halve}). At the start of each epoch, we initialize \pref{alg:strong} with $A_m$ and $\theta_{m,\lambda}$ and set $S=0$. The complete pseudocode is shown in \pref{alg:strong-adaptive}, and \pref{thm:strong-adaptive} establishes the claimed regret guarantee.

\begin{algorithm2e}[t]
\caption{Doubling trick with \pref{alg:strong} as an oracle}
\label{alg:strong-adaptive}
\KwIn{strongly observable graph $G$, horizon $T$}
Set $m=1$, $\lambda=1$, $A_1=1$, $\theta_{1,1}=\frac c{\Gamma_{A_1}}$ and $S=0$\;
Start meta-epoch $m$ and epoch $(m,\lambda)$ by initializing an instance of \pref{alg:strong} with $(G,T,A_m,\theta_{m,\lambda})$\;
\For{$t=1,\ldots,T$}{
  \If{$0\in\Bcal$}{
    Compute $q_t$ by \pref{line:strong-fixed-point} of the instance of \pref{alg:strong} for epoch $(m,\lambda)$\;
    \While{$\mathcal H^{\Gself}(q_t)>A_m$\label{line:strong-adaptive-check}}{
      $A_{m+1}=2A_m$, $\theta_{m+1,1}=\frac c{\Gamma_{A_{m+1}}}$, $m\leftarrow m+1$, $\lambda\leftarrow1$ and $S\leftarrow0$\;
      Start meta-epoch $m$ and epoch $(m,1)$ by initializing a new instance of \pref{alg:strong} with $(G,T,A_m,\theta_{m,1})$, starting at round $t$\;
      Compute $q_t$ by \pref{line:strong-fixed-point} of this new instance\;
    }
  }
  Execute round $t$ of the instance of \pref{alg:strong} for epoch $(m,\lambda)$, and denote by $x_t$ its master distribution and by $\widetilde g_t$ its loss estimate in \pref{line:strong-master-est}\;
  Set $\widetilde Z_t=\ip{x_t}{\widetilde g_t}$ and $S\leftarrow S+\widetilde Z_t$\label{line:strong-adaptive-S}\;
  \If{$t<T$ and $\Gamma_{A_m}\theta_{m,\lambda}^2S\ge\ln(MT)$\label{line:strong-adaptive-halve}}{
    $\theta_{m,\lambda+1}=\frac{\theta_{m,\lambda}}2$, $\lambda\leftarrow\lambda+1$ and $S\leftarrow0$\;
    Start epoch $(m,\lambda)$ by initializing a new instance of \pref{alg:strong} with $(G,T,A_m,\theta_{m,\lambda})$, starting at round $t+1$\;
  }
}
\end{algorithm2e}

Unlike the incurred loss in \pref{alg:u-adaptive}, the observable loss of the last round of an epoch is not bounded by $1$.
We control it by the following pathwise bounds.

\begin{lemma}
\label[lemma]{lem:observable-loss-bounds}
For every round $t$, $\Et[\widetilde Z_t]=\ip{x_t}{g_t}$ and $0\le\mathcal V_t\le\min\{2\widetilde Z_t,5\}$, where $\mathcal V_t$ is defined in \pref{lem:strong-master-step}.
If $0\in\Bcal$, then also $\widetilde Z_t\le3+\frac1{x_{t,0}}$.
\end{lemma}

\begin{proof}
The first claim follows from $\Et[\widetilde g_t]=g_t$.
For the other claims, \pref{eq:strong-master-estimator} and $\ell_t\in[0,1]^K$ give $x_{t,0}^2\widetilde g_{t,0}^2=\ell_{t,I_t}^2\ind\{B_t=0\}\le x_{t,0}\widetilde g_{t,0}\le1$, $\widetilde g_{t,i}\le\frac1{1-x_{t,i}}$ for $i\in\Vbarself$, and
\begin{align*}
\min\{x_{t,i},(1-x_{t,i})^2\}\widetilde g_{t,i}^2
\le\min\left\{\frac1{1-x_{t,i}},\frac{1-x_{t,i}}{x_{t,i}}\right\}
x_{t,i}\widetilde g_{t,i}
\le2x_{t,i}\widetilde g_{t,i},
\end{align*}
where the last inequality uses $\frac1{1-x_{t,i}}\le2$ if $x_{t,i}\le\frac12$ and $\frac{1-x_{t,i}}{x_{t,i}}\le1$ otherwise. Taking the summation over $i\in\Vbarself$ and adding $x_{t,0}^2\widetilde g_{t,0}^2\le x_{t,0}\widetilde g_{t,0}$ gives $\mathcal V_t\le2\widetilde Z_t$, since $\widetilde Z_t=\ind\{0\in\Bcal\}x_{t,0}\widetilde g_{t,0}+\sum_{i\in\Vbarself}x_{t,i}\widetilde g_{t,i}$.

To prove $\mathcal V_t\le5$, note that for $i\in\Vbarself$, $\widetilde g_{t,i}\le\frac1{1-x_{t,i}}$ gives $\min\{x_{t,i},(1-x_{t,i})^2\}\widetilde g_{t,i}^2\le\min\{\frac{x_{t,i}}{(1-x_{t,i})^2},1\}$, which is at most $4x_{t,i}$ if $x_{t,i}\le\frac12$ and at most $1$ otherwise. Since at most one $j\in\Vbarself$ has $x_{t,j}>\frac12$, in which case $\sum_{i\in\Vbarself\setminus\{j\}}x_{t,i}<\frac12$, taking the summation over $i\in\Vbarself$ gives
\begin{align*}
\sum_{i\in\Vbarself}\min\{x_{t,i},(1-x_{t,i})^2\}\widetilde g_{t,i}^2
\le\max\left\{4\sum_{i\in\Vbarself}x_{t,i},\,1+4\cdot\frac12\right\}\le4.
\end{align*}
Together with $x_{t,0}^2\widetilde g_{t,0}^2\le1$, we have $\mathcal V_t\le5$.

Finally, to prove $\widetilde Z_t\le3+\frac1{x_{t,0}}$, we have
\begin{align*}
\widetilde Z_t
&=x_{t,0}\widetilde g_{t,0}+\sum_{i\in\Vbarself}x_{t,i}\widetilde g_{t,i}
\le1+\sum_{i\in\Vbarself:\,x_{t,i}\le\frac12}2x_{t,i}+\sum_{i\in\Vbarself:\,x_{t,i}>\frac12}\frac1{1-x_{t,i}}
\le3+\frac1{x_{t,0}},
\end{align*}
where the first inequality uses $x_{t,0}\widetilde g_{t,0}\le1$ and $x_{t,i}\widetilde g_{t,i}\le\frac{x_{t,i}}{1-x_{t,i}}$, which is at most $2x_{t,i}$ if $x_{t,i}\le\frac12$ and at most $\frac1{1-x_{t,i}}$ otherwise, and the second uses $\sum_{i\in\Vbarself}x_{t,i}\le1$, that at most one $i\in\Vbarself$ has $x_{t,i}>\frac12$, and $1-x_{t,i}\ge x_{t,0}$.
\end{proof}

Next, we first bound the regret of each epoch, conditionally on its start.

\begin{lemma}
\label[lemma]{lem:one-loss-epoch}
For an epoch $(m,\lambda)$ of \pref{alg:strong-adaptive}, denote by $\mathcal T_{m,\lambda}$ its rounds, by $\tstart_{m,\lambda}$ and $\tend_{m,\lambda}$ its first and last rounds, and by
$\Reg_{m,\lambda,k}\triangleq\E_{\tstart_{m,\lambda}}[\sum_{t\in\mathcal T_{m,\lambda}}(\ip{x_t}{g_t}-\ell_{t,k})]$ its expected regret conditional on the history before $\tstart_{m,\lambda}$.
Then, for every $k\in V$,
\begin{align}
\Reg_{m,\lambda,k}
\le\order\left(\frac{\ln(MT)}{\theta_{m,\lambda}}\right).
\label{eq:one-loss-epoch}
\end{align}
\end{lemma}

\begin{proof}
Abbreviate $A=A_m$ and $\theta=\theta_{m,\lambda}$, and assume $\Reg_{m,\lambda,k}>0$.
Since whether $t\in\mathcal T_{m,\lambda}$ is determined before round $t$ is sampled, \pref{lem:observable-loss-bounds} and the tower property give
\begin{align}
\E_{\tstart_{m,\lambda}}\left[\sum_{t\in\mathcal T_{m,\lambda}}\widetilde Z_t\right]
=\E_{\tstart_{m,\lambda}}\left[\sum_{t\in\mathcal T_{m,\lambda}}\ip{x_t}{g_t}\right],
\label{eq:stopped-unbiasedness}
\end{align}
and similarly, $\E_{\tstart_{m,\lambda}}[\sum_{t\in\mathcal T_{m,\lambda}}\widetilde g_{t,i}]=\E_{\tstart_{m,\lambda}}[\sum_{t\in\mathcal T_{m,\lambda}}\ell_{t,i}]$ for $i\in\Vbarself$.
By \pref{alg:strong-adaptive}, $\Gamma_A\theta^2S<\ln(MT)$ after every round $t<\tend_{m,\lambda}$ of the epoch, so
\begin{align}
\sum_{\substack{t\in\mathcal T_{m,\lambda}\\t<\tend_{m,\lambda}}}\widetilde Z_t
\le\frac{\ln(MT)}{\Gamma_A\theta^2}.
\label{eq:strong-epoch-loss}
\end{align}

For $k\in\Vself$, denote by $\bar\rho_{m,\lambda}\triangleq\max_{t\in\mathcal T_{m,\lambda}}\frac1{x_{t,0}}$ the largest inverse probability of meta-action $0$ in the epoch. Since $g_{t,0}=\ip{q_t}{\ell_t}$, we have
\begin{align}
\Reg_{m,\lambda,k}
=\E_{\tstart_{m,\lambda}}\left[\sum_{t\in\mathcal T_{m,\lambda}}\left(\ip{x_t}{g_t}-g_{t,0}\right)\right]+\E_{\tstart_{m,\lambda}}\left[\sum_{t\in\mathcal T_{m,\lambda}}\left(\ip{q_t}{\ell_t}-\ell_{t,k}\right)\right],
\label{eq:strong-epoch-split}
\end{align}
and we bound the two terms on the right-hand side of \pref{eq:strong-epoch-split} separately. Since \pref{alg:strong-adaptive} ensures $\mathcal H^{\Gself}(q_t)\le A$ for every $t\in\mathcal T_{m,\lambda}$, \pref{lem:scaled-stability-base} applies with $\mathcal I=\mathcal T_{m,\lambda}$, and the same steps as in the proof of \pref{eq:strong-base-ledger} give
\begin{align*}
&\E_{\tstart_{m,\lambda}}\left[\sum_{t\in\mathcal T_{m,\lambda}}\left(\ip{q_t}{\ell_t}-\ell_{t,k}\right)\right]\\
&\le74A\ln(KT)\E_{\tstart_{m,\lambda}}[\bar\rho_{m,\lambda}]+8\sqrt{2A\ln(KT)\E_{\tstart_{m,\lambda}}[\bar\rho_{m,\lambda}]\E_{\tstart_{m,\lambda}}\left[\sum_{t\in\mathcal T_{m,\lambda}}\ell_{t,k}\right]}.
\end{align*}
Applying the proof of \pref{eq:strong-master-ledger} with $b=0$ to the instance of \pref{alg:strong} for this epoch, conditionally on the history before $\tstart_{m,\lambda}$, gives
\begin{align*}
\E_{\tstart_{m,\lambda}}\left[\sum_{t\in\mathcal T_{m,\lambda}}\left(\ip{x_t}{g_t}-g_{t,0}\right)\right]\le\frac{{9}\ln(MT)}{\theta}+12\theta\E_{\tstart_{m,\lambda}}\left[\sum_{t\in\mathcal T_{m,\lambda}}\ip{x_t}{g_t}\right]-\frac{\E_{\tstart_{m,\lambda}}[\bar\rho_{m,\lambda}]}{16e^2\theta J}.
\end{align*}
Plugging the two bounds into \pref{eq:strong-epoch-split} and applying \pref{eq:adaptive-base-cancellation} with $\E_{\tstart_{m,\lambda}}[\bar\rho_{m,\lambda}]$ in place of $\E[\bar\rho]$ and $\E_{\tstart_{m,\lambda}}[\sum_{t\in\mathcal T_{m,\lambda}}\ell_{t,k}]$ in place of $L_{T,k}$, we have
\begin{align}
\Reg_{m,\lambda,k}&\le\frac{{9}\ln(MT)}{\theta}+12\theta\E_{\tstart_{m,\lambda}}\left[\sum_{t\in\mathcal T_{m,\lambda}}\ip{x_t}{g_t}\right]+2^{11}e^2\Gamma_A\theta\E_{\tstart_{m,\lambda}}\left[\sum_{t\in\mathcal T_{m,\lambda}}\ell_{t,k}\right]-\frac{\E_{\tstart_{m,\lambda}}[\bar\rho_{m,\lambda}]}{32e^2\theta J}\nonumber\\
&\le\frac{C_0\ln(MT)}{\theta}+C_0\Gamma_A\theta\E_{\tstart_{m,\lambda}}\left[\sum_{t\in\mathcal T_{m,\lambda}}\ip{x_t}{g_t}\right]-\frac{\E_{\tstart_{m,\lambda}}[\bar\rho_{m,\lambda}]}{32e^2\theta J},
\label{eq:strong-epoch-Vself}
\end{align}
where $C_0\triangleq2^{12}e^2$, and the last inequality uses $\E_{\tstart_{m,\lambda}}[\sum_{t\in\mathcal T_{m,\lambda}}\ell_{t,k}]\le\E_{\tstart_{m,\lambda}}[\sum_{t\in\mathcal T_{m,\lambda}}\ip{x_t}{g_t}]$, which holds since $\Reg_{m,\lambda,k}>0$, together with $\Gamma_A\ge1$ and $C_0\ge\max\{{9},12+2^{11}e^2\}$.
To bound $\E_{\tstart_{m,\lambda}}[\sum_{t\in\mathcal T_{m,\lambda}}\ip{x_t}{g_t}]$ in \pref{eq:strong-epoch-Vself}, we have
\begin{align*}
&\E_{\tstart_{m,\lambda}}\left[\sum_{t\in\mathcal T_{m,\lambda}}\ip{x_t}{g_t}\right]
=\E_{\tstart_{m,\lambda}}\left[\sum_{t\in\mathcal T_{m,\lambda}:\,t<\tend_{m,\lambda}}\widetilde Z_t+\widetilde Z_{\tend_{m,\lambda}}\right]
\le\frac{\ln(MT)}{\Gamma_A\theta^2}+3+\E_{\tstart_{m,\lambda}}[\bar\rho_{m,\lambda}],
\end{align*}
where the equality uses \pref{eq:stopped-unbiasedness}, and the inequality uses \pref{eq:strong-epoch-loss} and $\widetilde Z_{\tend_{m,\lambda}}\le3+\frac1{x_{\tend_{m,\lambda},0}}\le3+\bar\rho_{m,\lambda}$ from \pref{lem:observable-loss-bounds}. Plugging this bound into \pref{eq:strong-epoch-Vself}, we have
\begin{align*}
\Reg_{m,\lambda,k}
&\le\frac{2C_0\ln(MT)}{\theta}+3C_0\Gamma_A\theta
+\left(C_0\Gamma_A\theta-\frac1{32e^2\theta J}\right)\E_{\tstart_{m,\lambda}}[\bar\rho_{m,\lambda}]=\order\left(\frac{\ln(MT)}{\theta}\right),
\end{align*}
where the last step uses $\Gamma_A\theta\le c$ and $C_0\Gamma_A\theta\le\frac1{32e^2\theta J}$. The latter holds since $32e^2C_0\Gamma_A\theta^2J\le2^{17}e^4c^2\frac{J}{\Gamma_A}\le2^{-9}$, where we use $\theta\le\frac c{\Gamma_A}$, $c=\frac1{2^{13}e^2}$ and $\Gamma_A=AJ\ln(KT)\ge J$.

For $k\in\Vbarself$, we have $g_{t,k}=\ell_{t,k}$, so \pref{eq:stopped-unbiasedness} gives
\begin{align}
\Reg_{m,\lambda,k}=\E_{\tstart_{m,\lambda}}\left[\sum_{t\in\mathcal T_{m,\lambda}}\left(\widetilde Z_t-\widetilde g_{t,k}\right)\right].
\label{eq:strong-epoch-Vbar-target}
\end{align}
We first bound $\sum_{t\in\mathcal T_{m,\lambda}}(\widetilde Z_t-\widetilde g_{t,k})$ pathwise, and then take the expectation. Since $\widetilde Z_t=\ip{x_t}{\widetilde g_t}$, we have $\widetilde Z_t-\widetilde g_{t,k}=\ip{x_t-u^{(k)}}{\widetilde g_t}+\ip{u^{(k)}-e_k}{\widetilde g_t}$, where $u^{(k)}$ is defined in \pref{lem:strong-barrier-reward}. To bound the sum of $\ip{x_t-u^{(k)}}{\widetilde g_t}$ over the epoch, summing \pref{eq:strong-master-step-bound} with $u=u^{(k)}$ over $t\in\mathcal T_{m,\lambda}$, we have
\begin{align*}
\sum_{t\in\mathcal T_{m,\lambda}}\ip{x_t-u^{(k)}}{\widetilde g_t}
&\le D_{\Psi_{\tstart_{m,\lambda}}}(u^{(k)},x_{\tstart_{m,\lambda}})-D_{\Psi_{\tend_{m,\lambda}}}(u^{(k)},x_{\tend_{m,\lambda}+1})\\
&\phantom{{}\le{}}+\sum_{t\in\mathcal T_{m,\lambda}:\,t<\tend_{m,\lambda}}\left(D_{\Psi_{t+1}}(u^{(k)},x_{t+1})-D_{\Psi_t}(u^{(k)},x_{t+1})\right)+2\theta\sum_{t\in\mathcal T_{m,\lambda}}\mathcal V_t\\
&\le\frac{{8}\ln(MT)}{\theta}+2\theta\sum_{t\in\mathcal T_{m,\lambda}}\mathcal V_t,
\end{align*}
where the last inequality uses $D_{\Psi_{\tstart_{m,\lambda}}}(u^{(k)},x_{\tstart_{m,\lambda}})\le\frac{{8}\ln(MT)}{\theta}$ and \pref{eq:strong-barrier-zero} from \pref{lem:strong-barrier-reward} applied to this instance, and $D_{\Psi_{\tend_{m,\lambda}}}(u^{(k)},x_{\tend_{m,\lambda}+1})\ge0$. Adding $\sum_{t\in\mathcal T_{m,\lambda}}\ip{u^{(k)}-e_k}{\widetilde g_t}$ to both sides, we have
\begin{align}
\sum_{t\in\mathcal T_{m,\lambda}}(\widetilde Z_t-\widetilde g_{t,k})
\le\frac{{8}\ln(MT)}{\theta}
+2\theta\sum_{t\in\mathcal T_{m,\lambda}}\mathcal V_t
+\sum_{t\in\mathcal T_{m,\lambda}}\ip{u^{(k)}-e_k}{\widetilde g_t}.
\label{eq:strong-epoch-complement}
\end{align}
By \pref{lem:observable-loss-bounds}, $\mathcal V_t\le2\widetilde Z_t$ for $t<\tend_{m,\lambda}$ and $\mathcal V_{\tend_{m,\lambda}}\le5$, so \pref{eq:strong-epoch-loss} gives $\sum_{t\in\mathcal T_{m,\lambda}}\mathcal V_t\le\frac{2\ln(MT)}{\Gamma_A\theta^2}+5$. Moreover, since $\Et[\widetilde g_t]=g_t$ and whether $t\in\mathcal T_{m,\lambda}$ is determined before round $t$ is sampled, the tower property and \pref{lem:strong-barrier-reward} give
\begin{align}
\E_{\tstart_{m,\lambda}}\left[\sum_{t\in\mathcal T_{m,\lambda}}\ip{u^{(k)}-e_k}{\widetilde g_t}\right]=\E_{\tstart_{m,\lambda}}\left[\sum_{t\in\mathcal T_{m,\lambda}}\ip{u^{(k)}-e_k}{g_t}\right]\le\frac12.
\label{eq:strong-epoch-smoothing}
\end{align}
 Taking $\E_{\tstart_{m,\lambda}}$ of both sides of \pref{eq:strong-epoch-complement}, and using \pref{eq:strong-epoch-Vbar-target}, $\sum_{t\in\mathcal T_{m,\lambda}}\mathcal V_t\le\frac{2\ln(MT)}{\Gamma_A\theta^2}+5$ and \pref{eq:strong-epoch-smoothing}, we have
\begin{align*}
\Reg_{m,\lambda,k}
\le\frac{{8}\ln(MT)}{\theta}
+\frac{4\ln(MT)}{\Gamma_A\theta}+10\theta+\frac12=\order\left(\frac{\ln(MT)}{\theta}\right).
\end{align*}
\end{proof}

We now prove \pref{thm:strong-adaptive} by summing \pref{lem:one-loss-epoch} over the epochs.

\begin{proof}[Proof of \pref{thm:strong-adaptive}.]
Denote by $m^\star$ the index of the last meta-epoch started by \pref{alg:strong-adaptive}, by $\lambda_m^\star$ the index of the last epoch started within meta-epoch $m$, and by $S_m\triangleq\sum_{t\in\mathcal T_m}\widetilde Z_t$ the observable loss of the rounds $\mathcal T_m$ of meta-epoch $m$.
Let $A^\star\ge1$ satisfy $\mathcal H^{\Gself}(q)\le A^\star$ for every $q\in\Delta(\Vself)$, with $A^\star=\order(\alpha\ln(KT))$ by \pref{lem:soft-ratio}, and set $A^\star=1$ if $0\notin\Bcal$.
Since the condition $\mathcal H^{\Gself}(q_t)>A_m$ in \pref{alg:strong-adaptive} never holds once $A_m\ge A^\star$, and $A_{m+1}=2A_m$ with $A_1=1$, we have $A_{m^\star}<2A^\star$ and $\sum_{m=1}^{m^\star}A_m<4A^\star$.

Fix $k\in V$.
Since each epoch starts at a round determined before it is sampled, summing \pref{eq:one-loss-epoch} over the epochs with the tower property gives
\begin{align}
\Reg_{T,k}
\le\order\left(\E\left[\sum_{m=1}^{m^\star}\sum_{\lambda=1}^{\lambda_m^\star}\frac{\ln(MT)}{\theta_{m,\lambda}}\right]\right).
\label{eq:strong-adaptive-rate-bound}
\end{align}
Fix a meta-epoch $m\in[m^\star]$. Since $\theta_{m,\lambda+1}=\frac{\theta_{m,\lambda}}2$ by \pref{line:strong-adaptive-halve}, the inverse rates $\frac1{\theta_{m,\lambda}}$ double from one epoch to the next, so
\begin{align}
\sum_{\lambda=1}^{\lambda_m^\star}\frac1{\theta_{m,\lambda}}<\frac2{\theta_{m,\lambda_m^\star}}.
\label{eq:strong-adaptive-geometric}
\end{align}
To bound $\frac1{\theta_{m,\lambda_m^\star}}$ in \pref{eq:strong-adaptive-geometric}, we consider two cases. If $\lambda_m^\star=1$, then $\frac1{\theta_{m,1}}=\frac{\Gamma_{A_m}}c$. If $\lambda_m^\star\ge2$, then epoch $(m,\lambda_m^\star-1)$ ended because the condition in \pref{line:strong-adaptive-halve} held, that is, $\Gamma_{A_m}\theta_{m,\lambda_m^\star-1}^2S\ge\ln(MT)$, where $S$ is the observable loss of epoch $(m,\lambda_m^\star-1)$ and hence at most $S_m$. Therefore, $\theta_{m,\lambda_m^\star-1}\ge\sqrt{\frac{\ln(MT)}{\Gamma_{A_m}S_m}}$, and
\begin{align*}
\frac1{\theta_{m,\lambda_m^\star}}=\frac2{\theta_{m,\lambda_m^\star-1}}\le2\sqrt{\frac{\Gamma_{A_m}S_m}{\ln(MT)}}.
\end{align*}
In both cases, $\frac1{\theta_{m,\lambda_m^\star}}\le\frac{\Gamma_{A_m}}c+2\sqrt{\frac{\Gamma_{A_m}S_m}{\ln(MT)}}$. Plugging this bound into \pref{eq:strong-adaptive-geometric} gives
\begin{align*}
\sum_{\lambda=1}^{\lambda_m^\star}\frac{\ln(MT)}{\theta_{m,\lambda}}
<\frac{2\Gamma_{A_m}\ln(MT)}{c}+4\sqrt{\Gamma_{A_m}\ln(MT)S_m}.
\end{align*}
Taking the summation over $m\in[m^\star]$, we have
\begin{align}
\sum_{m=1}^{m^\star}\sum_{\lambda=1}^{\lambda_m^\star}\frac{\ln(MT)}{\theta_{m,\lambda}}
&\le\order\left(\sum_{m=1}^{m^\star}\left(\sqrt{\Gamma_{A_m}\ln(MT)S_m}+\Gamma_{A_m}\ln(MT)\right)\right)\nonumber\\*
&\le\order\left(\sqrt{\Gamma_{A^\star}\ln(MT)\sum_{t=1}^T\widetilde Z_t}+\Gamma_{A^\star}\ln(MT)\right),
\label{eq:strong-adaptive-sum}
\end{align}
where the last inequality uses $\Gamma_{A_m}=A_mJ\ln(KT)$, the Cauchy--Schwarz inequality, $\sum_{m=1}^{m^\star}A_m<4A^\star$ and $\sum_{m=1}^{m^\star}S_m=\sum_{t=1}^T\widetilde Z_t$.
Plugging \pref{eq:strong-adaptive-sum} into \pref{eq:strong-adaptive-rate-bound} and applying Jensen's inequality with $\E[\sum_{t=1}^T\widetilde Z_t]=L_{T,k}+\Reg_{T,k}$ from \pref{lem:observable-loss-bounds} yields
\begin{align*}
\Reg_{T,k}
\le\order\left(\sqrt{\Gamma_{A^\star}\ln(MT)(L_{T,k}+\Reg_{T,k})}+\Gamma_{A^\star}\ln(MT)\right).
\end{align*}
For $\Reg_{T,k}>0$, using $\sqrt{x+y}\le\sqrt x+\sqrt y$ and the AM--GM inequality, and rearranging, gives $\Reg_{T,k}=\order(\sqrt{\Gamma_{A^\star}\ln(MT)L_{T,k}}+\Gamma_{A^\star}\ln(MT))$, which is $\otil(\sqrt{\alpha L_{T,k}}+\alpha)$ since $\Gamma_{A^\star}=\otil(\alpha)$. The case $\Reg_{T,k}\le0$ is immediate.
\end{proof}

\section{Smoothing in Fixed-Point Mapping}
\label{app:u-unsmoothed}

We show that the unsmoothed equation \pref{eq:u-coordinate-design} cannot ensure that $Z_0(p,q)=\otil(\eta\alpha)$.
The following example shows that $Z_0(p,q)$ can grow linearly with $K$ even when $\alpha=1$.

\begin{proposition}
\label[proposition]{prop:u-unsmoothed-obstruction}
For every $0<\eta\le\frac1{K+1}$, there exist a self-aware graph $G$ with independence number $\alpha=1$ and a distribution $p\in\operatorname{int}(\DeltaK)$ such that \pref{eq:u-coordinate-design} has an interior fixed point $q$ satisfying
$
  \sum_{j=1}^K\frac{p_j}{Q_j(q)+\eta}=\frac{K+1}{2+\eta(K+1)}.
$
Moreover, every fixed point $r\in\DeltaK$ of the same equation satisfies $Z_0(p,r)\ge\frac{\eta(K+1)}{2+\eta(K+1)}\ge\frac{\eta(K+1)}3$.
\end{proposition}

\begin{proof}
Let $G$ have the edges $i\to j$ exactly when $i\ge j$.
Then every vertex has a self-loop, and every pair of distinct vertices has an edge, so $\alpha=1$.
Choose $p\in\operatorname{int}(\DeltaK)$ with coordinates
\begin{equation*}
  p_j=
  \begin{cases}
    \dfrac{2^{1-j}+\eta}{2+\eta(K+1)},&j<K,\\[4pt]
    \dfrac{2^{2-K}+2\eta}{2+\eta(K+1)},&j=K.
  \end{cases}
\end{equation*}
For this $p$, we claim that the distribution with coordinates
\begin{equation*}
  q_j=
  \begin{cases}
    2^{-j},&j<K,\\
    2^{1-K},&j=K
  \end{cases}
\end{equation*}
is an interior fixed point.
To verify the claim, the observation probabilities satisfy $Q_j(q)=\sum_{i=j}^Kq_i=2^{1-j}$, and substitution of $p_j$ gives
\begin{equation*}
  p_j\phi(Q_j(q))
  =\frac{p_jQ_j(q)}{Q_j(q)+\eta}
  =\frac{2q_j}{2+\eta(K+1)}
\end{equation*}
for every $j\in[K]$.
Summing over $j$ gives $\sum_{j=1}^Kp_j\phi(Q_j(q))=\frac2{2+\eta(K+1)}$, so normalization yields \pref{eq:u-coordinate-design} and $Z_0(p,q)=\frac{\eta(K+1)}{2+\eta(K+1)}$.
Moreover,
\begin{equation*}
  \sum_{j=1}^K\frac{p_j}{Q_j(q)+\eta}
  =\frac2{2+\eta(K+1)}\sum_{j=1}^K\frac{q_j}{Q_j(q)}
  =\frac2{2+\eta(K+1)}\left(\frac{K-1}{2}+1\right)
  =\frac{K+1}{2+\eta(K+1)}.
\end{equation*}

It remains to show that no other fixed point gives a smaller coefficient.
Fix any fixed point $r\in\DeltaK$, denote by $R\triangleq\max_{i\in[K]}r_i/q_i$ the largest coordinate ratio, and take $j\in[K]$ attaining it.
Since $R\ge1$ and $Q_j(r)\le RQ_j(q)$, we obtain
\begin{align*}
  1-Z_0(p,r)
  =\frac{p_j\phi(Q_j(r))}{r_j}
  \le\frac{p_j\phi(RQ_j(q))}{Rq_j}=\frac2{2+\eta(K+1)}\frac{Q_j(q)+\eta}{RQ_j(q)+\eta}
  \le\frac2{2+\eta(K+1)},
\end{align*}
where the first inequality uses that $\phi(s)=s/(s+\eta)$ is increasing.
Thus $Z_0(p,r)\ge\frac{\eta(K+1)}{2+\eta(K+1)}\ge\frac{\eta(K+1)}3$.
\end{proof}
\section{Auxiliary Lemma}\label{app:aux}
In this section, we prove \pref{lem:soft-ratio}, which is a direct implication of Lemma 1 of~\citet{kocak2014implicit}.
We use the case $m=1$ of their lemma, stated in our notation as follows.

\begin{lemma}[Lemma 1 of~\citet{kocak2014implicit}]\label{lem:kocak}
Let $G'=(V',E')$ be a self-aware directed graph with $|V'|=d$, and denote by $\alpha'$ its independence number. For every $c>0$ and nonnegative numbers $(p_i)_{i\in V'}$ with $\sum_{i\in V'}p_i\le1$, we have
$$
  \sum_{i\in V'}\frac{p_i}{c+\sum_{j\in V':j\to i}p_j}
  \le
  2\alpha'\ln\left(1+\frac{\lceil d^2/c\rceil+d}{\alpha'}\right)+2.
$$
\end{lemma}

\begin{proof}[Proof of~\pref{lem:soft-ratio}]
If $\Vself=\emptyset$, then $\mathcal H^G(q)=0$ and the bound is immediate. Otherwise, recall that $\Gself$ is the subgraph induced by $\Vself$. As $G$ is strongly observable, we know that $\alpha'=\alpha$.

To compare $\mathcal H^G(q)$ with the left-hand side of \pref{lem:kocak}, for every $i\in\Vself$, we have
$$
  Q_i^G(q)=\sum_{j\in V:j\to i}q_j\ge \sum_{j\in\Vself:j\to i\text{ in }\Gself}q_j,
$$
since the sum contains the term $q_i$ from the self-loop at $i$, and we drop the nonnegative terms from the vertices in $V\setminus\Vself$. Since $\sum_{i\in\Vself}q_i\le1$, we apply \pref{lem:kocak} to $\Gself$ with $d=|\Vself|$, $p_i=q_i$ for $i\in\Vself$, and $c=1/(KT)$, and obtain that
\begin{align*}
  \mathcal H^G(q)
  &\le\sum_{i\in\Vself}\frac{q_i}{\sum_{j\in\Vself:j\to i\text{ in }\Gself}q_j+1/(KT)}\\
  &\le2\alpha'\ln\left(1+\frac{|\Vself|^2KT+|\Vself|}{\alpha'}\right)+2
  \tag{\pref{lem:kocak}}\\
  &\le2\alpha\ln\left(1+\frac{K^3T+K}{\alpha}\right)+2,
\end{align*}
where the last inequality uses $|\Vself|\le K$.
\end{proof}
\section*{AI Methodology Statement}
The authors used GPT-5.6 and GPT-6 Astra to generate an initial algorithmic proposal and discuss its analysis. The authors also used Opus 5.5 to assist with language editing. The authors substantially simplified and revised the proposed algorithm and worked through and refined its analysis. The authors determined the paper's organization and extensively rewrote the generated material. The authors take full responsibility for the correctness, originality, and final content of the paper.
\bibliographystyle{alpha}
\bibliography{references}

@inproceedings{eldowa2023minimax,
  author = {Eldowa, Khaled and Esposito, Emmanuel and Cesari, Tom and Cesa-Bianchi, Nicol{\`o}},
  title = {On the minimax regret for online learning with feedback graphs},
  booktitle = {Advances in Neural Information Processing Systems},
  year = {2023},
}

@inproceedings{allenberg2006hannan,
  author = {Allenberg, Chamy and Auer, Peter and Gy{\"o}rfi, L{\'a}szl{\'o} and Ottucs{\'a}k, Gy{\"o}rgy},
  title = {Hannan consistency in on-line learning in case of unbounded losses under partial monitoring},
  booktitle = {Algorithmic Learning Theory},
  year = {2006},
}

@inproceedings{arora2019switching,
  author = {Raman Arora and Teodor Vanislavov Marinov and Mehryar Mohri},
  title = {Bandits with Feedback Graphs and Switching Costs},
  booktitle = {Advances in Neural Information Processing Systems},
  year = {2019},
}

@inproceedings{esposito2022edge,
  author = {Emmanuel Esposito and Federico Fusco and Dirk van der Hoeven and Nicol{\`o} Cesa-Bianchi},
  title = {Learning on the Edge: Online Learning with Stochastic Feedback Graphs},
  booktitle = {Advances in Neural Information Processing Systems},
  year = {2022},
}

@inproceedings{marinov2022stochastic,
  author = {Teodor Vanislavov Marinov and Mehryar Mohri and Julian Zimmert},
  title = {Stochastic Online Learning with Feedback Graphs: Finite-Time and Asymptotic Optimality},
  booktitle = {Advances in Neural Information Processing Systems},
  year = {2022},
}

@inproceedings{ito2022nearly,
  author = {Shinji Ito and Taira Tsuchiya and Junya Honda},
  title = {Nearly Optimal Best-of-Both-Worlds Algorithms for Online Learning with Feedback Graphs},
  booktitle = {Advances in Neural Information Processing Systems},
  year = {2022},
}

@inproceedings{rouyer2022near,
  author = {Chlo{\'e} Rouyer and Dirk van der Hoeven and Nicol{\`o} Cesa-Bianchi and Yevgeny Seldin},
  title = {A Near-Optimal Best-of-Both-Worlds Algorithm for Online Learning with Feedback Graphs},
  booktitle = {Advances in Neural Information Processing Systems},
  year = {2022},
}

@inproceedings{wen2024contextual,
  author = {Yuxiao Wen and Yanjun Han and Zhengyuan Zhou},
  title = {Stochastic Contextual Bandits with Graph Feedback: From Independence Number to {MAS} Number},
  booktitle = {Advances in Neural Information Processing Systems},
  year = {2024},
}

@inproceedings{wen2025combinatorial,
  author = {Yuxiao Wen},
  title = {Adversarial Combinatorial Semi-bandits with Graph Feedback},
  booktitle = {International Conference on Machine Learning},
  year = {2025},
}

@inproceedings{zhang2023practical,
  author = {Mengxiao Zhang and Yuheng Zhang and Olga Vrousgou and Haipeng Luo and Paul Mineiro},
  title = {Practical Contextual Bandits with Feedback Graphs},
  booktitle = {Advances in Neural Information Processing Systems},
  year = {2023},
}

@inproceedings{zhang2024uninformed,
  author = {Mengxiao Zhang and Yuheng Zhang and Haipeng Luo and Paul Mineiro},
  title = {Efficient Contextual Bandits with Uninformed Feedback Graphs},
  booktitle = {International Conference on Machine Learning},
  year = {2024},
}

@inproceedings{mannor2011side,
  author = {Mannor, Shie and Shamir, Ohad},
  title = {From Bandits to Experts: On the Value of Side-Observations},
  booktitle = {Advances in Neural Information Processing Systems},
  year = {2011},
}

@inproceedings{alon2013domination,
  author = {Alon, Noga and Cesa-Bianchi, Nicol\`o and Gentile, Claudio and Mansour, Yishay},
  title = {From Bandits to Experts: A Tale of Domination and Independence},
  booktitle = {Advances in Neural Information Processing Systems},
  year = {2013},
}

@inproceedings{alon2015feedback,
  author = {Alon, Noga and Cesa-Bianchi, Nicol\`o and Dekel, Ofer and Koren, Tomer},
  title = {Online Learning with Feedback Graphs: Beyond Bandits},
  booktitle = {Conference on Learning Theory},
  year = {2015},
}

@article{alon2017nonstochastic,
  author = {Alon, Noga and Cesa-Bianchi, Nicol\`o and Gentile, Claudio and Mannor, Shie and Mansour, Yishay and Shamir, Ohad},
  title = {Nonstochastic Multi-Armed Bandits with Graph-Structured Feedback},
  journal = {SIAM Journal on Computing},
  volume = {46},
  number = {6},
  year = {2017},
}

@inproceedings{cohen2016without,
  author = {Cohen, Alon and Hazan, Tamir and Koren, Tomer},
  title = {Online Learning with Feedback Graphs Without the Graphs},
  booktitle = {International Conference on Machine Learning},
  year = {2016},
}

@inproceedings{luo2023highprob,
  author = {Luo, Haipeng and Tong, Hanghang and Zhang, Mengxiao and Zhang, Yuheng},
  title = {Improved High-Probability Regret for Adversarial Bandits with Time-Varying Feedback Graphs},
  booktitle = {Algorithmic Learning Theory},
  year = {2023},
}

@inproceedings{kocak2023shape,
  author = {Koc\'ak, Tom\'a\v{s} and Carpentier, Alexandra},
  title = {Online Learning with Feedback Graphs: The True Shape of Regret},
  booktitle = {International Conference on Machine Learning},
  year = {2023},
}

@article{he2023decomposition,
  author = {He, Yuchen and Zhang, Chihao},
  title = {Improved Algorithms for Bandit with Graph Feedback via Regret Decomposition},
  journal = {Theoretical Computer Science},
  volume = {979},
  year = {2023},
}

@inproceedings{lykouris2018small,
  author = {Lykouris, Thodoris and Sridharan, Karthik and Tardos, {\'E}va},
  title = {Small-Loss Bounds for Online Learning with Partial Information},
  booktitle = {Conference on Learning Theory},
  year = {2018},
}

@inproceedings{lee2020graph,
  author = {Lee, Chung-Wei and Luo, Haipeng and Zhang, Mengxiao},
  title = {A Closer Look at Small-Loss Bounds for Bandits with Graph Feedback},
  booktitle = {Conference on Learning Theory},
  year = {2020},
}

@inproceedings{kocak2014implicit,
  author = {Koc\'ak, Tom\'a\v{s} and Neu, Gergely and Valko, Michal and Munos, R\'emi},
  title = {Efficient Learning by Implicit Exploration in Bandit Problems with Side Observations},
  booktitle = {Advances in Neural Information Processing Systems},
  year = {2014},
}

@inproceedings{neu2015explore,
  author = {Neu, Gergely},
  title = {Explore No More: Improved High-Probability Regret Bounds for Non-Stochastic Bandits},
  booktitle = {Advances in Neural Information Processing Systems},
  year = {2015},
}

@inproceedings{neu2015firstorder,
  author = {Neu, Gergely},
  title = {First-Order Regret Bounds for Combinatorial Semi-Bandits},
  booktitle = {Conference on Learning Theory},
  year = {2015},
}

@inproceedings{allenzhu2018minority,
  author = {Allen-Zhu, Zeyuan and Bubeck, Sebastien and Li, Yuanzhi},
  title = {Make the Minority Great Again: First-Order Regret Bound for Contextual Bandits},
  booktitle = {International Conference on Machine Learning},
  year = {2018},
}

@inproceedings{wei2018adaptive,
  author = {Wei, Chen-Yu and Luo, Haipeng},
  title = {More Adaptive Algorithms for Adversarial Bandits},
  booktitle = {Conference on Learning Theory},
  year = {2018},
}

@inproceedings{lattimore2020exploration,
  author = {Lattimore, Tor and Szepesv\'ari, Csaba},
  title = {Exploration by Optimisation in Partial Monitoring},
  booktitle = {Conference on Learning Theory},
  year = {2020},
}

@inproceedings{lattimore2019information,
  author = {Lattimore, Tor and Szepesv\'ari, Csaba},
  title = {An Information-Theoretic Approach to Minimax Regret in Partial Monitoring},
  booktitle = {Conference on Learning Theory},
  year = {2019},
}

@inproceedings{zimmert2019connections,
  author = {Zimmert, Julian and Lattimore, Tor},
  title = {Connections Between Mirror Descent, {Thompson} Sampling and the Information Ratio},
  booktitle = {Advances in Neural Information Processing Systems},
  year = {2019},
}

@inproceedings{tsuchiya2024exo,
  author = {Tsuchiya, Taira and Ito, Shinji and Honda, Junya},
  title = {Exploration by Optimization with Hybrid Regularizers: Logarithmic Regret with Adversarial Robustness in Partial Monitoring},
  booktitle = {International Conference on Machine Learning},
  year = {2024},
}

@article{auer2002nonstochastic,
  author = {Auer, Peter and Cesa-Bianchi, Nicol\`o and Freund, Yoav and Schapire, Robert E.},
  title = {The Nonstochastic Multiarmed Bandit Problem},
  journal = {SIAM Journal on Computing},
  volume = {32},
  number = {1},
  year = {2002},
}

@article{freund1997decision,
  author = {Freund, Yoav and Schapire, Robert E.},
  title = {A Decision-Theoretic Generalization of On-Line Learning and an Application to Boosting},
  journal = {Journal of Computer and System Sciences},
  volume = {55},
  number = {1},
  year = {1997},
}

@article{brouwer1911abbildung,
  author = {Brouwer, L. E. J.},
  title = {{\"U}ber {Abbildung} von {Mannigfaltigkeiten}},
  journal = {Mathematische Annalen},
  volume = {71},
  number = {1},
  year = {1911},
}

@inproceedings{agarwal2017corralling,
  author = {Alekh Agarwal and Haipeng Luo and Behnam Neyshabur and Robert E. Schapire},
  title = {Corralling a Band of Bandit Algorithms},
  booktitle = {Conference on Learning Theory},
  year = {2017},
}

@article{banach1922operations,
  author = {Banach, Stefan},
  title = {Sur les op{\'e}rations dans les ensembles abstraits et leur application aux {\'e}quations int{\'e}grales},
  journal = {Fundamenta Mathematicae},
  volume = {3},
  year = {1922},
}

@article{picard1890memoire,
  author = {Picard, {\'E}mile},
  title = {M{\'e}moire sur la th{\'e}orie des {\'e}quations aux d{\'e}riv{\'e}es partielles et la m{\'e}thode des approximations successives},
  journal = {Journal de Math{\'e}matiques Pures et Appliqu{\'e}es},
  volume = {6},
  year = {1890},
}

\end{document}